\documentclass[runningheads]{llncs}
\usepackage{graphicx}
\usepackage{tikz}
\usepackage{comment}
\usepackage{amsmath,amssymb} % define this before the line numbering.
\usepackage{color}

\usepackage[accsupp]{axessibility}  % Improves PDF readability for those with disabilities.

\usepackage{booktabs}
\usepackage{subcaption}
\usepackage{float}
\usepackage[group-separator={,}]{siunitx}
\usepackage{bm}
\usepackage[pagebackref=true,breaklinks=true,letterpaper=true,colorlinks,citecolor=citecolor,bookmarks=false]{hyperref}
\definecolor{citecolor}{RGB}{30,130,255}

\usepackage{etoolbox}
\renewrobustcmd{\bfseries}{\fontseries{b}\selectfont}

\newcommand\textoverlay[2]{%
\begin{tikzpicture}%
\draw (0, 0) node[inner sep=0] {#2};%
\draw (current bounding box.north west) [anchor=north west] node[text=white] {\small #1};%
\end{tikzpicture}%
}

\begin{document}
% \linenumbers
\pagestyle{headings}
\mainmatter
\def\ECCVSubNumber{4739}  % Insert your submission number here

\title{Hallucinating Pose-Compatible Scenes} % Replace with your title

\begin{comment}
\titlerunning{ECCV-22 submission ID \ECCVSubNumber}
\authorrunning{ECCV-22 submission ID \ECCVSubNumber}
\author{Anonymous ECCV submission}
\institute{Paper ID \ECCVSubNumber}
\end{comment}
%******************

% \begin{comment}
\titlerunning{Hallucinating Pose-Compatible Scenes}
\author{Tim Brooks \qquad Alexei A. Efros}
\authorrunning{T. Brook \and A.A. Efros}
\institute{UC Berkeley}
% \end{comment}
%******************
\maketitle

\begin{abstract}
What does human pose tell us about a scene? We propose a task to answer this question: given human pose as input, hallucinate a compatible scene. Subtle cues captured by human pose --- action semantics, environment affordances, object interactions --- provide surprising insight into which scenes are compatible. We present a large-scale generative adversarial network for pose-conditioned scene generation. We significantly scale the size and complexity of training data, curating a massive meta-dataset containing over 19 million frames of humans in everyday environments. We double the capacity of our model with respect to StyleGAN2 to handle such complex data, and design a pose conditioning mechanism that drives our model to learn the nuanced relationship between pose and scene. We leverage our trained model for various applications: hallucinating pose-compatible scene(s) with or without humans, visualizing incompatible scenes and poses, placing a person from one generated image into another scene, and animating pose. Our model produces diverse samples and outperforms pose-conditioned StyleGAN2 and Pix2Pix/Pix2PixHD baselines in terms of accurate human placement (percent of correct keypoints) and quality (Fréchet inception distance).
\end{abstract}

\section{Introduction}

\begin{figure}[t]
  \captionsetup[subfigure]{labelformat=empty}
  \centering
  \begin{subfigure}[b]{0.81\linewidth}
    \includegraphics[width=0.19\linewidth]{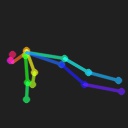}%
    \hfill%
    \hfill%
    \includegraphics[width=0.19\linewidth]{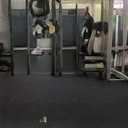}%
    \hfill%
    \includegraphics[width=0.19\linewidth]{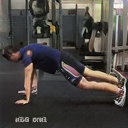}%
    \hfill%
    \hfill%
    \includegraphics[width=0.19\linewidth]{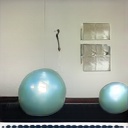}%
    \hfill%
    \includegraphics[width=0.19\linewidth]{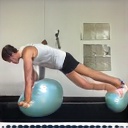}%
  \end{subfigure}
  \begin{subfigure}[b]{0.81\linewidth}
    \includegraphics[width=0.19\linewidth]{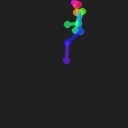}%
    \hfill%
    \hfill%
    \includegraphics[width=0.19\linewidth]{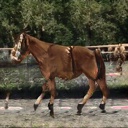}%
    \hfill
    \includegraphics[width=0.19\linewidth]{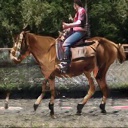}%
    \hfill%
    \hfill%
    \includegraphics[width=0.19\linewidth]{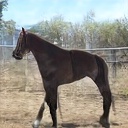}%
    \hfill%
    \includegraphics[width=0.19\linewidth]{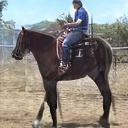}%
  \end{subfigure}
  \begin{subfigure}[b]{0.81\linewidth}
    \includegraphics[width=0.19\linewidth]{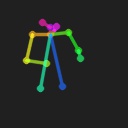}%
    \hfill%
    \hfill%
    \includegraphics[width=0.19\linewidth]{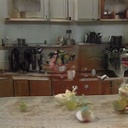}%
    \hfill%
    \includegraphics[width=0.19\linewidth]{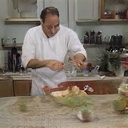}%
    \hfill%
    \hfill%
    \includegraphics[width=0.19\linewidth]{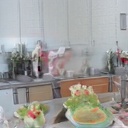}%
    \hfill%
    \includegraphics[width=0.19\linewidth]{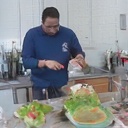}%
  \end{subfigure}

   \smallskip

  \begin{subfigure}[b]{0.8\linewidth}
    \begin{subfigure}[b]{0.19\linewidth}
        \caption{Input poses}
    \end{subfigure}
    \begin{subfigure}[b]{0.8\linewidth}
        \caption{Sample output scenes}
    \end{subfigure}
  \end{subfigure}

  \caption{Given a human pose as input, the goal of this paper is to hallucinate scene(s) that are compatible with that pose. Our model can generate isolated scenes as well as scenes containing humans.
}
  \label{fig:teaser}

\end{figure}

Human pose can reveal a lot about a scene. For example, mime artists\footnote{For those unfamiliar with mime artists, here is a wonderful example performance: \url{https://youtu.be/FPMBV3rd_hI}} invoke vivid scenes in a viewer's mind through pose and movement alone, despite performing on a bare stage. The viewer is able to imagine the invisible objects and scene elements because of the strong relationship between human poses and scenes learned through a lifetime of daily observations.

Psychologists have long been interested in understanding this symbiotic relationship between human and scene~\cite{gibson1979,biederman1981}.  J.J.~Gibson proposed the notion of {\em affordances}~\cite{gibson1979}, which can be described as “opportunities for interactions” furnished by the environment.  In computer vision, affordances have been used to provide a functional description of the scene. Given an image, a number of approaches try to predict likely human poses these scenes afford~\cite{GuptaSatkinEfrosHebertCVPR11,delaitre2012,fouhey2015defense,Hassan2021}.

This work, on the other hand, considers the opposite problem: given a human pose as input, the goal is to hallucinate scene(s) that are compatible with that pose. Consider Figure~\ref{fig:teaser}. A push-up pose (top) places severe constraints on the space of compatible scenes: they must not only be semantically compatible (e.g., gym, exercise room), but also have compatible spatial affordances (enough floor space or appropriate equipment). Objects in the scene can afford interaction with the human (e.g., squishing down an exercise ball). Other poses might not appear as constraining, but even a simple standing pose (bottom) --- head looking down, hands reaching in, legs occluded --- is actually a strong indicator of a cooking scene, and signals that an object (e.g., countertop) must be occluding the legs.

Rather than explicitly model scene affordances and contextual compatibility, we employ a modern large-scale generative model (based on a souped-up StyleGAN2~\cite{karras2020analyzing} architecture) to {\em discover} these relationships implicitly, from data. While GANs have performed well at capturing disentangled visual models in specialized scenarios (e.g., faces, churches, categories from ImageNet~\cite{imagenetcvpr09}), they have not been demonstrated {\em in situ}, on complex, real-world data across varying environments.

We curate a massive meta-dataset of humans interacting with everyday environments, containing over $19$ million frames. The complexity and scale of data is much higher than common GAN datasets, such as FFHQ~\cite{karras2019style} (\num{70000} face images) and ImageNet~\cite{imagenetcvpr09} (1.3M object images). With an appropriate pose conditioning mechanism, increased model capacity, and removal of style mixing, we are able to successfully train a pose-conditioned GAN on this highly complex data. Our model and meta-dataset mark substantial progress leveraging GANs in real-world settings containing humans and diverse environments. Through numerous visual experiments, we demonstrate our model's emergent ability to capture affordances and contextual relationships between poses and scenes.

See our webpage\footnote{\url{https://www.timothybrooks.com/tech/hallucinating-scenes}} for our supplemental video and code release.

\section{Related Work}
\label{sec:related}

\paragraph{Scene and object affordances.} Affordances~\cite{gibson1979} describe the possible uses of a given object or environment. A significant body of work learns scene affordances, such as where a person can stand or sit, from observing data of humans~\cite{GuptaSatkinEfrosHebertCVPR11,Grabner2011chair,fouhey2012people,delaitre2012,Jiang2013CVPR,fouhey2015defense,WangaffordanceCVPR2017,chuang2018learning,3d-affordance}. Overlapping areas of work focus on human interactions with objects~\cite{Yao2010ModelingMC,koppula2013learning,zhu2014reasoning,gkioxari2018detecting,cao2020reconstructing} or synthesize human pose conditioned on an input scene~\cite{lee2002interactive,caoHMP2020,wang2020synthesizing}. We propose the reverse task of hallucinating a scene conditioned on pose.

\paragraph{Pose-conditioned human synthesis.}
There are a plethora of methods that take a source image (or video) of a human plus a new pose and generate an image of the human in the new pose~\cite{ma2017pose,siarohin2018deformable,balakrishnan2018synthesizing,wang2018vid2vid,aberman2019deep,chan2019everybody,Li2019CVPR}.
Although we too condition on pose, our goals are almost entirely opposite: we aim to generate novel scenes compatible with a given pose, whereas the above methods reuse the scene from the source image/video and only focus on reposing within that provided scene.

\paragraph{GANs for image synthesis.} Introduced by Goodfellow \textit{et al.}~\cite{goodfellow2014generative} a generative adversarial network (GAN) is an implicit generative model that learns to synthesize data samples by optimizing a minimax objective. The generator is tasked with fooling a discriminator, and the discriminator is tasked with differentiating real and generated samples. Modern GANs are capable of producing high quality images~\cite{brock2018large,karras2018progressive,karras2019style,karras2020analyzing}. Image translation~\cite{isola2017image,wang2018vid2vid} utilizes conditional GANs~\cite{mirza2014conditional} to translate from one domain to another. While our task is pose-conditional scene generation, we leverage benefits of modern unconditional GANs~\cite{karras2020analyzing}.

\paragraph{Visual disentanglement.} Disentanglement methods attempt to separate out independent controllable attributes of images. This can be achieved with unsupervised methods~\cite{karras2019style,ganspace2020,gansteerability,peebles2020hessian}, or an auxiliary signal~\cite{goetschalckx2019ganalyze,li-cvpr2020}. Components of image samples can be added, removed and composed using pretrained GANs~\cite{bau2019gandissect,chai2021latent}. Recent work has applied similar strategies to image translation models to compose style and content from different images~\cite{park2020swapping}.
The most related to us is the work of Ma {\em et al.}~\cite{ma2017disentangled}, who synthesize images of people, while independently controlling foreground, background, and pose. However, the focus is on generating humans in very tightly cropped images with simple backgrounds, rather than generating scenes with appropriate affordances.
Many disentanglement methods assume all images or image attributes can be combined with all others~\cite{karras2019style,ganspace2020,goetschalckx2019ganalyze,gansteerability,peebles2020hessian,park2020swapping,li-cvpr2020,ma2017disentangled}. In this work, we seek disentangled representations of pose, human appearance and scene, yet it is essential our model understand which scenes can or cannot be composed with which poses.

\paragraph{Contextual relationships.} Many works leverage contextual relationships among objects and scenes~\cite{biederman1981} to improve vision models  such as object recognition and semantic segmentation ~\cite{torralba2003context,rabinovich2007objects,divvala2009empirical,mottaghi2014role}.
Divvala {\em et al.}~\cite{divvala2009empirical} explicitly enumerate (Table 1) a taxonomy of possible contextual information.
In this paper we are specifically interested in contextual relationships between humans and their environments, and aim to recover them implicitly, from data.

\section{{\em Humans in Context} Meta-dataset}
\label{sec:dataset}

\begin{figure}[t]
    \centering

    \begin{subfigure}[b]{0.25\linewidth}
        \centering
        \includegraphics[width=.98\linewidth]{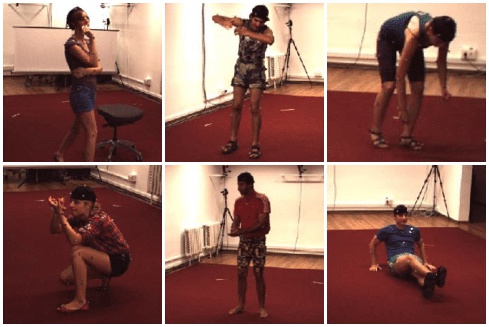}%
        \caption{Humans 3.6M~\cite{ionescu2013human3}}%
        \label{fig:humans_36m}%
    \end{subfigure}%
    \hfill\hfill\hfill\hfill
    \begin{subfigure}[b]{0.25\linewidth}
        \centering
        \includegraphics[width=.98\linewidth]{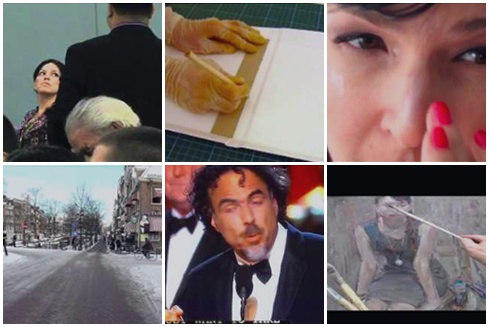}%
        \caption{Kinetics~\cite{kay2017kinetics}}%
        \label{fig:kinetics}%
    \end{subfigure}%
    \hfill
    \begin{subfigure}[b]{0.495\linewidth}
        \centering
        \includegraphics[width=.985\linewidth]{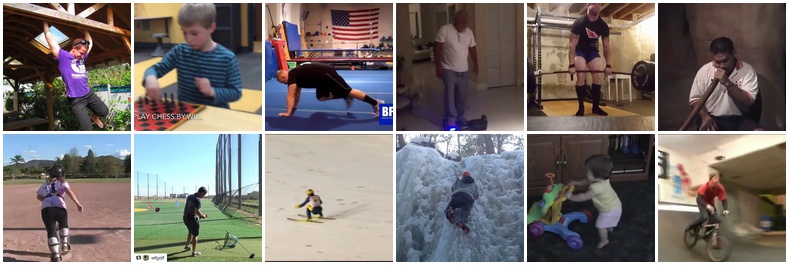}%
        \caption{Humans in Context (ours)}%
        \label{fig:humans_dataset}%
    \end{subfigure}
    \caption{\textbf{Dataset comparison.} (a) The largest human-centric video dataset with ground truth poses uses a fixed background, missing scene interactions. (b) Action recognition datasets include scenes, but contain videos without people or of close-up content. (c) Our dataset is a massive curation of humans in scenes.}

    \label{fig:dataset}
\end{figure}

To study the rich relationship between scenes and human poses requires large-scale data of people interacting with many different environments. Internet videos are a natural source, containing vast data of daily human activities. Unfortunately, large-scale action recognition datasets~\cite{diba2020large,monfort2019moments,kay2017kinetics} include substantial content without humans, as well as close-up footage not of scenes. Most existing human-centric datasets are insufficiently small~\cite{sigurdsson2016hollywood,andriluka20142d}, narrow in scene type~\cite{fouhey2018lifestyle,zhang2013actemes}, or captured on a fixed background~\cite{ionescu2013human3}.

We therefore curate a meta-dataset of \num{229595} video clips, each containing a single person in a scene, sourced from 10 existing human and action recognition video datasets~\cite{diba2020large,monfort2019moments,kay2017kinetics,sigurdsson2016hollywood,kanazawa2019learning,epstein2020oops,andriluka20142d,fouhey2018lifestyle,zhang2013actemes,xu2018youtube}, and supplemented with pseudo-ground truth pose obtained using OpenPose~\cite{cao2017realtime,cao2019openpose}. Video offers a massive source of real-world data, and ensures all poses of human activity are represented, rather than only poses photographers choose to capture in still images.

Videos are extensively filtered for quality, ensuring satisfactory framerate, bitrate and resolution. \num{1509032} videos ($75 \%$ of source videos) pass quality filtering. Frames are then filtered with pretrained Keypoint R-CNN~\cite{he2017mask,wu2019detectron2} person detection and OpenPose~\cite{cao2017realtime,cao2019openpose} keypoint prediction models. The final dataset only includes clips of at least 30 frames where Keypoint R-CNN detects a single person and OpenPose predicts sufficient keypoints. This results in \num{19503700} frames ($7.8 \%$ of high quality frames), with each clip averaging $85$ frames long.

While we train on images, we split data into partitions based on video clips, reserving \num{12800} clips for testing and the remaining \num{216795} for training. See the supplement for dataset details.

\section{Pose-compatible Scene GAN}
\label{sec:method}

\begin{figure*}[t]

    \includegraphics[width=\linewidth]{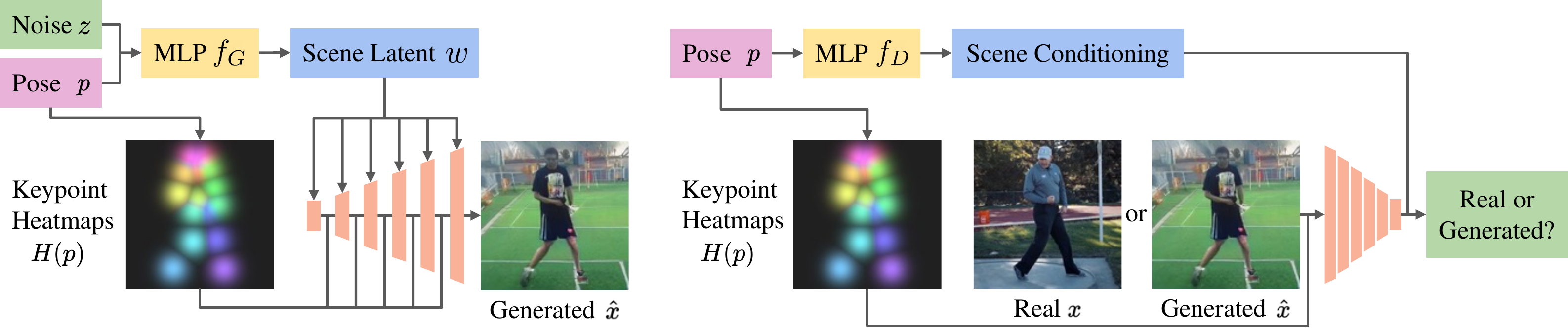}%
    \par\smallskip%
    \begin{subfigure}[b]{0.49\linewidth}
        \caption{Generator}
    \end{subfigure}
    \begin{subfigure}[b]{0.38\linewidth}
        \caption{Discriminator}
    \end{subfigure}%
    \caption{\textbf{Network architectures.} Our networks are based on StyleGAN2, with simple modifications to ensure accurately placed humans and compatible scenes. In particular, the conditional generator and discriminator networks utilize pose $p$ via two mechanisms: keypoint heatmaps and pose latent conditioning. Keypoint heatmaps correctly positions a human, and pose latent conditioning drives latent codes $w$ to generate compatible scenes. Multiple plausible scenes can be produced for the same input pose by sampling different noise vectors $z$.}
    \label{fig:networks}

\end{figure*}

We design a conditional GAN~\cite{goodfellow2014generative,mirza2014conditional} to produce scenes compatible with human pose. Our network architectures are based on StyleGAN2~\cite{karras2020analyzing} and are depicted in Figure~\ref{fig:networks}. Generating high quality pose-compatible scenes arises from simple yet important modifications: dual pose conditioning, removal of style mixing, and large-scale training. Our model can produce isolated scene images without any human by zeroing out keypoint heatmaps when generating images.

\subsection{Dual pose conditioning}
The conditional generator $G$ and discriminator $D$ both utilize input pose via two mechanisms: keypoint heatmap conditioning, which specifies spatial placement of a human subject, and pose latent conditioning, which infers compatible scenes. To succeed at our task, humans must be positioned correctly and generated scenes must be compatible. Dual pose conditioning drives strong performance in both respects, and outperforms conditioning on either alone in our ablation experiment (Table~\ref{tab:conditioning}). Furthermore, dual pose conditioning disentangles control of scene and human pose. We leverage these separate controls for numerous applications: generating scenes without humans, visualizing incompatible scenes and poses, placing a person in a new scene, and animating pose.

\paragraph{Keypoint heatmaps.} Let pose $p = (p_1, ..., p_K)$ denote 2D locations of the $K=18$ human keypoints detected by OpenPose~\cite{cao2019openpose}, and let $v = (v_1, ..., v_K)$ indicate visibility of each keypoint. Following the works of~\cite{siarohin2018deformable,balakrishnan2018synthesizing,aberman2019deep}, our keypoint heatmaps $H(p)$ consist of radial basis function kernels centered at each keypoint. For heatmap $k \in \{1, ..., K\}$, the intensity at location $q$ is given by Equation~\ref{eq:heatmaps}. We concatenate heatmaps at each scale of the generator, and at the input of the discriminator. We set $\sigma^2 = \max(0.5, 0.005R^2)$ where $R$ is the spatial resolution of the heatmaps. After training, we generate images of scenes without humans by simply zeroing out all keypoint heatmaps.

\begin{equation}
H_{k, q}(p) =
\begin{cases}
    \exp \left( - \frac{||q - p_k||^2}{2\sigma^2} \right)   & \text{if } v_k = 1 \\
    0                                                       & \text{otherwise}
\end{cases}
\label{eq:heatmaps}
\end{equation}

\paragraph{Pose latent conditioning.} To generate compatible scenes, we condition scene latent codes on the input pose. Akin to intermediate latents in StyleGAN2~\cite{karras2020analyzing}, the scene latent code $w$ controls generation by modulating convolutional weights. To condition the latent code, pose locations and visibility are flattened and mapped to a $512$-dimensional input via a learned linear projection. A noise sample $z \sim \mathcal{Z}$ is concatenated with the input vector and passed through a multi-layer perceptron (MLP) $f_G$ to produce a scene latent code $w \in \mathcal{W}$. Multiple plausible scenes can be generated by sampling different noise vectors $z$ for the same pose. The discriminator learns a separate linear projection and MLP $f_D$.

\subsection{Removal of style mixing}
Style mixing regularization~\cite{karras2019style,park2020swapping} encourages disentanglement by randomly mixing intermediate latent codes during training. The technique assumes image attributes at each layer are compatible with all other image attributes (e.g.\ any face could have any color hair). This assumption is not true when composing scenes and humans, which we visually demonstrate through the incompatible scenes and poses in Figure~\ref{fig:compat}. This motivates removing style mixing regularization during training, which improves results in our ablation experiments (Table~\ref{tab:stylegan2_a}).

\subsection{Large-scale GAN training}
Typical datasets used with StyleGAN2 (e.g.\ faces, bedrooms, churches~\cite{liu2015faceattributes,lsun}) are relatively homogeneous. Increasing model capacity is a natural extension given the diversity and complexity of scene images in our dataset. We find that increasing the channel width of convolutional layers by~$2 \times$ significantly improves our model (see ablation in Table~\ref{tab:stylegan2_a}). Following prior work in scaling GANs~\cite{brock2018large}, we also increase minibatch size (from $40$ to $120$). Concurrent work~\cite{mokady2022self,sauer2022stylegan} also explores scaling StyleGAN, and proposes strategies such as self-filtering the training dataset~\cite{mokady2022self}, progressive growing and leveraging pretrained classifiers~\cite{sauer2022stylegan}.

\subsection{Model details}
We train all models at $128 \times 128$ resolution with non-saturating logistic loss~\cite{goodfellow2014generative}, path length~\cite{karras2020analyzing} and $R_1$~\cite{Mescheder2018ICML} regularization, and exponential moving average of generator parameters~\cite{karras2018progressive}. We remove spatial noise maps from StyleGAN2 for all models, employ differentiable augmentation of both real and generated images~\cite{zhao2020differentiable,karras2020training}, and train the discriminator with an additional fake example containing real images with mismatched labels~\cite{reed2016generative}. See the supplement for details.

\section{Experiments}
\label{sec:results}

Our model hallucinates diverse, high quality images of scenes compatible with input pose. We generate scenes in isolation as well as scenes containing humans, and analyze our model through several visual experiments. Generating scene images is challenging due to the high complexity of data, and our model outperforms Pix2Pix/Pix2PixHD~\cite{isola2017image,wang2018pix2pixHD} and pose-conditioned StyleGAN2~\cite{karras2020analyzing} baselines in terms of image quality and accurate human placement. We present characteristic success and failure results in Figure~\ref{fig:success_results} and Figure~\ref{fig:failure_results}. See the supplement for more results, including multiple pages of random uncurated samples.

\begin{figure*}[p]
\centering

\begin{subfigure}[b]{1.0\linewidth}
\textoverlay{A}{\includegraphics[width=0.087\linewidth]{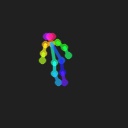}}%
\hfill%
\hfill%
\hfill%
\includegraphics[width=0.087\linewidth]{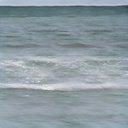}%
\hfill%
\includegraphics[width=0.087\linewidth]{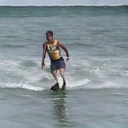}%
\hfill%
\hfill%
\hfill%
\includegraphics[width=0.087\linewidth]{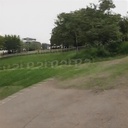}%
\hfill%
\includegraphics[width=0.087\linewidth]{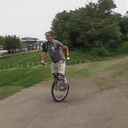}%
\hfill%
\hfill%
\hfill%
\includegraphics[width=0.087\linewidth]{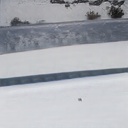}%
\hfill%
\includegraphics[width=0.087\linewidth]{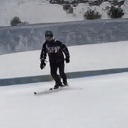}%
\hfill%
\hfill%
\hfill%
\includegraphics[width=0.087\linewidth]{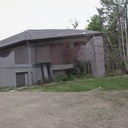}%
\hfill%
\includegraphics[width=0.087\linewidth]{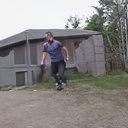}%
\hfill%
\hfill%
\hfill%
\includegraphics[width=0.087\linewidth]{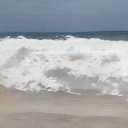}%
\hfill%
\includegraphics[width=0.087\linewidth]{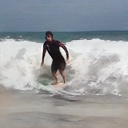}%
\end{subfigure}

% \vspace{0.9pt}

\begin{subfigure}[b]{1.0\linewidth}
\textoverlay{B}{\includegraphics[width=0.087\linewidth]{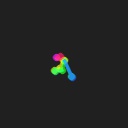}}%
\hfill%
\hfill%
\hfill%
\includegraphics[width=0.087\linewidth]{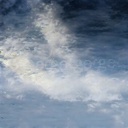}%
\hfill%
\includegraphics[width=0.087\linewidth]{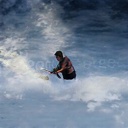}%
\hfill%
\hfill%
\hfill%
\includegraphics[width=0.087\linewidth]{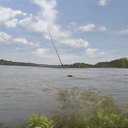}%
\hfill%
\includegraphics[width=0.087\linewidth]{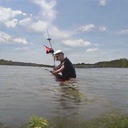}%
\hfill%
\hfill%
\hfill%
\includegraphics[width=0.087\linewidth]{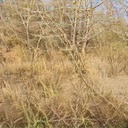}%
\hfill%
\includegraphics[width=0.087\linewidth]{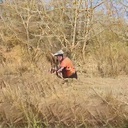}%
\hfill%
\hfill%
\hfill%
\includegraphics[width=0.087\linewidth]{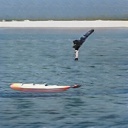}%
\hfill%
\includegraphics[width=0.087\linewidth]{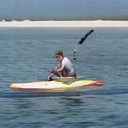}%
\hfill%
\hfill%
\hfill%
\includegraphics[width=0.087\linewidth]{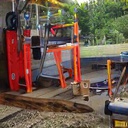}%
\hfill%
\includegraphics[width=0.087\linewidth]{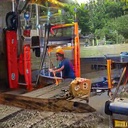}%
\end{subfigure}

% \vspace{0.9pt}

\begin{subfigure}[b]{1.0\linewidth}
\textoverlay{C}{\includegraphics[width=0.087\linewidth]{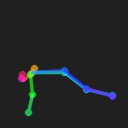}}%
\hfill%
\hfill%
\hfill%
\includegraphics[width=0.087\linewidth]{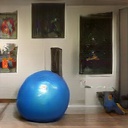}%
\hfill%
\includegraphics[width=0.087\linewidth]{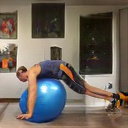}%
\hfill%
\hfill%
\hfill%
\includegraphics[width=0.087\linewidth]{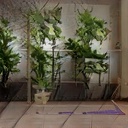}%
\hfill%
\includegraphics[width=0.087\linewidth]{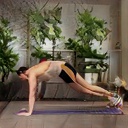}%
\hfill%
\hfill%
\hfill%
\includegraphics[width=0.087\linewidth]{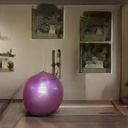}%
\hfill%
\includegraphics[width=0.087\linewidth]{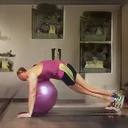}%
\hfill%
\hfill%
\hfill%
\includegraphics[width=0.087\linewidth]{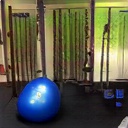}%
\hfill%
\includegraphics[width=0.087\linewidth]{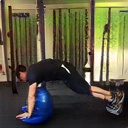}%
\hfill%
\hfill%
\hfill%
\includegraphics[width=0.087\linewidth]{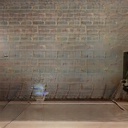}%
\hfill%
\includegraphics[width=0.087\linewidth]{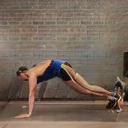}%
\end{subfigure}

% \vspace{0.9pt}

\begin{subfigure}[b]{1.0\linewidth}
\textoverlay{D}{\includegraphics[width=0.087\linewidth]{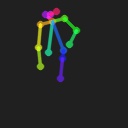}}%
\hfill%
\hfill%
\hfill%
\includegraphics[width=0.087\linewidth]{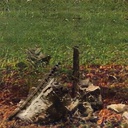}%
\hfill%
\includegraphics[width=0.087\linewidth]{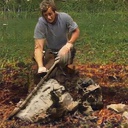}%
\hfill%
\hfill%
\hfill%
\includegraphics[width=0.087\linewidth]{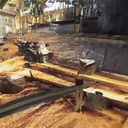}%
\hfill%
\includegraphics[width=0.087\linewidth]{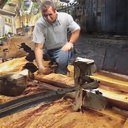}%
\hfill%
\hfill%
\hfill%
\includegraphics[width=0.087\linewidth]{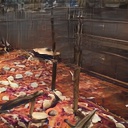}%
\hfill%
\includegraphics[width=0.087\linewidth]{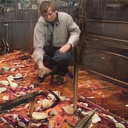}%
\hfill%
\hfill%
\hfill%
\includegraphics[width=0.087\linewidth]{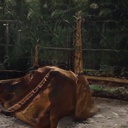}%
\hfill%
\includegraphics[width=0.087\linewidth]{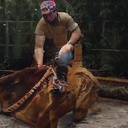}%
\hfill%
\hfill%
\hfill%
\includegraphics[width=0.087\linewidth]{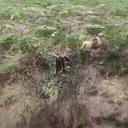}%
\hfill%
\includegraphics[width=0.087\linewidth]{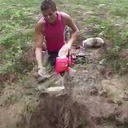}%
\end{subfigure}

% \vspace{0.9pt}

\begin{subfigure}[b]{1.0\linewidth}
\textoverlay{E}{\includegraphics[width=0.087\linewidth]{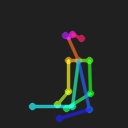}}%
\hfill%
\hfill%
\hfill%
\includegraphics[width=0.087\linewidth]{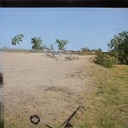}%
\hfill%
\includegraphics[width=0.087\linewidth]{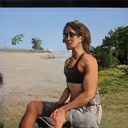}%
\hfill%
\hfill%
\hfill%
\includegraphics[width=0.087\linewidth]{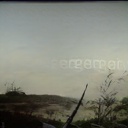}%
\hfill%
\includegraphics[width=0.087\linewidth]{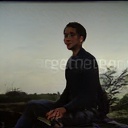}%
\hfill%
\hfill%
\hfill%
\includegraphics[width=0.087\linewidth]{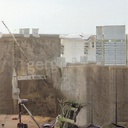}%
\hfill%
\includegraphics[width=0.087\linewidth]{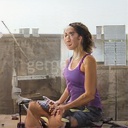}%
\hfill%
\hfill%
\hfill%
\includegraphics[width=0.087\linewidth]{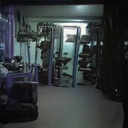}%
\hfill%
\includegraphics[width=0.087\linewidth]{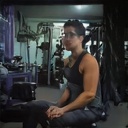}%
\hfill%
\hfill%
\hfill%
\includegraphics[width=0.087\linewidth]{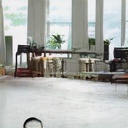}%
\hfill%
\includegraphics[width=0.087\linewidth]{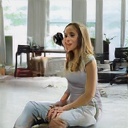}%
\end{subfigure}

% \vspace{0.9pt}

\begin{subfigure}[b]{1.0\linewidth}
\textoverlay{F}{\includegraphics[width=0.087\linewidth]{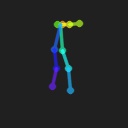}}%
\hfill%
\hfill%
\hfill%
\includegraphics[width=0.087\linewidth]{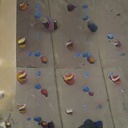}%
\hfill%
\includegraphics[width=0.087\linewidth]{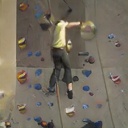}%
\hfill%
\hfill%
\hfill%
\includegraphics[width=0.087\linewidth]{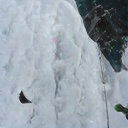}%
\hfill%
\includegraphics[width=0.087\linewidth]{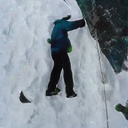}%
\hfill%
\hfill%
\hfill%
\includegraphics[width=0.087\linewidth]{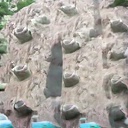}%
\hfill%
\includegraphics[width=0.087\linewidth]{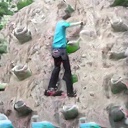}%
\hfill%
\hfill%
\hfill%
\includegraphics[width=0.087\linewidth]{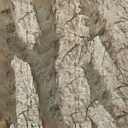}%
\hfill%
\includegraphics[width=0.087\linewidth]{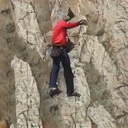}%
\hfill%
\hfill%
\hfill%
\includegraphics[width=0.087\linewidth]{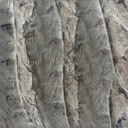}%
\hfill%
\includegraphics[width=0.087\linewidth]{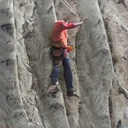}%
\end{subfigure}

% \vspace{0.9pt}

\begin{subfigure}[b]{1.0\linewidth}
\textoverlay{G}{\includegraphics[width=0.087\linewidth]{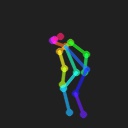}}%
\hfill%
\hfill%
\hfill%
\includegraphics[width=0.087\linewidth]{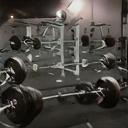}%
\hfill%
\includegraphics[width=0.087\linewidth]{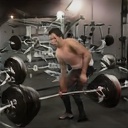}%
\hfill%
\hfill%
\hfill%
\includegraphics[width=0.087\linewidth]{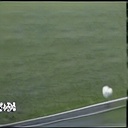}%
\hfill%
\includegraphics[width=0.087\linewidth]{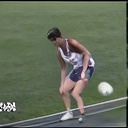}%
\hfill%
\hfill%
\hfill%
\includegraphics[width=0.087\linewidth]{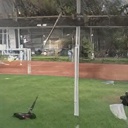}%
\hfill%
\includegraphics[width=0.087\linewidth]{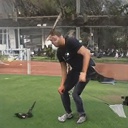}%
\hfill%
\hfill%
\hfill%
\includegraphics[width=0.087\linewidth]{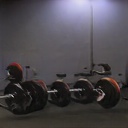}%
\hfill%
\includegraphics[width=0.087\linewidth]{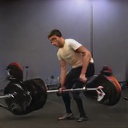}%
\hfill%
\hfill%
\hfill%
\includegraphics[width=0.087\linewidth]{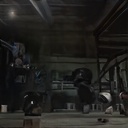}%
\hfill%
\includegraphics[width=0.087\linewidth]{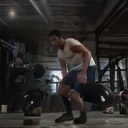}%
\end{subfigure}

% \vspace{0.9pt}

\begin{subfigure}[b]{1.0\linewidth}
\textoverlay{H}{\includegraphics[width=0.087\linewidth]{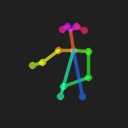}}%
\hfill%
\hfill%
\hfill%
\includegraphics[width=0.087\linewidth]{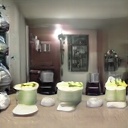}%
\hfill%
\includegraphics[width=0.087\linewidth]{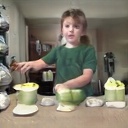}%
\hfill%
\hfill%
\hfill%
\includegraphics[width=0.087\linewidth]{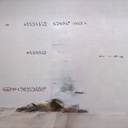}%
\hfill%
\includegraphics[width=0.087\linewidth]{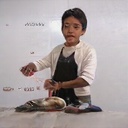}%
\hfill%
\hfill%
\hfill%
\includegraphics[width=0.087\linewidth]{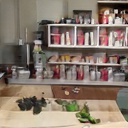}%
\hfill%
\includegraphics[width=0.087\linewidth]{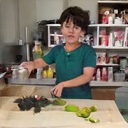}%
\hfill%
\hfill%
\hfill%
\includegraphics[width=0.087\linewidth]{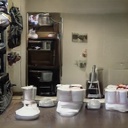}%
\hfill%
\includegraphics[width=0.087\linewidth]{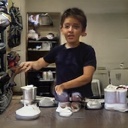}%
\hfill%
\hfill%
\hfill%
\includegraphics[width=0.087\linewidth]{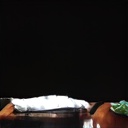}%
\hfill%
\includegraphics[width=0.087\linewidth]{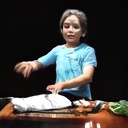}%
\end{subfigure}

% \vspace{0.9pt}

\begin{subfigure}[b]{1.0\linewidth}
\textoverlay{I}{\includegraphics[width=0.087\linewidth]{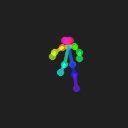}}%
\hfill%
\hfill%
\hfill%
\includegraphics[width=0.087\linewidth]{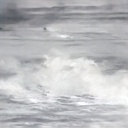}%
\hfill%
\includegraphics[width=0.087\linewidth]{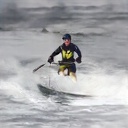}%
\hfill%
\hfill%
\hfill%
\includegraphics[width=0.087\linewidth]{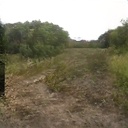}%
\hfill%
\includegraphics[width=0.087\linewidth]{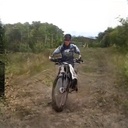}%
\hfill%
\hfill%
\hfill%
\includegraphics[width=0.087\linewidth]{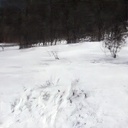}%
\hfill%
\includegraphics[width=0.087\linewidth]{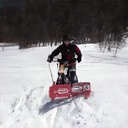}%
\hfill%
\hfill%
\hfill%
\includegraphics[width=0.087\linewidth]{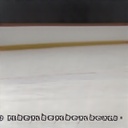}%
\hfill%
\includegraphics[width=0.087\linewidth]{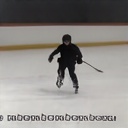}%
\hfill%
\hfill%
\hfill%
\includegraphics[width=0.087\linewidth]{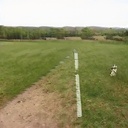}%
\hfill%
\includegraphics[width=0.087\linewidth]{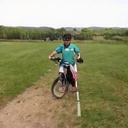}%
\end{subfigure}

% \vspace{0.9pt}

\begin{subfigure}[b]{1.0\linewidth}
\textoverlay{J}{\includegraphics[width=0.087\linewidth]{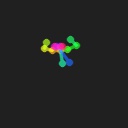}}%
\hfill%
\hfill%
\hfill%
\includegraphics[width=0.087\linewidth]{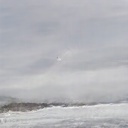}%
\hfill%
\includegraphics[width=0.087\linewidth]{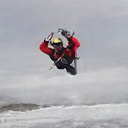}%
\hfill%
\hfill%
\hfill%
\includegraphics[width=0.087\linewidth]{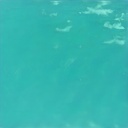}%
\hfill%
\includegraphics[width=0.087\linewidth]{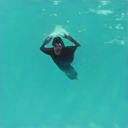}%
\hfill%
\hfill%
\hfill%
\includegraphics[width=0.087\linewidth]{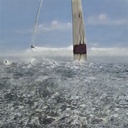}%
\hfill%
\includegraphics[width=0.087\linewidth]{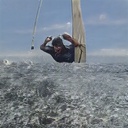}%
\hfill%
\hfill%
\hfill%
\includegraphics[width=0.087\linewidth]{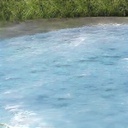}%
\hfill%
\includegraphics[width=0.087\linewidth]{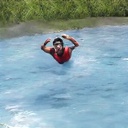}%
\hfill%
\hfill%
\hfill%
\includegraphics[width=0.087\linewidth]{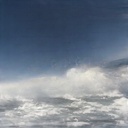}%
\hfill%
\includegraphics[width=0.087\linewidth]{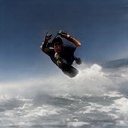}%
\end{subfigure}

% \vspace{0.9pt}

\begin{subfigure}[b]{1.0\linewidth}
\textoverlay{K}{\includegraphics[width=0.087\linewidth]{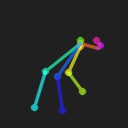}}%
\hfill%
\hfill%
\hfill%
\includegraphics[width=0.087\linewidth]{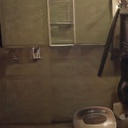}%
\hfill%
\includegraphics[width=0.087\linewidth]{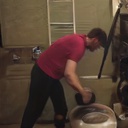}%
\hfill%
\hfill%
\hfill%
\includegraphics[width=0.087\linewidth]{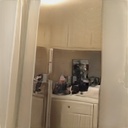}%
\hfill%
\includegraphics[width=0.087\linewidth]{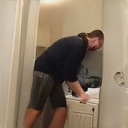}%
\hfill%
\hfill%
\hfill%
\includegraphics[width=0.087\linewidth]{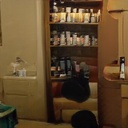}%
\hfill%
\includegraphics[width=0.087\linewidth]{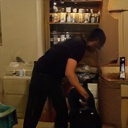}%
\hfill%
\hfill%
\hfill%
\includegraphics[width=0.087\linewidth]{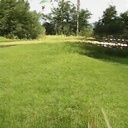}%
\hfill%
\includegraphics[width=0.087\linewidth]{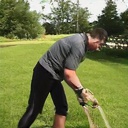}%
\hfill%
\hfill%
\hfill%
\includegraphics[width=0.087\linewidth]{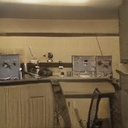}%
\hfill%
\includegraphics[width=0.087\linewidth]{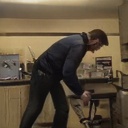}%
\end{subfigure}

% \vspace{0.9pt}

% \begin{subfigure}[b]{1.0\linewidth}
% \textoverlay{L}{\includegraphics[width=0.087\linewidth]{assets/success_results/213231_pose.jpg}}%
% \hfill%
% \hfill%
% \hfill%
% \includegraphics[width=0.087\linewidth]{assets/success_results/213231_213285_scene.jpg}%
% \hfill%
% \includegraphics[width=0.087\linewidth]{assets/success_results/213231_213285_image.jpg}%
% \hfill%
% \hfill%
% \hfill%
% \includegraphics[width=0.087\linewidth]{assets/success_results/213231_213244_scene.jpg}%
% \hfill%
% \includegraphics[width=0.087\linewidth]{assets/success_results/213231_213244_image.jpg}%
% \hfill%
% \hfill%
% \hfill%
% \includegraphics[width=0.087\linewidth]{assets/success_results/213231_213292_scene.jpg}%
% \hfill%
% \includegraphics[width=0.087\linewidth]{assets/success_results/213231_213292_image.jpg}%
% \hfill%
% \hfill%
% \hfill%
% \includegraphics[width=0.087\linewidth]{assets/success_results/213231_213266_scene.jpg}%
% \hfill%
% \includegraphics[width=0.087\linewidth]{assets/success_results/213231_213266_image.jpg}%
% \hfill%
% \hfill%
% \hfill%
% \includegraphics[width=0.087\linewidth]{assets/success_results/213231_213311_scene.jpg}%
% \hfill%
% \includegraphics[width=0.087\linewidth]{assets/success_results/213231_213311_image.jpg}%
% \end{subfigure}

% \vspace{0.9pt}

\begin{subfigure}[b]{1.0\linewidth}
\textoverlay{L}{\includegraphics[width=0.087\linewidth]{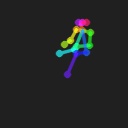}}%
\hfill%
\hfill%
\hfill%
\includegraphics[width=0.087\linewidth]{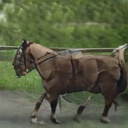}%
\hfill%
\includegraphics[width=0.087\linewidth]{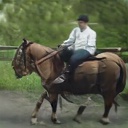}%
\hfill%
\hfill%
\hfill%
\includegraphics[width=0.087\linewidth]{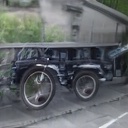}%
\hfill%
\includegraphics[width=0.087\linewidth]{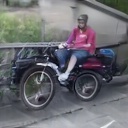}%
\hfill%
\hfill%
\hfill%
\includegraphics[width=0.087\linewidth]{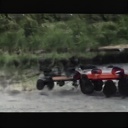}%
\hfill%
\includegraphics[width=0.087\linewidth]{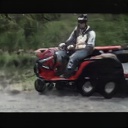}%
\hfill%
\hfill%
\hfill%
\includegraphics[width=0.087\linewidth]{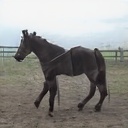}%
\hfill%
\includegraphics[width=0.087\linewidth]{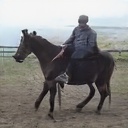}%
\hfill%
\hfill%
\hfill%
\includegraphics[width=0.087\linewidth]{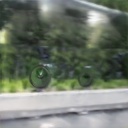}%
\hfill%
\includegraphics[width=0.087\linewidth]{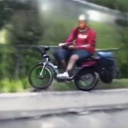}%
\end{subfigure}

% \vspace{0.9pt}

\begin{subfigure}[b]{1.0\linewidth}
\textoverlay{M}{\includegraphics[width=0.087\linewidth]{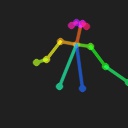}}%
\hfill%
\hfill%
\hfill%
\includegraphics[width=0.087\linewidth]{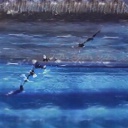}%
\hfill%
\includegraphics[width=0.087\linewidth]{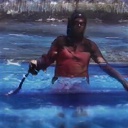}%
\hfill%
\hfill%
\hfill%
\includegraphics[width=0.087\linewidth]{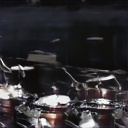}%
\hfill%
\includegraphics[width=0.087\linewidth]{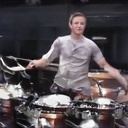}%
\hfill%
\hfill%
\hfill%
\includegraphics[width=0.087\linewidth]{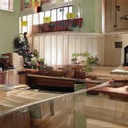}%
\hfill%
\includegraphics[width=0.087\linewidth]{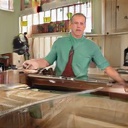}%
\hfill%
\hfill%
\hfill%
\includegraphics[width=0.087\linewidth]{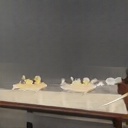}%
\hfill%
\includegraphics[width=0.087\linewidth]{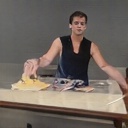}%
\hfill%
\hfill%
\hfill%
\includegraphics[width=0.087\linewidth]{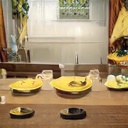}%
\hfill%
\includegraphics[width=0.087\linewidth]{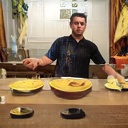}%
\end{subfigure}

% \vspace{0.9pt}

\begin{subfigure}[b]{1.0\linewidth}
\textoverlay{N}{\includegraphics[width=0.087\linewidth]{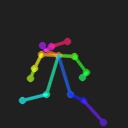}}%
\hfill%
\hfill%
\hfill%
\includegraphics[width=0.087\linewidth]{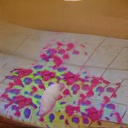}%
\hfill%
\includegraphics[width=0.087\linewidth]{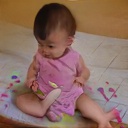}%
\hfill%
\hfill%
\hfill%
\includegraphics[width=0.087\linewidth]{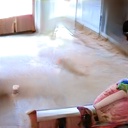}%
\hfill%
\includegraphics[width=0.087\linewidth]{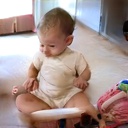}%
\hfill%
\hfill%
\hfill%
\includegraphics[width=0.087\linewidth]{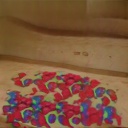}%
\hfill%
\includegraphics[width=0.087\linewidth]{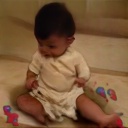}%
\hfill%
\hfill%
\hfill%
\includegraphics[width=0.087\linewidth]{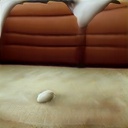}%
\hfill%
\includegraphics[width=0.087\linewidth]{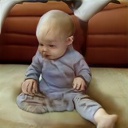}%
\hfill%
\hfill%
\hfill%
\includegraphics[width=0.087\linewidth]{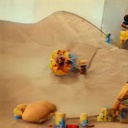}%
\hfill%
\includegraphics[width=0.087\linewidth]{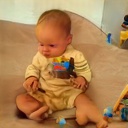}%
\end{subfigure}

% \vspace{-5pt}

\caption{\textbf{Success cases.} Our model learns complex scene-pose relationships. For each input pose, we show many hallucinated scenes, with and without a human. Diverse outputs include a person paddling a kayak (B), lifting a barbell in their hand (G), cleaning the toilet (K), and playing the drums (M). Our model produces multiple plausible scenes for the same pose, providing insight into scenes with related affordances: in the same pose, a person may climb in an indoor gym or on a snowy ledge (F); a person can ride a horse, ride a bicycle, or ride a tractor (L). Please see the appendix for multiple pages of random results.
% {\em See example K for the first ever GAN-generated image of a person cleaning a toilet.}
}
\label{fig:success_results}

\end{figure*}

\begin{figure*}[h!]
\centering

\begin{subfigure}[b]{1.0\linewidth}
\begin{subfigure}[b]{0.24\linewidth}
\textoverlay{A}{\includegraphics[width=0.325\linewidth]{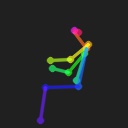}}%
\hfill%
\includegraphics[width=0.325\linewidth]{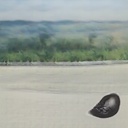}%
\hfill%
\includegraphics[width=0.325\linewidth]{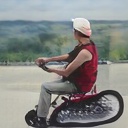}%
\end{subfigure}
\hfill%
\begin{subfigure}[b]{0.24\linewidth}
\textoverlay{B}{\includegraphics[width=0.325\linewidth]{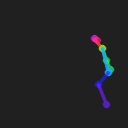}}%
\hfill%
\includegraphics[width=0.325\linewidth]{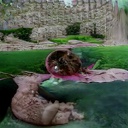}%
\hfill%
\includegraphics[width=0.325\linewidth]{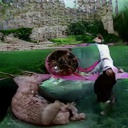}%
\end{subfigure}
\hfill%
\begin{subfigure}[b]{0.24\linewidth}
\textoverlay{C}{\includegraphics[width=0.325\linewidth]{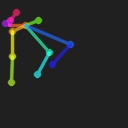}}%
\hfill%
\includegraphics[width=0.325\linewidth]{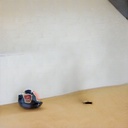}%
\hfill%
\includegraphics[width=0.325\linewidth]{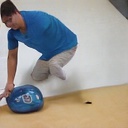}%
\end{subfigure}
\hfill%
\begin{subfigure}[b]{0.24\linewidth}
\textoverlay{D}{\includegraphics[width=0.325\linewidth]{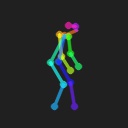}}%
\hfill%
\includegraphics[width=0.325\linewidth]{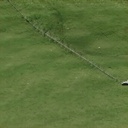}%
\hfill%
\includegraphics[width=0.325\linewidth]{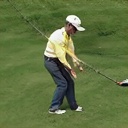}%
\end{subfigure}

% \vspace{0.9pt}

\begin{subfigure}[b]{0.24\linewidth}
\textoverlay{E}{\includegraphics[width=0.325\linewidth]{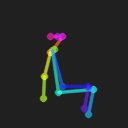}}%
\hfill%
\includegraphics[width=0.325\linewidth]{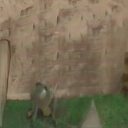}%
\hfill%
\includegraphics[width=0.325\linewidth]{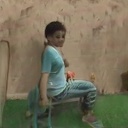}%
\end{subfigure}
\hfill%
\begin{subfigure}[b]{0.24\linewidth}
\textoverlay{F}{\includegraphics[width=0.325\linewidth]{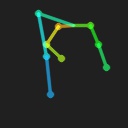}}%
\hfill%
\includegraphics[width=0.325\linewidth]{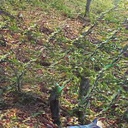}%
\hfill%
\includegraphics[width=0.325\linewidth]{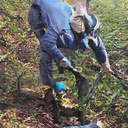}%
\end{subfigure}
\hfill%
\begin{subfigure}[b]{0.24\linewidth}
\textoverlay{G}{\includegraphics[width=0.325\linewidth]{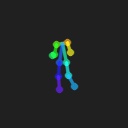}}%
\hfill%
\includegraphics[width=0.325\linewidth]{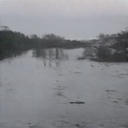}%
\hfill%
\includegraphics[width=0.325\linewidth]{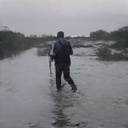}%
\end{subfigure}
\hfill%
\begin{subfigure}[b]{0.24\linewidth}
\textoverlay{H}{\includegraphics[width=0.325\linewidth]{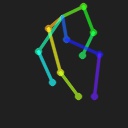}}%
\hfill%
\includegraphics[width=0.325\linewidth]{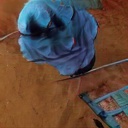}%
\hfill%
\includegraphics[width=0.325\linewidth]{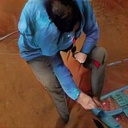}%
\end{subfigure}
\end{subfigure}

% \vspace{-5pt}

\caption{\textbf{Failure cases.} Causes for failure include: partially generating objects, such as a bike (A); poor overall image quality (B); missing limbs without proper occluders (C); difficulty placing objects, such as a golf club, in a person's hands (D); difficulty hallucinating an object on which to sit (E); overly repetitive textures (F); infeasible scenes, such as walking on water (G); and leaving behind a partial human when hallucinating the scene in isolation (H).}
\label{fig:failure_results}

% \vspace{-10pt}

\end{figure*}

\begin{figure}[h!]
    \centering

    \includegraphics[width=0.84\linewidth]{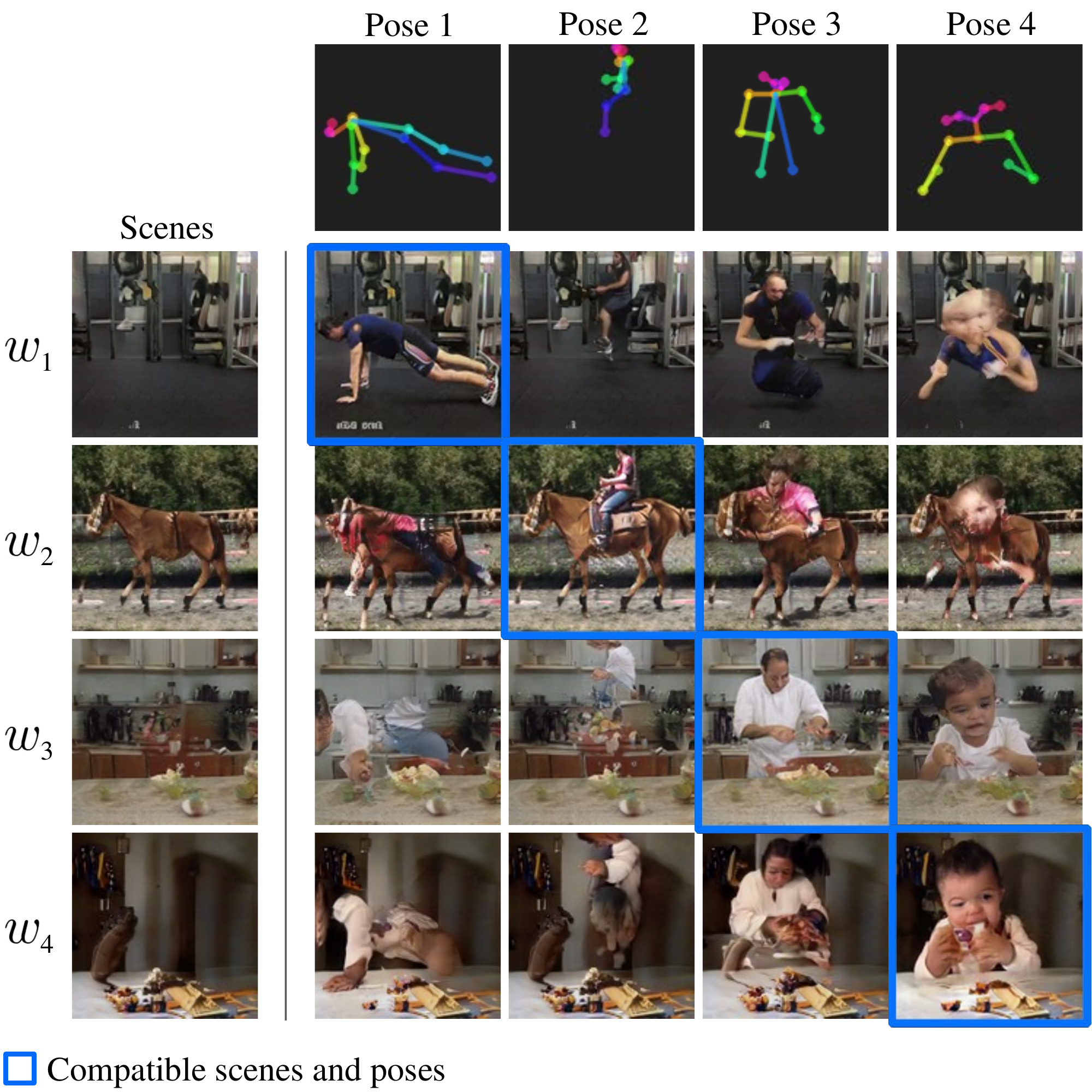}
    \par\smallskip
    \caption{A central theme of our paper is that scenes must be compatible with human poses to produce realistic images --- here we visualize what happens when scenes and poses are {\em not} compatible. Correctly paired images are shown in blue on the diagonal --- a person doing a pushup in a gym, riding a horse, cooking in a kitchen, and a baby leaning on a table. These exemplify interesting relationships between human pose and scene learned by our model. Other images mix scene latent codes with keypoint heatmaps from the wrong pose, often producing unrealistic images. Generating pose-compatible scenes is essential to avoid these incorrect pairings.}

    \label{fig:compat}
\end{figure}

\subsection{Not all scenes and poses are compatible}
\label{sec:compat}

It is essential that we model which scenes are compatible with which poses. A person cannot do a push-up in the middle of a horse, ride atop a kitchen countertop, or be occluded by thin air. These scenarios sound obviously false, yet could occur if the scene and human pose are incompatible. We visualize images generated with correctly and incorrectly paired scenes and poses in Figure~\ref{fig:compat}.

These examples of incompatible scenes and poses highlight an important difference between our scene data and other datasets commonly used for GAN training, such as cropped faces in the CelebA~\cite{liu2015faceattributes} and FFHQ~\cite{karras2019style} datasets. Any face can be given glasses, longer or shorter hair, or a darker or lighter skin tone and still remain a feasible image. This enables global disentanglement of attributes, and applications like style mixing, which combines different intermediate latent codes of any two samples (see Figure~3 of the original StyleGAN paper~\cite{karras2019style} for a wonderful example). The assumption of compatibility between all attribute pairs no longer holds for data of scenes with humans, which motivates conditioning scene latent codes on pose. Relatedly, we find that removing style mixing from training significantly improves performance (Table~\ref{tab:stylegan2_a}).

\subsection{Scene occlusion reasoning}

\begin{figure}[h]
    \centering

    \begin{subfigure}[b]{0.85\linewidth}
        \includegraphics[width=0.193\linewidth]{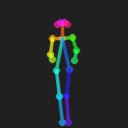}%
        \hfill%
        \hfill%
        \includegraphics[width=0.193\linewidth]{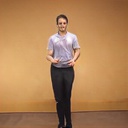}%
        \hfill%
        \includegraphics[width=0.193\linewidth]{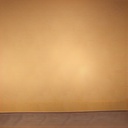}%
        \hfill%
        \hfill%
        \includegraphics[width=0.193\linewidth]{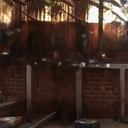}%
        \hfill%
        \includegraphics[width=0.193\linewidth]{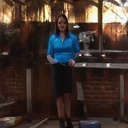}%
        \caption{Scenes generated with a full-body input pose.}
        \label{fig:occ_fb}
    \end{subfigure}
    % \hfill
    \par\smallskip
    \begin{subfigure}[b]{0.85\linewidth}
        \includegraphics[width=0.193\linewidth]{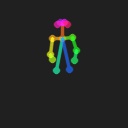}%
        \hfill%
        \hfill%
        \includegraphics[width=0.193\linewidth]{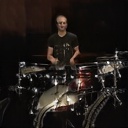}%
        \hfill%
        \includegraphics[width=0.193\linewidth]{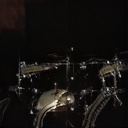}%
        \hfill%
        \hfill%
        \includegraphics[width=0.193\linewidth]{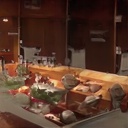}%
        \hfill%
        \includegraphics[width=0.193\linewidth]{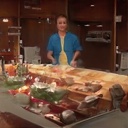}%
        \caption{Scenes generated from the same input pose with legs not visible.}
        \label{fig:occ_occ}
    \end{subfigure}
    \par\smallskip
    \caption{(a) A full-body input pose and corresponding scenes. (b) When the legs from an otherwise identical pose are hidden, our model hallucinates scenes with foreground objects, such as a drum kit or table, to occlude the missing legs.}
    \label{fig:occlusions}

\end{figure}

Portions of a human pose may be occluded by foreground objects, such as pieces of furniture. Provided a partially visible human pose, our model hallucinates scenes with foreground objects to occlude portions of the pose not visible. Figure~\ref{fig:occlusions}a shows an example full-body pose and output scenes. When the legs are not visible in the input pose in \ref{fig:occlusions}b, our model produces scenes with occluders blocking the legs, demonstrating its emergent ability to reason about occlusions.

\begin{figure}[t]
    \centering
    \includegraphics[width=0.75\linewidth]{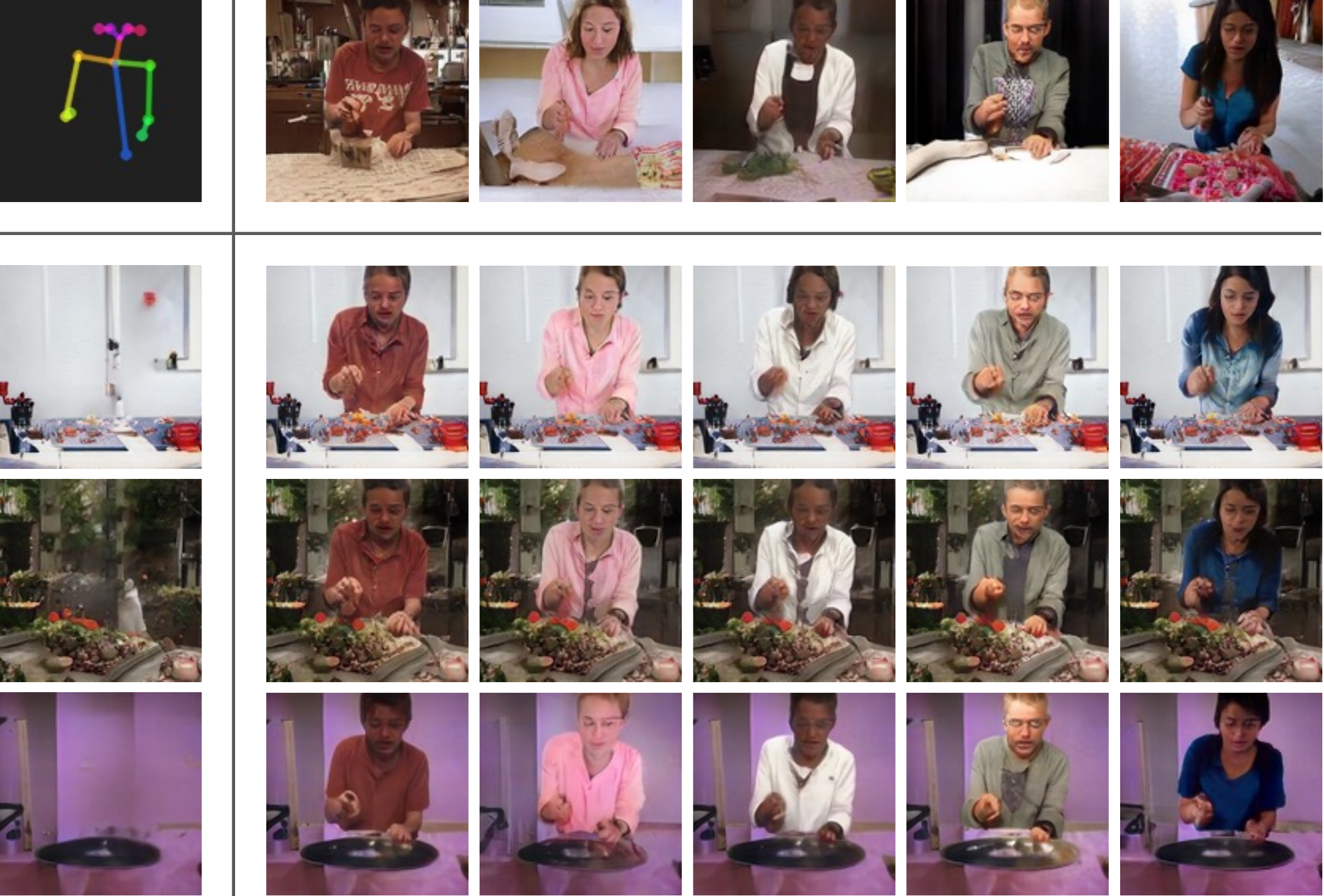}

    \caption{Given an input pose (top left), our method can compose human appearances (top row) and scenes (left column) from different generated images.}
    \label{fig:disentanglement}

\end{figure}

\begin{figure}[h]
    \centering

    \begin{subfigure}[b]{0.81\linewidth}
        \includegraphics[width=\linewidth]{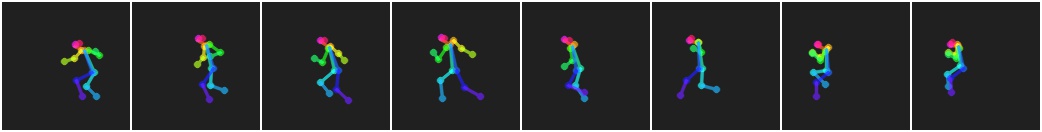}
        \includegraphics[width=\linewidth]{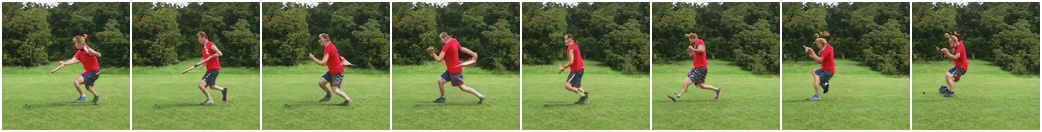}
        \includegraphics[width=\linewidth]{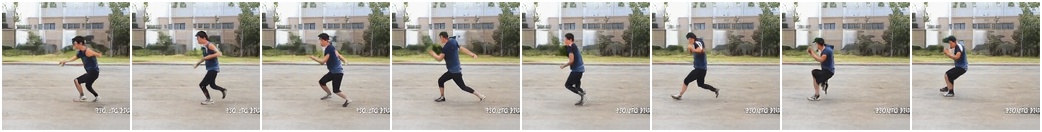}%
    \end{subfigure}

    \caption{Provided an input pose sequence (top), we infer scenes based on the first pose, then generate animations (middle/bottom) by keeping the scene latent fixed and passing keypoint heatmaps for each subsequent pose.}
    \label{fig:vid2vid}

\end{figure}

\subsection{Human appearance and scene disentanglement}

Section~\ref{sec:compat} demonstrates why complete separation of pose and scene is undesirable. We can, however, disentangle human appearance from scene when both are conditioned on the same pose, as shown in Figure~\ref{fig:disentanglement}. To achieve this, we optimize a latent code to compose two samples.
We minimize perceptual loss~\cite{johnson2016perceptual,zhang2018perceptual} between person-only crops of the composition and first sample, %obtained using the bounding box surrounding human pose; we also minimize perceptual loss between
and scene-only images of the composition and second sample.
See the supplement for details.

\subsection{Animating pose}

After training, our model is capable of animating pose in a stationary scene. In Figure~\ref{fig:vid2vid} we demonstrate a sequence of images generated by fixing the scene and animating the human pose. The scene is inferred from only the first pose, and is limited to small human motion and stationary backgrounds.

\subsection{Scene clustering and truncation}

\begin{figure}[t]
    \centering
    \includegraphics[width=.9\linewidth]{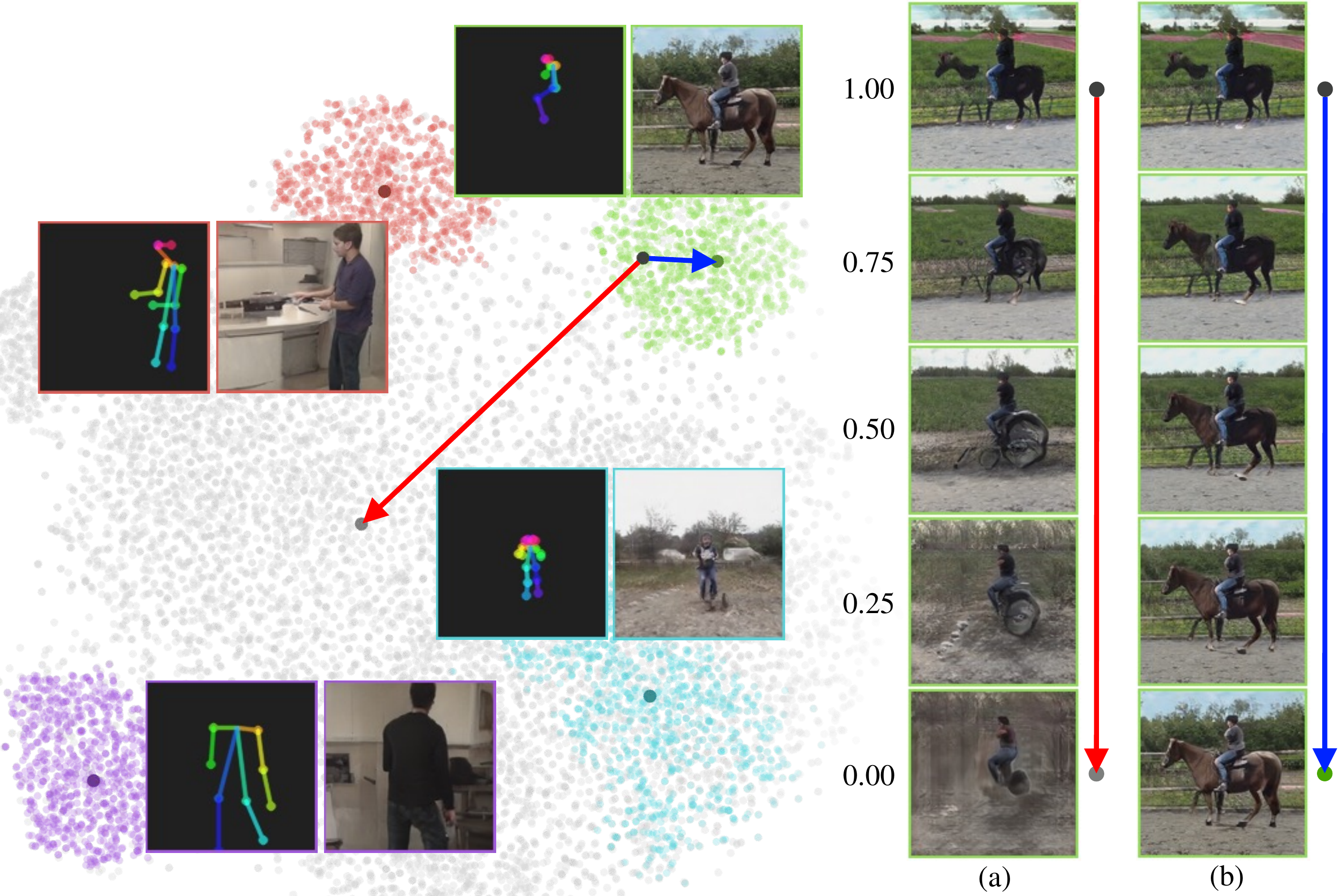}
    \par\medskip
    % \vspace{-2pt}
    \caption{We contrast truncation via (a) interpolation toward the mean of random latents, and (b) interpolation toward the mean of conditional latent clusters. The left plot shows a t-SNE~\cite{van2008visualizing} visualization of latent codes. Gray points are \num{10000} random latents. Colored sets of points are each \num{1000} latent samples conditioned on the same pose. The formation of clusters signifies that different scene latents conditioned on the same pose are close to each other in the intermediate latent space. The dark gray point in the center is the mean of all random latents, and dark colored points are the means for each pose. Beside each cluster is the input pose and image generated using the corresponding mean cluster latent. Conditional truncation (b) works significantly better than unconditional (a).}
    \label{fig:truncation}
    % \vspace{-2pt}
\end{figure}

Regions of low density in the data distribution are particularly challenging to model. Quality can be improved (at the loss of some diversity) by sampling from a shrunk distribution~\cite{ackley1985learning,marchesi2017megapixel,glow2018,brock2018large,fan2018hierarchical,holtzman2019curious}. StyleGAN~\cite{karras2019style} interpolates intermediate latents $w$ toward the mean $\bar w =~\mathbb{E}_{z \sim \mathcal{Z}} [w]$ to shrink the sampling distribution, which improves generation quality for models trained on data such as faces. However, on our more complex data, interpolating toward the mean scene latent produces a gray scene rather than improving quality, as shown in Figure~\ref{fig:truncation}a.

In visualizing a t-SNE~\cite{van2008visualizing} plot of scene latents in Figure~\ref{fig:truncation}, we observe that latents sampled from different noise vectors $z$ yet conditioned on the same pose $p$ form clusters. We apply conditional truncation by interpolating a latent $w$ toward the conditional mean $\bar w_p =~\mathbb{E}_{z \sim \mathcal{Z}} [w | p]$, shifting the sample toward the cluster center. Shown in part (b) of Figure~\ref{fig:truncation}, conditional truncation works significantly better for our model. We apply conditional truncation $w^\prime = \bar w_p + \psi (w - \bar w_p)$ of $\psi = 0.75$ to generated images throughout the paper. Concurrent work~\cite{mokady2022self} proposes a similar method for applying truncation toward the centers of perceptual clusters.

\subsection{Baseline comparisons}
\label{sec:metrics}

\begin{figure}[t]
    \centering
    \captionsetup[subfigure]{labelformat=empty}
    \begin{subfigure}{0.71\linewidth}
    \includegraphics[width=0.1835\linewidth]{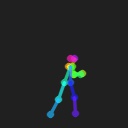}%
    \hfill
    \includegraphics[width=0.1835\linewidth]{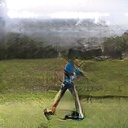}%
    \hfill
    \includegraphics[width=0.1835\linewidth]{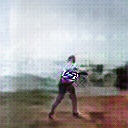}%
    \hfill
    \includegraphics[width=0.1835\linewidth]{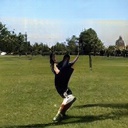}%
    \hfill
    \includegraphics[width=0.1835\linewidth]{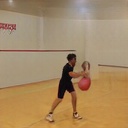}%

    \includegraphics[width=0.1835\linewidth]{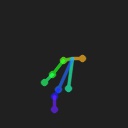}%
    \hfill
    \includegraphics[width=0.1835\linewidth]{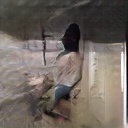}%
    \hfill
    \includegraphics[width=0.1835\linewidth]{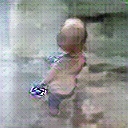}%
    \hfill
    \includegraphics[width=0.1835\linewidth]{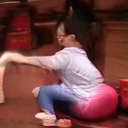}%
    \hfill
    \includegraphics[width=0.1835\linewidth]{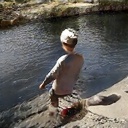}%

    \includegraphics[width=0.1835\linewidth]{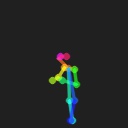}%
    \hfill
    \includegraphics[width=0.1835\linewidth]{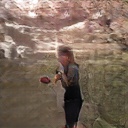}%
    \hfill
    \includegraphics[width=0.1835\linewidth]{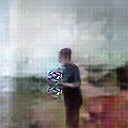}%
    \hfill
    \includegraphics[width=0.1835\linewidth]{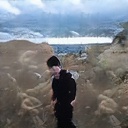}%
    \hfill
    \includegraphics[width=0.1835\linewidth]{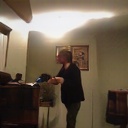}%
    \par\smallskip
    \begin{subfigure}{0.18\linewidth}
        \caption{Pose}
    \end{subfigure}
    \begin{subfigure}{0.195\linewidth}
        \caption{Pix2Pix}
    \end{subfigure}
    \begin{subfigure}{0.195\linewidth}
        \caption{Pix2PixHD}
    \end{subfigure}
    \begin{subfigure}{0.205\linewidth}
        \caption{SG2}
    \end{subfigure}
    \begin{subfigure}{0.165\linewidth}
        \caption{Ours}
    \end{subfigure}%
    \end{subfigure}%
    \caption{Baseline comparisons. Pix2Pix/Pix2PixHD struggle to produce realistic images; pose-conditioned StyleGAN2 (SG2) often generates humans in the wrong pose; our model generates realistic scenes with humans in the correct pose.}
    \label{fig:baselines}
\end{figure}

Please see Figure~\ref{fig:baselines} for visual comparisons with baseline methods. Pix2Pix and Pix2PixHD were designed for image translation tasks with stronger conditioning, such as segmentation masks. These methods struggle to produce reasonable images on our more challenging task and dataset. StyleGAN2 (SG2) with latent pose conditioning provides a stronger baseline, but still has notable issues with image quality and often places humans in the incorrect pose. These observations are corroborated by metric performance in two respects: how accurately human subjects are positioned, and how realistic generated scenes look.

To succeed at our task, a model must both put a human in the correct pose and generate a compatible scene. Table~\ref{tab:main_short} compares our model with Pix2Pix, Pix2PixHD and StyleGAN2 baselines on these metrics, demonstrating that our model achieves superior performance. Note that StyleGAN2~\cite{karras2020analyzing} is primarily an unconditional GAN. The public code release and follow-up work~\cite{karras2020training} support class-conditional generation. We refer to the version of our model with only pose latent conditioning as StyleGAN2, since it is the most straightforward extension of StyleGAN2 for our task.

\paragraph{Accurate human positioning.} PCKh~\cite{andriluka20142d} measures the percent of correct pose keypoints (within a radius relative to the head size), where a higher percent is better. We use OpenPose~\cite{cao2019openpose} to extract poses from generated images for comparison with input poses. PCKh is computed on a held out test set, ensuring accurate placement of new poses not seen during training.

\paragraph{Realistic scene images.} FID --- Fréchet inception distance~\cite{fidscore} --- measures realism by comparing distributions of Inception network~\cite{szegedy2015going} features between the training dataset and generated images. Lower FID scores are better and correlate with higher quality, more realistic images.

\begin{table}[h!]
    \centering
    \caption{\textbf{Baseline metric comparisons.} We report PCKh (higher is better) as a measure of how accurately humans are positioned, and FID (lower is better) as a measure of how realistic generated scenes look. Our model outperforms Pix2Pix, Pix2PixHD and pose-conditioned StyleGAN2 baselines on both metrics. While the poor performance of baselines may appear surprising, note that our task is much more challenging than standard conditional generation tasks: the dataset is diverse and complex, and conditioning on pose requires the network to infer scene contents and layout.}
    \label{tab:main_short}
    \begin{tabular}{p{15em}SS}
        \toprule
                                                            & {\small PCKh $\uparrow$}  & {\small FID $\downarrow$} \\
        \midrule
        \small Pix2Pix                                      & 48.4                      & 71.2                  \\
        \small Pix2PixHD                                    & 73.8                     & 149.7                 \\
        \small StyleGAN2 \qquad\qquad\qquad\qquad\scriptsize (with pose latent conditioning)   & 32.4                      & 16.6                           \\
        \small Ours                         & 84.2            & 5.9                    \\
        \bottomrule
    \end{tabular}
\end{table}

\begin{table}[h!]
    \centering
    \caption{\textbf{StyleGAN2 ablation.} We enumerate modifications relative to a pose-conditioned StyleGAN2 baseline. In particular, removing style mixing, conditioning on keypoint heatmaps, augmenting discriminator inputs and passing a fake mismatched example to the discriminator, and increasing scale all contribute to our final model.}
    \label{tab:stylegan2_a}
    \begin{tabular}{p{15em}SS}
        \toprule
                                                            & {\small PCKh $\uparrow$}  & {\small FID $\downarrow$} \\
        \midrule
        \small StyleGAN2 \qquad\qquad\qquad\qquad\scriptsize (with pose latent conditioning)   & 32.4                      & 16.6                           \\
        \small -- style mixing                               & 36.4                      & 11.6                                 \\
        \small + keypoint heatmaps                          & 79.8                      & 12.2                                    \\
        \small + augmentation, mismatch                     & 80.7                      & 12.1                              \\
        \small + large scale (Ours)                         & 84.2            & 5.9                    \\
        \bottomrule
    \end{tabular}
\end{table}

\begin{table}[h!]
\begin{center}
\caption{\textbf{Pose conditioning ablation.} We contrast three options for pose conditioning: only conditioning the latent on pose, only conditioning on keypoint heatmaps, and dual conditioning of both latents and heatmaps. We conduct this ablation on the smaller version of our model. We find that keypoint heatmap conditioning is crucial for accurately placing a human (PCKh), whereas latent conditioning improves the quality of scene generation (FID). We condition with both mechanisms in our final model, which has the best metric trade-off, and enables separating control of human position and scene generation after training.}
\label{tab:conditioning}
\begin{tabular}{p{15em}SS}
\toprule
       {Conditioning method}    & {\small PCKh $\uparrow$}  & {\small FID $\downarrow$} \\
\midrule
\small Latent only                   & 36.4                      & 11.6          \\
\small Heatmap only                   & 79.7 &    15.1 \\
\small Both                          & 79.8                      & 12.2          \\
\bottomrule
\end{tabular}
\end{center}
\end{table}

\subsection{Ablations}
\label{sec:ablations}

We present two ablation experiments. Table~\ref{tab:stylegan2_a} enumerates changes relative to a pose-conditioned StyleGAN2 baseline, demonstrating improvements gained by our simple yet important modifications. Table~\ref{tab:conditioning} compares three options for pose conditioning: latents only, keypoint heatmaps only, and dual conditioning of both. Keypoint heatmaps are necessary to accurately position a human in the scene, which is shown by a substantially higher PCKh. Latent conditioning improves quality, which is shown by a lower FID score. We condition with both mechanisms --- in addition to offering the best trade-off in metric performance, dual conditioning enables applications of disentanglement, such as generating scenes without humans or visualizing incompatible scenes and poses.

\section{Discussion}
\label{sec:discussion}

\paragraph{Limitations.} Our dataset and model only consider images with a single human subject. Dataset curation is limited by the performance of Keypoint R-CNN~\cite{he2017mask,wu2019detectron2} and OpenPose~\cite{cao2017realtime,cao2019openpose} when filtering videos for humans. Training depends on OpenPose to correctly predict poses. Our model does not consider human movement when inferring scenes.

\paragraph{Societal impact.} There is some risk of this or future generative models being used to create fake and misleading content. Our model also inherits any demographic bias present in the existing datasets used to source our training data.

\paragraph{Conclusion.}
In this paper, we present a new task: provided a human pose as input, hallucinate the possible scene(s) which are compatible with that input pose. Strong relationships between humans, objects and environments dictate which scenes afford a given pose. Many prior works study human affordances from the angle of predicting possible poses given an input scene --- we study the other side of the same coin, and hallucinate scenes that afford an input pose.

We demonstrate the emergent ability of our model to capture affordance relationships between scenes and poses. This work marks a significant step toward using GANs to represent complex real-world environments. We hope it will motivate the broader research community to leverage modern generative approaches for scene understanding and modeling. %#

\paragraph{Acknowledgements.}
We thank William Peebles, Ilija Radosavovic, Matthew Tancik, Allan Jabri, Dave Epstein, Lucy Chai, Toru Lin, Shiry Ginosar, Angjoo Kanazawa, Vickie Ye, Karttikeya Mangalam, and Taesung Park for insightful discussion and feedback. Tim Brooks is supported by the National Science Foundation Graduate Research Fellowship under Grant No. 2020306087. Additional support for this project is provided by DARPA MCS, and SAP.

% \clearpage
\bibliographystyle{splncs04}
\bibliography{egbib}

\begin{thebibliography}{10}
\providecommand{\url}[1]{\texttt{#1}}
\providecommand{\urlprefix}{URL }
\providecommand{\doi}[1]{https://doi.org/#1}

\bibitem{im2style}
Abdal, R., Qin, Y., Wonka, P.: Image2stylegan: How to embed images into the
  stylegan latent space? In: ICCV. pp. 4431--4440 (2019),
  \url{https://doi.org/10.1109/ICCV.2019.00453}

\bibitem{aberman2019deep}
Aberman, K., Shi, M., Liao, J., Lischinski, D., Chen, B., Cohen-Or, D.: Deep
  video-based performance cloning. In: Computer Graphics Forum. vol.~38, pp.
  219--233. Wiley Online Library (2019)

\bibitem{ackley1985learning}
Ackley, D.H., Hinton, G.E., Sejnowski, T.J.: A learning algorithm for boltzmann
  machines. Cognitive science  \textbf{9}(1),  147--169 (1985)

\bibitem{andriluka20142d}
Andriluka, M., Pishchulin, L., Gehler, P., Schiele, B.: 2d human pose
  estimation: New benchmark and state of the art analysis. In: Proceedings of
  the IEEE Conference on computer Vision and Pattern Recognition. pp.
  3686--3693 (2014)

\bibitem{balakrishnan2018synthesizing}
Balakrishnan, G., Zhao, A., Dalca, A.V., Durand, F., Guttag, J.: Synthesizing
  images of humans in unseen poses. In: Proceedings of the IEEE Conference on
  Computer Vision and Pattern Recognition. pp. 8340--8348 (2018)

\bibitem{bau2019gandissect}
Bau, D., Zhu, J.Y., Strobelt, H., Zhou, B., Tenenbaum, J.B., Freeman, W.T.,
  Torralba, A.: Gan dissection: Visualizing and understanding generative
  adversarial networks. In: Proceedings of the International Conference on
  Learning Representations (ICLR) (2019)

\bibitem{biederman1981}
Biederman, I.: On the semantics of a glance at a scene. In: Perceptual
  Organization (1981)

\bibitem{brock2018large}
Brock, A., Donahue, J., Simonyan, K.: Large scale {GAN} training for high
  fidelity natural image synthesis. In: International Conference on Learning
  Representations (2019), \url{https://openreview.net/forum?id=B1xsqj09Fm}

\bibitem{caoHMP2020}
Cao, Z., Gao, H., Mangalam, K., Cai, Q., Vo, M., Malik, J.: Long-term human
  motion prediction with scene context. In: ECCV (2020)

\bibitem{cao2019openpose}
Cao, Z., Hidalgo, G., Simon, T., Wei, S.E., Sheikh, Y.: Openpose: realtime
  multi-person 2d pose estimation using part affinity fields. IEEE transactions
  on pattern analysis and machine intelligence  \textbf{43}(1),  172--186
  (2019)

\bibitem{cao2020reconstructing}
Cao, Z., Radosavovic, I., Kanazawa, A., Malik, J.: Reconstructing hand-object
  interactions in the wild. arXiv e-prints pp. arXiv--2012 (2020)

\bibitem{cao2017realtime}
Cao, Z., Simon, T., Wei, S.E., Sheikh, Y.: Realtime multi-person 2d pose
  estimation using part affinity fields. In: Proceedings of the IEEE conference
  on computer vision and pattern recognition. pp. 7291--7299 (2017)

\bibitem{carreira2019short}
Carreira, J., Noland, E., Hillier, C., Zisserman, A.: A short note on the
  kinetics-700 human action dataset. arXiv preprint arXiv:1907.06987  (2019)

\bibitem{chai2021latent}
Chai, L., Wulff, J., Isola, P.: Using latent space regression to analyze and
  leverage compositionality in gans. In: International Conference on Learning
  Representations (2021)

\bibitem{chan2019everybody}
Chan, C., Ginosar, S., Zhou, T., Efros, A.A.: Everybody dance now. In:
  Proceedings of the IEEE/CVF International Conference on Computer Vision. pp.
  5933--5942 (2019)

\bibitem{chuang2018learning}
Chuang, C.Y., Li, J., Torralba, A., Fidler, S.: Learning to act properly:
  Predicting and explaining affordances from images. In: Proceedings of the
  IEEE Conference on Computer Vision and Pattern Recognition. pp. 975--983
  (2018)

\bibitem{delaitre2012}
Delaitre, V., Fouhey, D., Laptev, I., Sivic, J., Gupta, A., Efros, A.: Scene
  semantics from long-term observation of people. In: Proc. 12th European
  Conference on Computer Vision (2012)

\bibitem{imagenetcvpr09}
Deng, J., Dong, W., Socher, R., Li, L.J., Li, K., Fei-Fei, L.: {ImageNet: A
  Large-Scale Hierarchical Image Database}. In: CVPR09 (2009)

\bibitem{diba2020large}
Diba, A., Fayyaz, M., Sharma, V., Paluri, M., Gall, J., Stiefelhagen, R.,
  Van~Gool, L.: Large scale holistic video understanding. In: European
  Conference on Computer Vision. pp. 593--610. Springer (2020)

\bibitem{divvala2009empirical}
Divvala, S.K., Hoiem, D., Hays, J.H., Efros, A.A., Hebert, M.: An empirical
  study of context in object detection. In: 2009 IEEE Conference on computer
  vision and Pattern Recognition. pp. 1271--1278. IEEE (2009)

\bibitem{LanczosFiltering}
Duchon, C.E.: Lanczos filtering in one and two dimensions. Journal of Applied
  Meteorology and Climatology  \textbf{18}(8),  1016 -- 1022 (1979).
  \doi{10.1175/1520-0450(1979)018<1016:LFIOAT>2.0.CO;2},
  \url{https://journals.ametsoc.org/view/journals/apme/18/8/1520-0450_1979_018_1016_lfioat_2_0_co_2.xml}

\bibitem{epstein2020oops}
Epstein, D., Chen, B., Vondrick, C.: Oops! predicting unintentional action in
  video. In: Proceedings of the IEEE/CVF conference on computer vision and
  pattern recognition. pp. 919--929 (2020)

\bibitem{fan2018hierarchical}
Fan, A., Lewis, M., Dauphin, Y.: Hierarchical neural story generation. arXiv
  preprint arXiv:1805.04833  (2018)

\bibitem{fouhey2012people}
Fouhey, D.F., Delaitre, V., Gupta, A., Efros, A.A., Laptev, I., Sivic, J.:
  People watching: Human actions as a cue for single view geometry. In:
  European Conference on Computer Vision. pp. 732--745. Springer (2012)

\bibitem{fouhey2018lifestyle}
Fouhey, D.F., Kuo, W.c., Efros, A.A., Malik, J.: From lifestyle vlogs to
  everyday interactions. In: Proceedings of the IEEE Conference on Computer
  Vision and Pattern Recognition. pp. 4991--5000 (2018)

\bibitem{fouhey2015defense}
Fouhey, D.F., Wang, X., Gupta, A.: In defense of the direct perception of
  affordances (2015)

\bibitem{gibson1979}
Gibson, J.J.: The Ecological Approach to Visual Perception. Houghton Mifflin,
  Boston (1979)

\bibitem{gkioxari2018detecting}
Gkioxari, G., Girshick, R., Doll{\'a}r, P., He, K.: Detecting and recognizing
  human-object interactions. In: Proceedings of the IEEE Conference on Computer
  Vision and Pattern Recognition. pp. 8359--8367 (2018)

\bibitem{goetschalckx2019ganalyze}
Goetschalckx, L., Andonian, A., Oliva, A., Isola, P.: Ganalyze: Toward visual
  definitions of cognitive image properties. In: Proceedings of the IEEE/CVF
  International Conference on Computer Vision. pp. 5744--5753 (2019)

\bibitem{goodfellow2014generative}
Goodfellow, I.J., Pouget-Abadie, J., Mirza, M., Xu, B., Warde-Farley, D.,
  Ozair, S., Courville, A., Bengio, Y.: Generative adversarial networks. arXiv
  preprint arXiv:1406.2661  (2014)

\bibitem{Grabner2011chair}
Grabner, H., Gall, J., Van~Gool, L.: What makes a chair a chair? In: Proc. IEEE
  Computer Society Conference on Computer Vision and Pattern Recognition. pp.
  1529--1536 (06 2011). \doi{10.1109/CVPR.2011.5995327}

\bibitem{GuptaSatkinEfrosHebertCVPR11}
Gupta, A., Satkin, S., Efros, A.A., Hebert, M.: From 3d scene geometry to human
  workspace. In: Computer Vision and Pattern Recognition(CVPR) (2011)

\bibitem{Hassan2021}
Hassan, M., Ghosh, P., Tesch, J., Tzionas, D., Black, M.J.: Populating {3D}
  scenes by learning human-scene interaction. In: Proceedings {IEEE/CVF}
  Conf.~on Computer Vision and Pattern Recognition ({CVPR}) (Jun 2021)

\bibitem{he2017mask}
He, K., Gkioxari, G., Doll{\'a}r, P., Girshick, R.: Mask r-cnn. In: Proceedings
  of the IEEE international conference on computer vision. pp. 2961--2969
  (2017)

\bibitem{fidscore}
Heusel, M., Ramsauer, H., Unterthiner, T., Nessler, B., Hochreiter, S.: Gans
  trained by a two time-scale update rule converge to a local nash equilibrium.
  In: Proceedings of the 31st International Conference on Neural Information
  Processing Systems. p. 6629–6640. NIPS'17 (2017)

\bibitem{holtzman2019curious}
Holtzman, A., Buys, J., Du, L., Forbes, M., Choi, Y.: The curious case of
  neural text degeneration. arXiv preprint arXiv:1904.09751  (2019)

\bibitem{ganspace2020}
Härkönen, E., Hertzmann, A., Lehtinen, J., Paris, S.: Ganspace: Discovering
  interpretable gan controls. In: Proc. NeurIPS (2020)

\bibitem{ionescu2013human3}
Ionescu, C., Papava, D., Olaru, V., Sminchisescu, C.: Human3. 6m: Large scale
  datasets and predictive methods for 3d human sensing in natural environments.
  IEEE transactions on pattern analysis and machine intelligence
  \textbf{36}(7),  1325--1339 (2013)

\bibitem{isola2017image}
Isola, P., Zhu, J.Y., Zhou, T., Efros, A.A.: Image-to-image translation with
  conditional adversarial networks. In: Computer Vision and Pattern Recognition
  (CVPR), 2017 IEEE Conference on (2017)

\bibitem{gansteerability}
Jahanian, A., Chai, L., Isola, P.: On the "steerability" of generative
  adversarial networks. In: International Conference on Learning
  Representations (2020)

\bibitem{Jiang2013CVPR}
Jiang, Y., Koppula, H., Saxena, A.: Hallucinated humans as the hidden context
  for labeling 3d scenes. In: Proceedings of the IEEE Conference on Computer
  Vision and Pattern Recognition (CVPR) (June 2013)

\bibitem{johnson2016perceptual}
Johnson, J., Alahi, A., Fei-Fei, L.: Perceptual losses for real-time style
  transfer and super-resolution. In: European conference on computer vision.
  pp. 694--711. Springer (2016)

\bibitem{kanazawa2019learning}
Kanazawa, A., Zhang, J.Y., Felsen, P., Malik, J.: Learning 3d human dynamics
  from video. In: Proceedings of the IEEE/CVF Conference on Computer Vision and
  Pattern Recognition. pp. 5614--5623 (2019)

\bibitem{karras2018progressive}
Karras, T., Aila, T., Laine, S., Lehtinen, J.: Progressive growing of {GAN}s
  for improved quality, stability, and variation. In: International Conference
  on Learning Representations (2018),
  \url{https://openreview.net/forum?id=Hk99zCeAb}

\bibitem{karras2020training}
Karras, T., Aittala, M., Hellsten, J., Laine, S., Lehtinen, J., Aila, T.:
  Training generative adversarial networks with limited data. arXiv preprint
  arXiv:2006.06676  (2020)

\bibitem{karras2019style}
Karras, T., Laine, S., Aila, T.: A style-based generator architecture for
  generative adversarial networks. In: Proceedings of the IEEE/CVF Conference
  on Computer Vision and Pattern Recognition. pp. 4401--4410 (2019)

\bibitem{karras2020analyzing}
Karras, T., Laine, S., Aittala, M., Hellsten, J., Lehtinen, J., Aila, T.:
  Analyzing and improving the image quality of stylegan. In: Proceedings of the
  IEEE/CVF Conference on Computer Vision and Pattern Recognition. pp.
  8110--8119 (2020)

\bibitem{kay2017kinetics}
Kay, W., Carreira, J., Simonyan, K., Zhang, B., Hillier, C., Vijayanarasimhan,
  S., Viola, F., Green, T., Back, T., Natsev, P., et~al.: The kinetics human
  action video dataset. arXiv preprint arXiv:1705.06950  (2017)

\bibitem{kingma2015adam}
Kingma, D.P., Ba, J.: Adam: A methodfor stochastic optimization. In:
  International Conference onLearning Representations (ICLR) (2015)

\bibitem{glow2018}
Kingma, D.P., Dhariwal, P.: Glow: Generative flow with invertible 1x1
  convolutions. In: NeurIPS. pp. 10236--10245 (2018),
  \url{http://papers.nips.cc/paper/8224-glow-generative-flow-with-invertible-1x1-convolutions}

\bibitem{koppula2013learning}
Koppula, H.S., Gupta, R., Saxena, A.: Learning human activities and object
  affordances from rgb-d videos. The International Journal of Robotics Research
   \textbf{32}(8),  951--970 (2013)

\bibitem{lee2002interactive}
Lee, J., Chai, J., Reitsma, P.S., Hodgins, J.K., Pollard, N.S.: Interactive
  control of avatars animated with human motion data. In: Proceedings of the
  29th annual conference on Computer graphics and interactive techniques. pp.
  491--500 (2002)

\bibitem{3d-affordance}
Li, X., Liu, S., Kim, K., Wang, X., Yang, M.H., Kautz, J.: Putting humans in a
  scene: Learning affordance in 3d indoor environments. In: CVPR (2019)

\bibitem{Li2019CVPR}
Li, Y., Huang, C., Loy, C.C.: Dense intrinsic appearance flow for human pose
  transfer. In: Proceedings of the IEEE/CVF Conference on Computer Vision and
  Pattern Recognition (CVPR) (June 2019)

\bibitem{li-cvpr2020}
Li, Y., Singh, K.K., Ojha, U., Lee, Y.J.: Mixnmatch: Multifactor
  disentanglement and encoding for conditional image generation. In: CVPR
  (2020)

\bibitem{liu2015faceattributes}
Liu, Z., Luo, P., Wang, X., Tang, X.: Deep learning face attributes in the
  wild. In: Proceedings of International Conference on Computer Vision (ICCV)
  (December 2015)

\bibitem{ma2017pose}
Ma, L., Jia, X., Sun, Q., Schiele, B., Tuytelaars, T., Van~Gool, L.: Pose
  guided person image generation. In: Advances in Neural Information Processing
  Systems. pp. 405--415 (2017)

\bibitem{ma2017disentangled}
Ma, L., Sun, Q., Georgoulis, S., Van~Gool, L., Schiele, B., Fritz, M.:
  Disentangled person image generation. In: The IEEE International Conference
  on Computer Vision and Pattern Recognition (CVPR) (June 2018)

\bibitem{van2008visualizing}
Van~der Maaten, L., Hinton, G.: Visualizing data using t-sne. Journal of
  machine learning research  \textbf{9}(11) (2008)

\bibitem{marchesi2017megapixel}
Marchesi, M.: Megapixel size image creation using generative adversarial
  networks (2017)

\bibitem{Mescheder2018ICML}
Mescheder, L., Nowozin, S., Geiger, A.: Which training methods for gans do
  actually converge? In: International Conference on Machine Learning (ICML)
  (2018)

\bibitem{mirza2014conditional}
Mirza, M., Osindero, S.: Conditional generative adversarial nets. arXiv
  preprint arXiv:1411.1784  (2014)

\bibitem{mokady2022self}
Mokady, R., Yarom, M., Tov, O., Lang, O., Cohen-Or, D., Dekel, T., Irani, M.,
  Mosseri, I.: Self-distilled stylegan: Towards generation from internet
  photos. arXiv preprint arXiv:2202.12211  (2022)

\bibitem{monfort2019moments}
Monfort, M., Andonian, A., Zhou, B., Ramakrishnan, K., Bargal, S.A., Yan, T.,
  Brown, L., Fan, Q., Gutfreund, D., Vondrick, C., et~al.: Moments in time
  dataset: one million videos for event understanding. IEEE transactions on
  pattern analysis and machine intelligence  \textbf{42}(2),  502--508 (2019)

\bibitem{mottaghi2014role}
Mottaghi, R., Chen, X., Liu, X., Cho, N.G., Lee, S.W., Fidler, S., Urtasun, R.,
  Yuille, A.: The role of context for object detection and semantic
  segmentation in the wild. In: Proceedings of the IEEE Conference on Computer
  Vision and Pattern Recognition. pp. 891--898 (2014)

\bibitem{park2020swapping}
Park, T., Zhu, J.Y., Wang, O., Lu, J., Shechtman, E., Efros, A.A., Zhang, R.:
  Swapping autoencoder for deep image manipulation. In: Advances in Neural
  Information Processing Systems (2020)

\bibitem{peebles2020hessian}
Peebles, W., Peebles, J., Zhu, J.Y., Efros, A.A., Torralba, A.: The hessian
  penalty: A weak prior for unsupervised disentanglement. In: Proceedings of
  European Conference on Computer Vision (ECCV) (2020)

\bibitem{rabinovich2007objects}
Rabinovich, A., Vedaldi, A., Galleguillos, C., Wiewiora, E., Belongie, S.:
  Objects in context. In: 2007 IEEE 11th International Conference on Computer
  Vision. pp.~1--8. IEEE (2007)

\bibitem{reed2016generative}
Reed, S., Akata, Z., Yan, X., Logeswaran, L., Schiele, B., Lee, H.: Generative
  adversarial text to image synthesis. In: International Conference on Machine
  Learning. pp. 1060--1069. PMLR (2016)

\bibitem{sauer2022stylegan}
Sauer, A., Schwarz, K., Geiger, A.: Stylegan-xl: Scaling stylegan to large
  diverse datasets. arXiv preprint arXiv:2202.00273  (2022)

\bibitem{siarohin2018deformable}
Siarohin, A., Sangineto, E., Lathuiliere, S., Sebe, N.: Deformable gans for
  pose-based human image generation. In: Proceedings of the IEEE Conference on
  Computer Vision and Pattern Recognition. pp. 3408--3416 (2018)

\bibitem{sigurdsson2016hollywood}
Sigurdsson, G.A., Varol, G., Wang, X., Farhadi, A., Laptev, I., Gupta, A.:
  Hollywood in homes: Crowdsourcing data collection for activity understanding.
  In: European Conference on Computer Vision. pp. 510--526. Springer (2016)

\bibitem{szegedy2015going}
Szegedy, C., Liu, W., Jia, Y., Sermanet, P., Reed, S., Anguelov, D., Erhan, D.,
  Vanhoucke, V., Rabinovich, A.: Going deeper with convolutions. In:
  Proceedings of the IEEE conference on computer vision and pattern
  recognition. pp.~1--9 (2015)

\bibitem{torralba2003context}
Torralba, A., Murphy, K.P., Freeman, W.T., Rubin, M.A.: Context-based vision
  system for place and object recognition. In: Computer Vision, IEEE
  International Conference on. vol.~2, pp. 273--273. IEEE Computer Society
  (2003)

\bibitem{wang2020synthesizing}
Wang, J., Xu, H., Xu, J., Liu, S., Wang, X.: Synthesizing long-term 3d human
  motion and interaction in 3d scenes (2020)

\bibitem{wang2018vid2vid}
Wang, T.C., Liu, M.Y., Zhu, J.Y., Liu, G., Tao, A., Kautz, J., Catanzaro, B.:
  Video-to-video synthesis. In: Advances in Neural Information Processing
  Systems (NeurIPS) (2018)

\bibitem{wang2018pix2pixHD}
Wang, T.C., Liu, M.Y., Zhu, J.Y., Tao, A., Kautz, J., Catanzaro, B.:
  High-resolution image synthesis and semantic manipulation with conditional
  gans. In: Proceedings of the IEEE Conference on Computer Vision and Pattern
  Recognition (2018)

\bibitem{WangaffordanceCVPR2017}
Wang, X., Girdhar, R., Gupta, A.: Binge watching: Scaling affordance learning
  from sitcoms. In: CVPR (2017)

\bibitem{wu2019detectron2}
Wu, Y., Kirillov, A., Massa, F., Lo, W.Y., Girshick, R.: Detectron2.
  \url{https://github.com/facebookresearch/detectron2} (2019)

\bibitem{xu2018youtube}
Xu, N., Yang, L., Fan, Y., Yang, J., Yue, D., Liang, Y., Price, B., Cohen, S.,
  Huang, T.: Youtube-vos: Sequence-to-sequence video object segmentation. In:
  Proceedings of the European Conference on Computer Vision (ECCV). pp.
  585--601 (2018)

\bibitem{Yao2010ModelingMC}
Yao, B., Fei-Fei, L.: Modeling mutual context of object and human pose in
  human-object interaction activities. 2010 IEEE Computer Society Conference on
  Computer Vision and Pattern Recognition pp. 17--24 (2010)

\bibitem{lsun}
Yu, F., Zhang, Y., Song, S., Seff, A., Xiao, J.: Lsun: Construction of a
  large-scale image dataset using deep learning with humans in the loop. CoRR
  \textbf{abs/1506.03365} (2015),
  \url{http://dblp.uni-trier.de/db/journals/corr/corr1506.html\#YuZSSX15}

\bibitem{zhang2018perceptual}
Zhang, R., Isola, P., Efros, A.A., Shechtman, E., Wang, O.: The unreasonable
  effectiveness of deep features as a perceptual metric. In: CVPR (2018)

\bibitem{zhang2013actemes}
Zhang, W., Zhu, M., Derpanis, K.G.: From actemes to action: A
  strongly-supervised representation for detailed action understanding. In:
  Proceedings of the IEEE International Conference on Computer Vision. pp.
  2248--2255 (2013)

\bibitem{zhao2020differentiable}
Zhao, S., Liu, Z., Lin, J., Zhu, J.Y., Han, S.: Differentiable augmentation for
  data-efficient gan training. arXiv preprint arXiv:2006.10738  (2020)

\bibitem{zhu2014reasoning}
Zhu, Y., Fathi, A., Fei-Fei, L.: Reasoning about object affordances in a
  knowledge base representation. In: European conference on computer vision.
  pp. 408--424. Springer (2014)

\end{thebibliography}

\appendix

\section{\emph{Humans in Context} Meta-dataset Details}

Our dataset contains diverse footage of humans immersed in everyday environments. Each image is supplemented with pseudo-ground truth human pose attained using OpenPose~\cite{cao2017realtime,cao2019openpose}. The data is sourced from 10 existing human and action recognition datasets, with the numbers of clips and frames from each source dataset detailed in Table~\ref{tab:dataset}. Video footage provides a vast source of diverse human activity, and ensures all poses are represented, rather than only human poses photographers choose to capture in still images. For the MPII~\cite{andriluka20142d} dataset, which is primarily a still image dataset, we use short video clips of the frames preceding and following each image.

Each dataset contains unique biases, and combining data sources is less subject to the bias of any particular dataset. Different datasets also offer different types of scenes. For example, Moments~\cite{monfort2019moments} includes classes absent from HVU~\cite{diba2020large}, and Oops~\cite{epstein2020oops} contains uncommon accidental actions. The number of useful examples from each source was only evident after extensive curation.

\begin{table*}[t]
\centering
\caption{\textbf{Humans in Context source data.} Our dataset consists of video clips filtered from 10 existing human and action recognition datasets. High quality clips have sufficient bitrate, framerate and resolution. Person clips are those where pretrained person detection and pose prediction networks assert that a single person is present. In total we curate \num{229595} clips and \num{19503700} frames.}
% \vspace{-10pt}
\label{tab:dataset}
\begin{tabular}{llllll}
\toprule
              & \multicolumn{3}{c}{\# Video Clips}    & \multicolumn{2}{c}{\# Frames} \\
\cmidrule(lr){2-4}\cmidrule(lr){5-6}
              & Source    & High Quality & Person  & High Quality & Person \\
\midrule
HVU~\cite{diba2020large}           & 566,489   & 353,174      & 105,634 & 98,603,223   & 9,590,407   \\
Moments~\cite{monfort2019moments}       & 757,804   & 653,368      & 54,156  & 56,074,418   & 3,374,112   \\
Kinetics-700-2020~\cite{kay2017kinetics,carreira2019short}     & 620,119   & 432,502      & 26,911  & 78,037,500   & 2,428,079   \\
Charades~\cite{sigurdsson2016hollywood}      & 9,848     & 7,319        & 16,967  & 6,256,421    & 2,157,074   \\
InstaVariety~\cite{kanazawa2019learning}  & 2,545     & 2,449        & 5,773   & 1,898,824    & 730,211     \\
Oops~\cite{epstein2020oops}          & 29,940    & 27,953       & 8,360   & 5,738,042    & 596,488     \\
MPII~\cite{andriluka20142d}          & 24,987    & 24,980       & 8,820   & 1,025,459    & 352,498     \\
VLOG-people~\cite{fouhey2018lifestyle} & 663       & 555          & 1,261   & 321,071      & 163,956     \\
PennAction~\cite{zhang2013actemes}    & 2,326     & 2,221        & 1,208   & 161,029      & 76,112      \\
YouTube-VOS~\cite{xu2018youtube}   & 4,519     & 4,511        & 505     & 613,441      & 34,763      \\
\midrule
Total         & 2,019,240 & 1,509,032    & \textbf{229,595} & 248,729,428  & \textbf{19,503,700} \\
\bottomrule
\end{tabular}
\end{table*}

We filter out videos where either dimension is shorter than $256$ pixels, and we resize remaining videos using Lanczos resampling~\cite{LanczosFiltering} such that the smaller edge is exactly $256$ pixels. We exclude videos with an average bitrate below $0.9$ bits per pixel, or with a framerate that does not fall between (and cannot be subsempled to fall between) 23.9 fps and 30 fps. Videos are truncated to $3000$ frames. Source datasets which provide pre-extracted frames only undergo quality filtering by spatial resolution.

Frames are then filtered to contain a single person using pretrained Keypoint R-CNN~\cite{he2017mask,wu2019detectron2} person detection. Person bounding boxes are detected for each frame, with a minimum accuracy of $95 \%$, a minimum bounding box area of $1\%$ of the total frame area, and non-maximum suppression of overlapping bounding boxes with an intersection over union greater than $0.3$. With these thresholds, any frame with more than a single person detected is removed. Stricter thresholds are then applied to the remaining frames with a single person bounding box: a minimum accuracy of $98 \%$, a minimum bounding box area of $4\%$ of the total frame area, and a maximum bounding box area of $80\%$ of the total frame area. These thresholds ensure with high accuracy that there is a single person present in the frame at a reasonable size. Frames are then cropped to a $256 \times 256$ resolution toward the average bounding box center for each contiguous segment of frames.

Pseudo-ground truth pose labels are computed for each frame using OpenPose~\cite{cao2017realtime,cao2019openpose} keypoint prediction. We use the single-scale OpenPose version to compute 18 body keypoints. Similar to person detection, we use a relaxed total score threshold of $2.5$ when filtering for multiple people, and a strict total score threshold of $10.0$ when ensuring there is a single person. Each individual keypoint has a score threshold of $0.3$, and keypoints below this threshold are marked as not visible in the frame. To avoid frames of just legs or torso, we only include frames where the keypoint at the base of the neck is visible, and where a total of at least $8$ of $14$ keypoints (excluding eyes and ears) are visible.

The final dataset only includes clips of at least $30$ adjacent frames where each frame passed filtering. Note that multiple clips may be sourced from the same video, and that duplicate videos from different source datasets are possible.

\subsection{Dataset licenses}

The HVU dataset~\cite{diba2020large} is released for non-commercial research and educational purposes only, and was attained directly from the dataset authors. The Moments in Time~\cite{monfort2019moments} dataset is released for non-commercial research and educational purposes, and was attained from the dataset project website. The Kinetics dataset~\cite{kay2017kinetics,carreira2019short} is licensed by Google Inc.\ under a Creative Commons Attribution 4.0 International License, and videos were downloaded directly from YouTube. The Charades dataset~\cite{sigurdsson2016hollywood} is released under a non-commercial license detailed here: \url{https://prior.allenai.org/projects/data/charades/license.txt}; data was downloaded from the project webpage. The InstaVariety dataset~\cite{kanazawa2019learning} is released for non-commercial academic use, and was attained directly from the dataset authors. The Oops dataset~\cite{epstein2020oops} is released under the Creative Commons Attribution-NonCommercial-ShareAlike 4.0 International License, and was downloaded from the project webpage. The MPII~\cite{andriluka20142d} dataset is released under the Simplified BSD License detailed here: \url{https://github.com/peiyunh/rg-mpii/blob/master/data/mpii_human/annotation/bsd.txt}; data was downloaded from the project webpage. The VLOG dataset~\cite{fouhey2018lifestyle} is released for non-commercial research purposes only, and was downloaded from the project webpage. The PennAction dataset~\cite{zhang2013actemes} is released without a license and was downloaded from the project webpage; dataset authors confirmed there are no terms of use and only ask the corresponding paper~\cite{zhang2013actemes} be cited. The YouTube-VOS dataset~\cite{xu2018youtube} is released for non-commercial research purposes only and was downloaded from the project challenge webpage.

\section{Model Implementation Details}

We train all models at $128 \times 128$ resolution. Many aspects of our model are borrowed directly from StyleGAN2~\cite{karras2020analyzing}, including non-saturating logistic loss~\cite{goodfellow2014generative}, equalized learning rates for
all parameters~\cite{karras2018progressive}, $R_1$ regularization~\cite{Mescheder2018ICML}, path length regularization~\cite{karras2020analyzing}, and exponential moving average of generator parameters~\cite{karras2018progressive}.

We use a learning rate of $2.5 \times 10^{-3}$, an exponential moving average rate of $\beta = 0.995$, a moving average warmup of \num{150000} steps, and $R_1$ regularization strength of $\gamma = 0.05$. We remove spatial noise maps to isolate control over the scene to the latent code. We also remove style mixing regularization during training. Our final model was trained with a minibatch size of $120$ on $10\times$ NVIDIA Quadro RTX GPUs, and for \num{1000000} steps. Our ablations trained for 1 week, and we let the final large model continue for 3 weeks. The large generator has 85.4M parameters and discriminator has 98.2M. Ablations and the Pix2PixHD baseline were trained with a batch size of $40$ on $5\times$ NVIDIA GeForce RTX 2080 GPUs for \num{600000} iterations. Multiple checkpoints were saved throughout training, and the checkpoint with the lowest FID score was used for all evaluation. The Pix2Pix baseline was trained for \num{10000000} iterations with a batch size of $40$ on $5\times$ NVIDIA GeForce RTX 2080 GPUs .

Code used from the public implementation of StyleGAN2-Ada is released under the NVIDIA code license found here: \sloppy\url{https://github.com/NVlabs/stylegan2-ada/blob/main/LICENSE.txt}. Code used to run the Pix2Pix baseline is released under the BSD License license found here: \url{https://github.com/junyanz/pytorch-CycleGAN-and-pix2pix/blob/master/LICENSE}.

\subsection{Data augmentation}

\begin{figure}[t]
    \centering
    \begin{subfigure}[b]{0.32\linewidth}
        \centering
        \includegraphics[width=\linewidth]{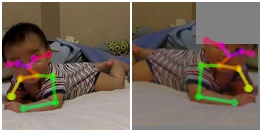}
    \end{subfigure}
    \hfill
    \begin{subfigure}[b]{0.32\linewidth}
        \centering
        \includegraphics[width=\linewidth]{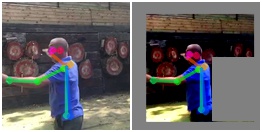}
    \end{subfigure}
    \hfill
    \begin{subfigure}[b]{0.32\linewidth}
        \centering
        \includegraphics[width=\linewidth]{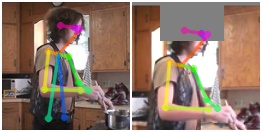}
    \end{subfigure}
    \caption{\textbf{Data augmentation.} We apply random spatial, cutout and color augmentations to frames and poses just prior to the discriminator network. Each pair above shows the original frame and pose on the left and the augmented output on the right.}
    \label{fig:augmentation}
\end{figure}

Data augmentation of both generated and real images just prior to the discriminator can improve robustness and prevent the discriminator from overfitting to the train dataset~\cite{karras2020training,zhao2020differentiable}. Our augmentation parameters are largely based on~\cite{zhao2020differentiable}. Brightness is augmented by randomly offsetting intensity by a value uniformly sampled from $-25\%$ to $+25\%$. Saturation is augmented by interpolating red, green and blue channels toward or away from the mean of all three at each pixel, with interpolation weights uniformly sampled from $0.0$ to $2.0$. Contrast is augmented by interpolating color values toward or away from the mean of all color values in an entire frame sequence, with interpolation weights uniformly sampled from $0.5$ to $1.5$. Horizontal flipping is applied with a $50\%$ chance to all frames and poses in a sequence. Frames are scaled by a factor uniformly sampled from $0.8$ to $1.25$ and translated by an offset sampled uniformly from $-12.5\%$ to $+12.5\%$. A random cutout, half the size of each dimension and randomly placed, is erased from each frame. Spatial transformations applied to frames are also applied to poses so that the frames and poses still correspond correctly. We briefly experimented with dropout augmentation of pose, but did not find it helpful. See Figure~\ref{fig:augmentation} for examples of our data augmentation.

\subsection{Mismatch discrimination}

\begin{figure}[t]
    \begin{subfigure}[b]{0.32\linewidth}
        \centering
        \includegraphics[width=\linewidth]{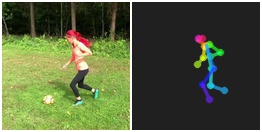}
        \caption{Real}
    \end{subfigure}
    \hfill
    \begin{subfigure}[b]{0.32\linewidth}
        \centering
        \includegraphics[width=\linewidth]{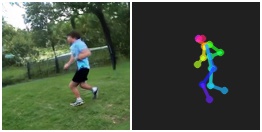}
        \caption{Generated}
    \end{subfigure}
    \hfill
    \begin{subfigure}[b]{0.32\linewidth}
        \centering
        \includegraphics[width=\linewidth]{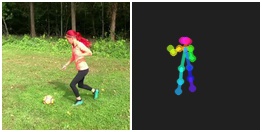}
        \caption{Mismatched}
    \end{subfigure}
    \caption{\textbf{Mismatch discrimination.} The discriminator must classify (a) a real image and pose as real, (b) a generated frame and conditional pose as fake, and (c) a real frame and mismatched pose as fake.}
    \label{fig:mismatch}
\end{figure}

We force the discriminator to pay attention to pose conditioning by providing a mismatched real image with the incorrect pose conditioning as an additional fake example. For the mismatched fake example, the pose embedding and keypoint heatmaps both take pose from another sample in the minibatch. This training method was first introduced in text-to-image generation~\cite{reed2016generative} but has not been widely used in the image or video translation literature; we found training with mismatch discrimination provides a slight improvement, forcing the discriminator to use conditioning.

\subsection{Human disentanglement}

Our generator can be used to place a human subject in a new scene, as we outline in Section~5.3 of the main paper. We accomplish this by optimizing for a latent code which produces a scene matching one image and a subject matching another. We separately optimize the latent code used at each \emph{scale} of our model, which is similar to the $\mathcal{W^+}$ space~\cite{im2style} used for inversion (although slightly lower dimensional). We minimize perceptual loss~\cite{johnson2016perceptual,zhang2018perceptual} between a subject-only crop of the first generated image and the composition. When generating subject-only images, we zero out the learned constant input to the StyleGAN2 generator~\cite{karras2020analyzing}, which we found helps isolate the subject from the background. The crop region is attained from human pose. We also minimize perceptual loss between scene-only versions of the second generated image and composition image. We optimize for $1000$ steps using the Adam optimizer~\cite{kingma2015adam} and a learning rate of $0.05$.

\section{Additional Results}

\begin{figure*}[h!]
\centering

\begin{subfigure}[b]{1.0\linewidth}
\textoverlay{A1}{\includegraphics[width=0.087\linewidth]{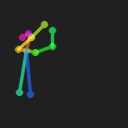}}%
\hfill%
\hfill%
\hfill%
\includegraphics[width=0.087\linewidth]{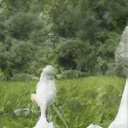}%
\hfill%
\includegraphics[width=0.087\linewidth]{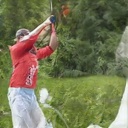}%
\hfill%
\hfill%
\hfill%
\includegraphics[width=0.087\linewidth]{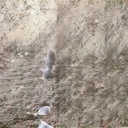}%
\hfill%
\includegraphics[width=0.087\linewidth]{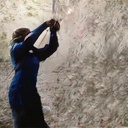}%
\hfill%
\hfill%
\hfill%
\includegraphics[width=0.087\linewidth]{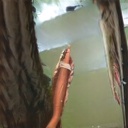}%
\hfill%
\includegraphics[width=0.087\linewidth]{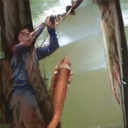}%
\hfill%
\hfill%
\hfill%
\includegraphics[width=0.087\linewidth]{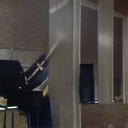}%
\hfill%
\includegraphics[width=0.087\linewidth]{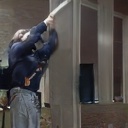}%
\hfill%
\hfill%
\hfill%
\includegraphics[width=0.087\linewidth]{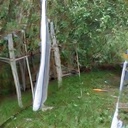}%
\hfill%
\includegraphics[width=0.087\linewidth]{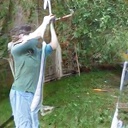}%
\end{subfigure}

% \vspace{0.9pt}

\begin{subfigure}[b]{1.0\linewidth}
\textoverlay{B1}{\includegraphics[width=0.087\linewidth]{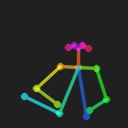}}%
\hfill%
\hfill%
\hfill%
\includegraphics[width=0.087\linewidth]{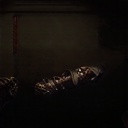}%
\hfill%
\includegraphics[width=0.087\linewidth]{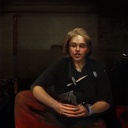}%
\hfill%
\hfill%
\hfill%
\includegraphics[width=0.087\linewidth]{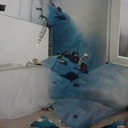}%
\hfill%
\includegraphics[width=0.087\linewidth]{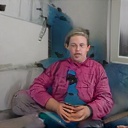}%
\hfill%
\hfill%
\hfill%
\includegraphics[width=0.087\linewidth]{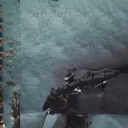}%
\hfill%
\includegraphics[width=0.087\linewidth]{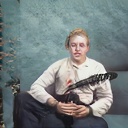}%
\hfill%
\hfill%
\hfill%
\includegraphics[width=0.087\linewidth]{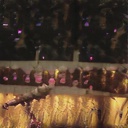}%
\hfill%
\includegraphics[width=0.087\linewidth]{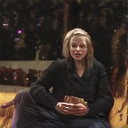}%
\hfill%
\hfill%
\hfill%
\includegraphics[width=0.087\linewidth]{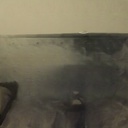}%
\hfill%
\includegraphics[width=0.087\linewidth]{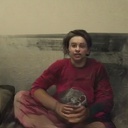}%
\end{subfigure}

% \vspace{0.9pt}

\begin{subfigure}[b]{1.0\linewidth}
\textoverlay{C1}{\includegraphics[width=0.087\linewidth]{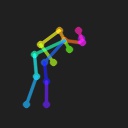}}%
\hfill%
\hfill%
\hfill%
\includegraphics[width=0.087\linewidth]{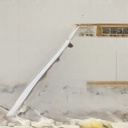}%
\hfill%
\includegraphics[width=0.087\linewidth]{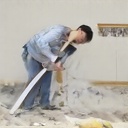}%
\hfill%
\hfill%
\hfill%
\includegraphics[width=0.087\linewidth]{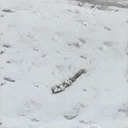}%
\hfill%
\includegraphics[width=0.087\linewidth]{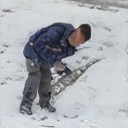}%
\hfill%
\hfill%
\hfill%
\includegraphics[width=0.087\linewidth]{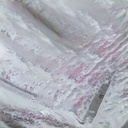}%
\hfill%
\includegraphics[width=0.087\linewidth]{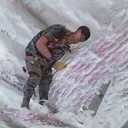}%
\hfill%
\hfill%
\hfill%
\includegraphics[width=0.087\linewidth]{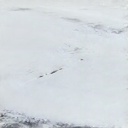}%
\hfill%
\includegraphics[width=0.087\linewidth]{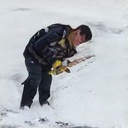}%
\hfill%
\hfill%
\hfill%
\includegraphics[width=0.087\linewidth]{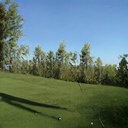}%
\hfill%
\includegraphics[width=0.087\linewidth]{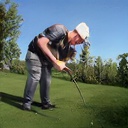}%
\end{subfigure}

% \vspace{0.9pt}

\begin{subfigure}[b]{1.0\linewidth}
\textoverlay{D1}{\includegraphics[width=0.087\linewidth]{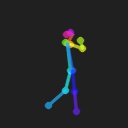}}%
\hfill%
\hfill%
\hfill%
\includegraphics[width=0.087\linewidth]{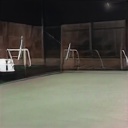}%
\hfill%
\includegraphics[width=0.087\linewidth]{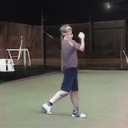}%
\hfill%
\hfill%
\hfill%
\includegraphics[width=0.087\linewidth]{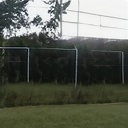}%
\hfill%
\includegraphics[width=0.087\linewidth]{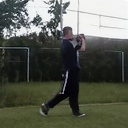}%
\hfill%
\hfill%
\hfill%
\includegraphics[width=0.087\linewidth]{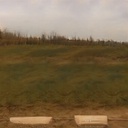}%
\hfill%
\includegraphics[width=0.087\linewidth]{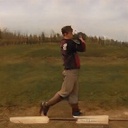}%
\hfill%
\hfill%
\hfill%
\includegraphics[width=0.087\linewidth]{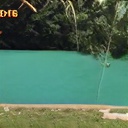}%
\hfill%
\includegraphics[width=0.087\linewidth]{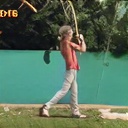}%
\hfill%
\hfill%
\hfill%
\includegraphics[width=0.087\linewidth]{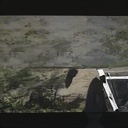}%
\hfill%
\includegraphics[width=0.087\linewidth]{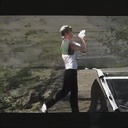}%
\end{subfigure}

% \vspace{0.9pt}

\begin{subfigure}[b]{1.0\linewidth}
\textoverlay{E1}{\includegraphics[width=0.087\linewidth]{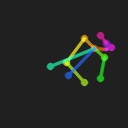}}%
\hfill%
\hfill%
\hfill%
\includegraphics[width=0.087\linewidth]{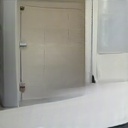}%
\hfill%
\includegraphics[width=0.087\linewidth]{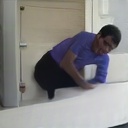}%
\hfill%
\hfill%
\hfill%
\includegraphics[width=0.087\linewidth]{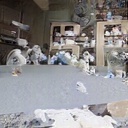}%
\hfill%
\includegraphics[width=0.087\linewidth]{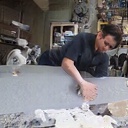}%
\hfill%
\hfill%
\hfill%
\includegraphics[width=0.087\linewidth]{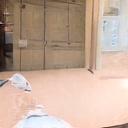}%
\hfill%
\includegraphics[width=0.087\linewidth]{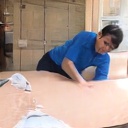}%
\hfill%
\hfill%
\hfill%
\includegraphics[width=0.087\linewidth]{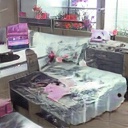}%
\hfill%
\includegraphics[width=0.087\linewidth]{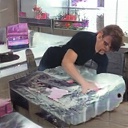}%
\hfill%
\hfill%
\hfill%
\includegraphics[width=0.087\linewidth]{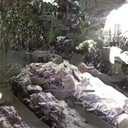}%
\hfill%
\includegraphics[width=0.087\linewidth]{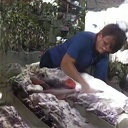}%
\end{subfigure}

% \vspace{0.9pt}

\begin{subfigure}[b]{1.0\linewidth}
\textoverlay{F1}{\includegraphics[width=0.087\linewidth]{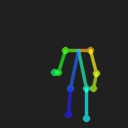}}%
\hfill%
\hfill%
\hfill%
\includegraphics[width=0.087\linewidth]{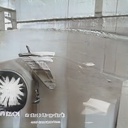}%
\hfill%
\includegraphics[width=0.087\linewidth]{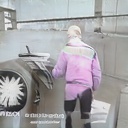}%
\hfill%
\hfill%
\hfill%
\includegraphics[width=0.087\linewidth]{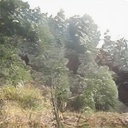}%
\hfill%
\includegraphics[width=0.087\linewidth]{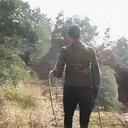}%
\hfill%
\hfill%
\hfill%
\includegraphics[width=0.087\linewidth]{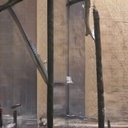}%
\hfill%
\includegraphics[width=0.087\linewidth]{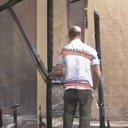}%
\hfill%
\hfill%
\hfill%
\includegraphics[width=0.087\linewidth]{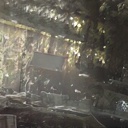}%
\hfill%
\includegraphics[width=0.087\linewidth]{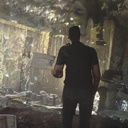}%
\hfill%
\hfill%
\hfill%
\includegraphics[width=0.087\linewidth]{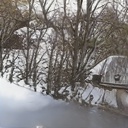}%
\hfill%
\includegraphics[width=0.087\linewidth]{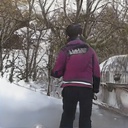}%
\end{subfigure}

% \vspace{0.9pt}

\begin{subfigure}[b]{1.0\linewidth}
\textoverlay{G1}{\includegraphics[width=0.087\linewidth]{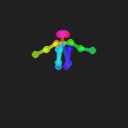}}%
\hfill%
\hfill%
\hfill%
\includegraphics[width=0.087\linewidth]{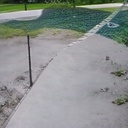}%
\hfill%
\includegraphics[width=0.087\linewidth]{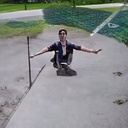}%
\hfill%
\hfill%
\hfill%
\includegraphics[width=0.087\linewidth]{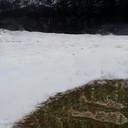}%
\hfill%
\includegraphics[width=0.087\linewidth]{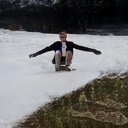}%
\hfill%
\hfill%
\hfill%
\includegraphics[width=0.087\linewidth]{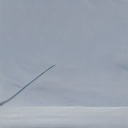}%
\hfill%
\includegraphics[width=0.087\linewidth]{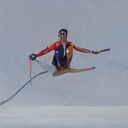}%
\hfill%
\hfill%
\hfill%
\includegraphics[width=0.087\linewidth]{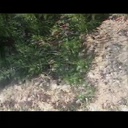}%
\hfill%
\includegraphics[width=0.087\linewidth]{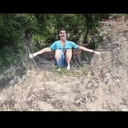}%
\hfill%
\hfill%
\hfill%
\includegraphics[width=0.087\linewidth]{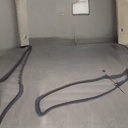}%
\hfill%
\includegraphics[width=0.087\linewidth]{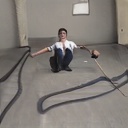}%
\end{subfigure}

% \vspace{0.9pt}

\begin{subfigure}[b]{1.0\linewidth}
\textoverlay{H1}{\includegraphics[width=0.087\linewidth]{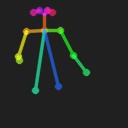}}%
\hfill%
\hfill%
\hfill%
\includegraphics[width=0.087\linewidth]{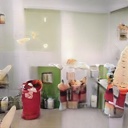}%
\hfill%
\includegraphics[width=0.087\linewidth]{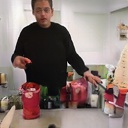}%
\hfill%
\hfill%
\hfill%
\includegraphics[width=0.087\linewidth]{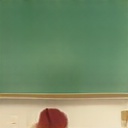}%
\hfill%
\includegraphics[width=0.087\linewidth]{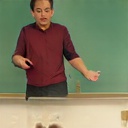}%
\hfill%
\hfill%
\hfill%
\includegraphics[width=0.087\linewidth]{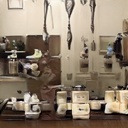}%
\hfill%
\includegraphics[width=0.087\linewidth]{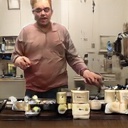}%
\hfill%
\hfill%
\hfill%
\includegraphics[width=0.087\linewidth]{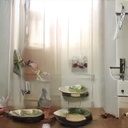}%
\hfill%
\includegraphics[width=0.087\linewidth]{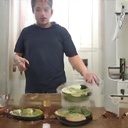}%
\hfill%
\hfill%
\hfill%
\includegraphics[width=0.087\linewidth]{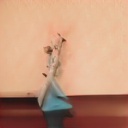}%
\hfill%
\includegraphics[width=0.087\linewidth]{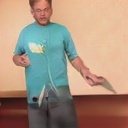}%
\end{subfigure}

% \vspace{0.9pt}

\begin{subfigure}[b]{1.0\linewidth}
\textoverlay{I1}{\includegraphics[width=0.087\linewidth]{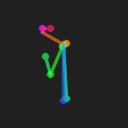}}%
\hfill%
\hfill%
\hfill%
\includegraphics[width=0.087\linewidth]{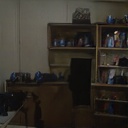}%
\hfill%
\includegraphics[width=0.087\linewidth]{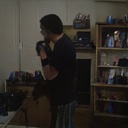}%
\hfill%
\hfill%
\hfill%
\includegraphics[width=0.087\linewidth]{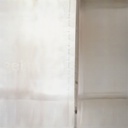}%
\hfill%
\includegraphics[width=0.087\linewidth]{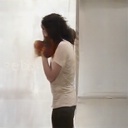}%
\hfill%
\hfill%
\hfill%
\includegraphics[width=0.087\linewidth]{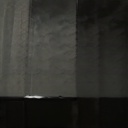}%
\hfill%
\includegraphics[width=0.087\linewidth]{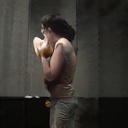}%
\hfill%
\hfill%
\hfill%
\includegraphics[width=0.087\linewidth]{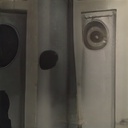}%
\hfill%
\includegraphics[width=0.087\linewidth]{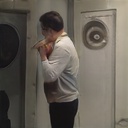}%
\hfill%
\hfill%
\hfill%
\includegraphics[width=0.087\linewidth]{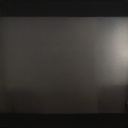}%
\hfill%
\includegraphics[width=0.087\linewidth]{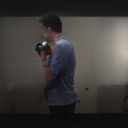}%
\end{subfigure}

% \vspace{0.9pt}

\begin{subfigure}[b]{1.0\linewidth}
\textoverlay{J1}{\includegraphics[width=0.087\linewidth]{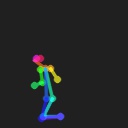}}%
\hfill%
\hfill%
\hfill%
\includegraphics[width=0.087\linewidth]{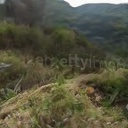}%
\hfill%
\includegraphics[width=0.087\linewidth]{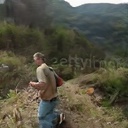}%
\hfill%
\hfill%
\hfill%
\includegraphics[width=0.087\linewidth]{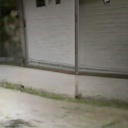}%
\hfill%
\includegraphics[width=0.087\linewidth]{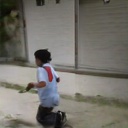}%
\hfill%
\hfill%
\hfill%
\includegraphics[width=0.087\linewidth]{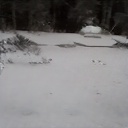}%
\hfill%
\includegraphics[width=0.087\linewidth]{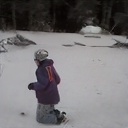}%
\hfill%
\hfill%
\hfill%
\includegraphics[width=0.087\linewidth]{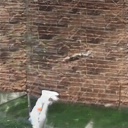}%
\hfill%
\includegraphics[width=0.087\linewidth]{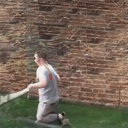}%
\hfill%
\hfill%
\hfill%
\includegraphics[width=0.087\linewidth]{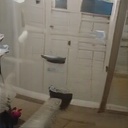}%
\hfill%
\includegraphics[width=0.087\linewidth]{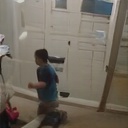}%
\end{subfigure}

% \vspace{0.9pt}

\begin{subfigure}[b]{1.0\linewidth}
\textoverlay{K1}{\includegraphics[width=0.087\linewidth]{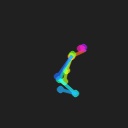}}%
\hfill%
\hfill%
\hfill%
\includegraphics[width=0.087\linewidth]{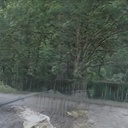}%
\hfill%
\includegraphics[width=0.087\linewidth]{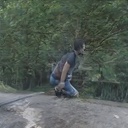}%
\hfill%
\hfill%
\hfill%
\includegraphics[width=0.087\linewidth]{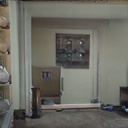}%
\hfill%
\includegraphics[width=0.087\linewidth]{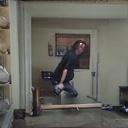}%
\hfill%
\hfill%
\hfill%
\includegraphics[width=0.087\linewidth]{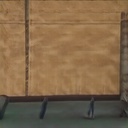}%
\hfill%
\includegraphics[width=0.087\linewidth]{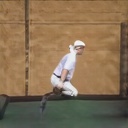}%
\hfill%
\hfill%
\hfill%
\includegraphics[width=0.087\linewidth]{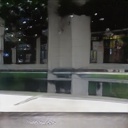}%
\hfill%
\includegraphics[width=0.087\linewidth]{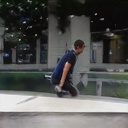}%
\hfill%
\hfill%
\hfill%
\includegraphics[width=0.087\linewidth]{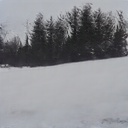}%
\hfill%
\includegraphics[width=0.087\linewidth]{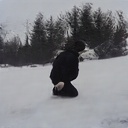}%
\end{subfigure}

% \vspace{0.9pt}

\begin{subfigure}[b]{1.0\linewidth}
\textoverlay{L1}{\includegraphics[width=0.087\linewidth]{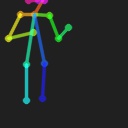}}%
\hfill%
\hfill%
\hfill%
\includegraphics[width=0.087\linewidth]{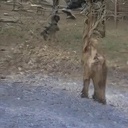}%
\hfill%
\includegraphics[width=0.087\linewidth]{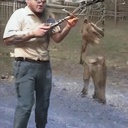}%
\hfill%
\hfill%
\hfill%
\includegraphics[width=0.087\linewidth]{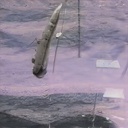}%
\hfill%
\includegraphics[width=0.087\linewidth]{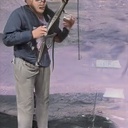}%
\hfill%
\hfill%
\hfill%
\includegraphics[width=0.087\linewidth]{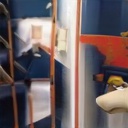}%
\hfill%
\includegraphics[width=0.087\linewidth]{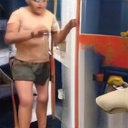}%
\hfill%
\hfill%
\hfill%
\includegraphics[width=0.087\linewidth]{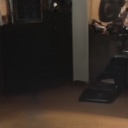}%
\hfill%
\includegraphics[width=0.087\linewidth]{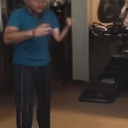}%
\hfill%
\hfill%
\hfill%
\includegraphics[width=0.087\linewidth]{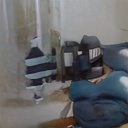}%
\hfill%
\includegraphics[width=0.087\linewidth]{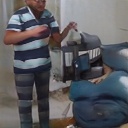}%
\end{subfigure}

% \vspace{0.9pt}

\begin{subfigure}[b]{1.0\linewidth}
\textoverlay{M1}{\includegraphics[width=0.087\linewidth]{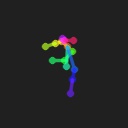}}%
\hfill%
\hfill%
\hfill%
\includegraphics[width=0.087\linewidth]{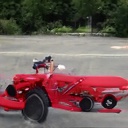}%
\hfill%
\includegraphics[width=0.087\linewidth]{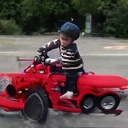}%
\hfill%
\hfill%
\hfill%
\includegraphics[width=0.087\linewidth]{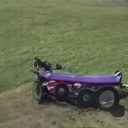}%
\hfill%
\includegraphics[width=0.087\linewidth]{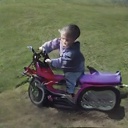}%
\hfill%
\hfill%
\hfill%
\includegraphics[width=0.087\linewidth]{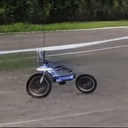}%
\hfill%
\includegraphics[width=0.087\linewidth]{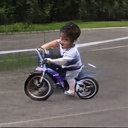}%
\hfill%
\hfill%
\hfill%
\includegraphics[width=0.087\linewidth]{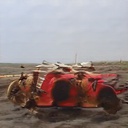}%
\hfill%
\includegraphics[width=0.087\linewidth]{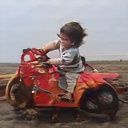}%
\hfill%
\hfill%
\hfill%
\includegraphics[width=0.087\linewidth]{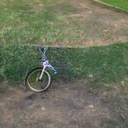}%
\hfill%
\includegraphics[width=0.087\linewidth]{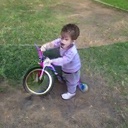}%
\end{subfigure}

% \vspace{0.9pt}

\begin{subfigure}[b]{1.0\linewidth}
\textoverlay{N1}{\includegraphics[width=0.087\linewidth]{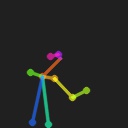}}%
\hfill%
\hfill%
\hfill%
\includegraphics[width=0.087\linewidth]{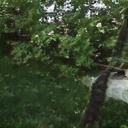}%
\hfill%
\includegraphics[width=0.087\linewidth]{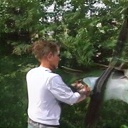}%
\hfill%
\hfill%
\hfill%
\includegraphics[width=0.087\linewidth]{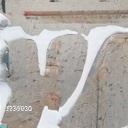}%
\hfill%
\includegraphics[width=0.087\linewidth]{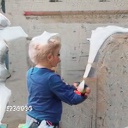}%
\hfill%
\hfill%
\hfill%
\includegraphics[width=0.087\linewidth]{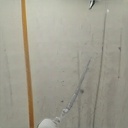}%
\hfill%
\includegraphics[width=0.087\linewidth]{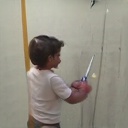}%
\hfill%
\hfill%
\hfill%
\includegraphics[width=0.087\linewidth]{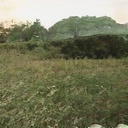}%
\hfill%
\includegraphics[width=0.087\linewidth]{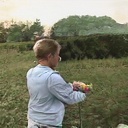}%
\hfill%
\hfill%
\hfill%
\includegraphics[width=0.087\linewidth]{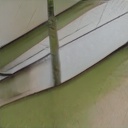}%
\hfill%
\includegraphics[width=0.087\linewidth]{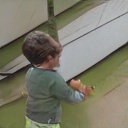}%
\end{subfigure}

% \vspace{0.9pt}

\begin{subfigure}[b]{1.0\linewidth}
\textoverlay{O1}{\includegraphics[width=0.087\linewidth]{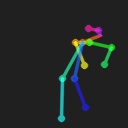}}%
\hfill%
\hfill%
\hfill%
\includegraphics[width=0.087\linewidth]{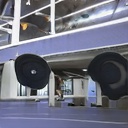}%
\hfill%
\includegraphics[width=0.087\linewidth]{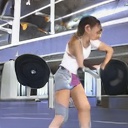}%
\hfill%
\hfill%
\hfill%
\includegraphics[width=0.087\linewidth]{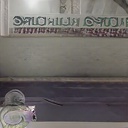}%
\hfill%
\includegraphics[width=0.087\linewidth]{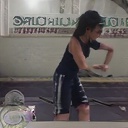}%
\hfill%
\hfill%
\hfill%
\includegraphics[width=0.087\linewidth]{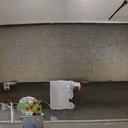}%
\hfill%
\includegraphics[width=0.087\linewidth]{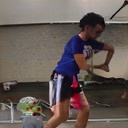}%
\hfill%
\hfill%
\hfill%
\includegraphics[width=0.087\linewidth]{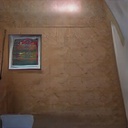}%
\hfill%
\includegraphics[width=0.087\linewidth]{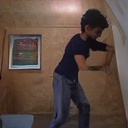}%
\hfill%
\hfill%
\hfill%
\includegraphics[width=0.087\linewidth]{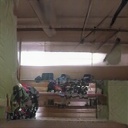}%
\hfill%
\includegraphics[width=0.087\linewidth]{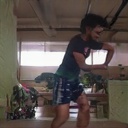}%
\end{subfigure}

% \vspace{-5pt}

\caption{\textbf{Random samples.}}
\label{fig:random_results_0}

\end{figure*}

\begin{figure*}[h!]
\centering

\begin{subfigure}[b]{1.0\linewidth}
\textoverlay{P1}{\includegraphics[width=0.087\linewidth]{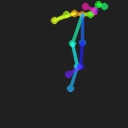}}%
\hfill%
\hfill%
\hfill%
\includegraphics[width=0.087\linewidth]{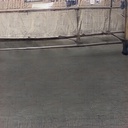}%
\hfill%
\includegraphics[width=0.087\linewidth]{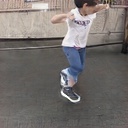}%
\hfill%
\hfill%
\hfill%
\includegraphics[width=0.087\linewidth]{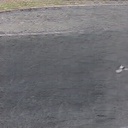}%
\hfill%
\includegraphics[width=0.087\linewidth]{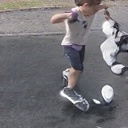}%
\hfill%
\hfill%
\hfill%
\includegraphics[width=0.087\linewidth]{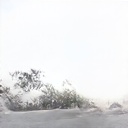}%
\hfill%
\includegraphics[width=0.087\linewidth]{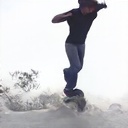}%
\hfill%
\hfill%
\hfill%
\includegraphics[width=0.087\linewidth]{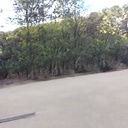}%
\hfill%
\includegraphics[width=0.087\linewidth]{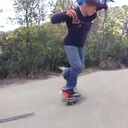}%
\hfill%
\hfill%
\hfill%
\includegraphics[width=0.087\linewidth]{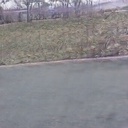}%
\hfill%
\includegraphics[width=0.087\linewidth]{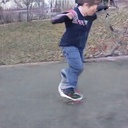}%
\end{subfigure}

% \vspace{0.9pt}

\begin{subfigure}[b]{1.0\linewidth}
\textoverlay{Q1}{\includegraphics[width=0.087\linewidth]{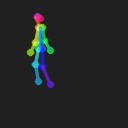}}%
\hfill%
\hfill%
\hfill%
\includegraphics[width=0.087\linewidth]{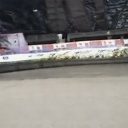}%
\hfill%
\includegraphics[width=0.087\linewidth]{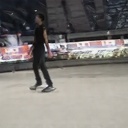}%
\hfill%
\hfill%
\hfill%
\includegraphics[width=0.087\linewidth]{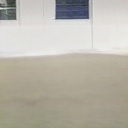}%
\hfill%
\includegraphics[width=0.087\linewidth]{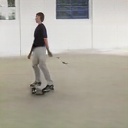}%
\hfill%
\hfill%
\hfill%
\includegraphics[width=0.087\linewidth]{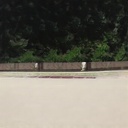}%
\hfill%
\includegraphics[width=0.087\linewidth]{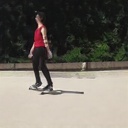}%
\hfill%
\hfill%
\hfill%
\includegraphics[width=0.087\linewidth]{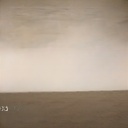}%
\hfill%
\includegraphics[width=0.087\linewidth]{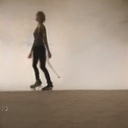}%
\hfill%
\hfill%
\hfill%
\includegraphics[width=0.087\linewidth]{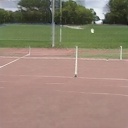}%
\hfill%
\includegraphics[width=0.087\linewidth]{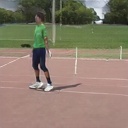}%
\end{subfigure}

% \vspace{0.9pt}

\begin{subfigure}[b]{1.0\linewidth}
\textoverlay{R1}{\includegraphics[width=0.087\linewidth]{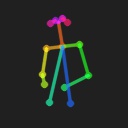}}%
\hfill%
\hfill%
\hfill%
\includegraphics[width=0.087\linewidth]{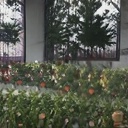}%
\hfill%
\includegraphics[width=0.087\linewidth]{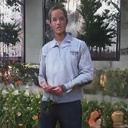}%
\hfill%
\hfill%
\hfill%
\includegraphics[width=0.087\linewidth]{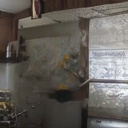}%
\hfill%
\includegraphics[width=0.087\linewidth]{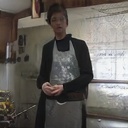}%
\hfill%
\hfill%
\hfill%
\includegraphics[width=0.087\linewidth]{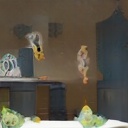}%
\hfill%
\includegraphics[width=0.087\linewidth]{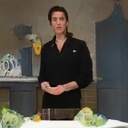}%
\hfill%
\hfill%
\hfill%
\includegraphics[width=0.087\linewidth]{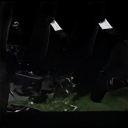}%
\hfill%
\includegraphics[width=0.087\linewidth]{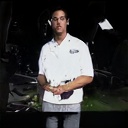}%
\hfill%
\hfill%
\hfill%
\includegraphics[width=0.087\linewidth]{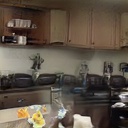}%
\hfill%
\includegraphics[width=0.087\linewidth]{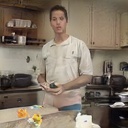}%
\end{subfigure}

% \vspace{0.9pt}

\begin{subfigure}[b]{1.0\linewidth}
\textoverlay{S1}{\includegraphics[width=0.087\linewidth]{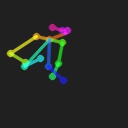}}%
\hfill%
\hfill%
\hfill%
\includegraphics[width=0.087\linewidth]{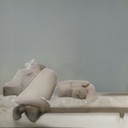}%
\hfill%
\includegraphics[width=0.087\linewidth]{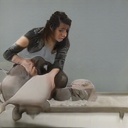}%
\hfill%
\hfill%
\hfill%
\includegraphics[width=0.087\linewidth]{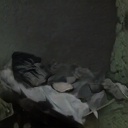}%
\hfill%
\includegraphics[width=0.087\linewidth]{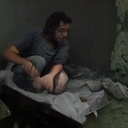}%
\hfill%
\hfill%
\hfill%
\includegraphics[width=0.087\linewidth]{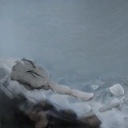}%
\hfill%
\includegraphics[width=0.087\linewidth]{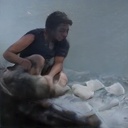}%
\hfill%
\hfill%
\hfill%
\includegraphics[width=0.087\linewidth]{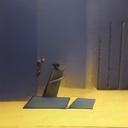}%
\hfill%
\includegraphics[width=0.087\linewidth]{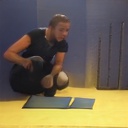}%
\hfill%
\hfill%
\hfill%
\includegraphics[width=0.087\linewidth]{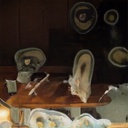}%
\hfill%
\includegraphics[width=0.087\linewidth]{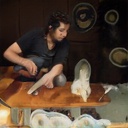}%
\end{subfigure}

% \vspace{0.9pt}

\begin{subfigure}[b]{1.0\linewidth}
\textoverlay{T1}{\includegraphics[width=0.087\linewidth]{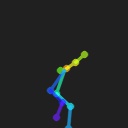}}%
\hfill%
\hfill%
\hfill%
\includegraphics[width=0.087\linewidth]{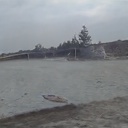}%
\hfill%
\includegraphics[width=0.087\linewidth]{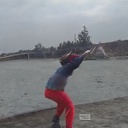}%
\hfill%
\hfill%
\hfill%
\includegraphics[width=0.087\linewidth]{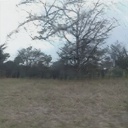}%
\hfill%
\includegraphics[width=0.087\linewidth]{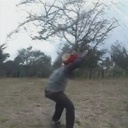}%
\hfill%
\hfill%
\hfill%
\includegraphics[width=0.087\linewidth]{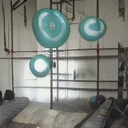}%
\hfill%
\includegraphics[width=0.087\linewidth]{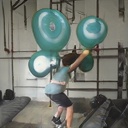}%
\hfill%
\hfill%
\hfill%
\includegraphics[width=0.087\linewidth]{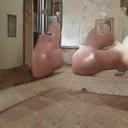}%
\hfill%
\includegraphics[width=0.087\linewidth]{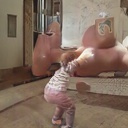}%
\hfill%
\hfill%
\hfill%
\includegraphics[width=0.087\linewidth]{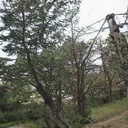}%
\hfill%
\includegraphics[width=0.087\linewidth]{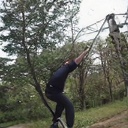}%
\end{subfigure}

% \vspace{0.9pt}

\begin{subfigure}[b]{1.0\linewidth}
\textoverlay{U1}{\includegraphics[width=0.087\linewidth]{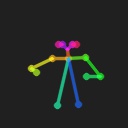}}%
\hfill%
\hfill%
\hfill%
\includegraphics[width=0.087\linewidth]{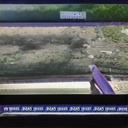}%
\hfill%
\includegraphics[width=0.087\linewidth]{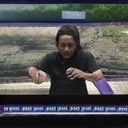}%
\hfill%
\hfill%
\hfill%
\includegraphics[width=0.087\linewidth]{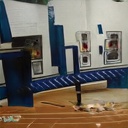}%
\hfill%
\includegraphics[width=0.087\linewidth]{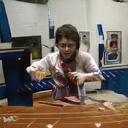}%
\hfill%
\hfill%
\hfill%
\includegraphics[width=0.087\linewidth]{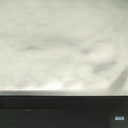}%
\hfill%
\includegraphics[width=0.087\linewidth]{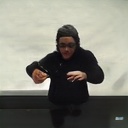}%
\hfill%
\hfill%
\hfill%
\includegraphics[width=0.087\linewidth]{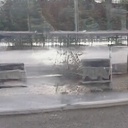}%
\hfill%
\includegraphics[width=0.087\linewidth]{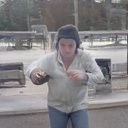}%
\hfill%
\hfill%
\hfill%
\includegraphics[width=0.087\linewidth]{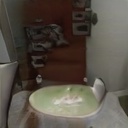}%
\hfill%
\includegraphics[width=0.087\linewidth]{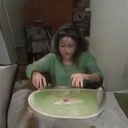}%
\end{subfigure}

% \vspace{0.9pt}

\begin{subfigure}[b]{1.0\linewidth}
\textoverlay{V1}{\includegraphics[width=0.087\linewidth]{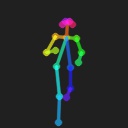}}%
\hfill%
\hfill%
\hfill%
\includegraphics[width=0.087\linewidth]{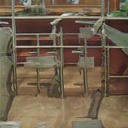}%
\hfill%
\includegraphics[width=0.087\linewidth]{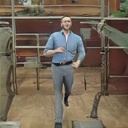}%
\hfill%
\hfill%
\hfill%
\includegraphics[width=0.087\linewidth]{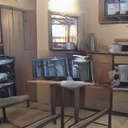}%
\hfill%
\includegraphics[width=0.087\linewidth]{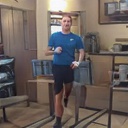}%
\hfill%
\hfill%
\hfill%
\includegraphics[width=0.087\linewidth]{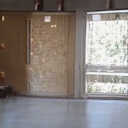}%
\hfill%
\includegraphics[width=0.087\linewidth]{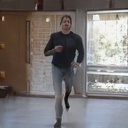}%
\hfill%
\hfill%
\hfill%
\includegraphics[width=0.087\linewidth]{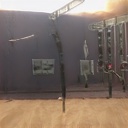}%
\hfill%
\includegraphics[width=0.087\linewidth]{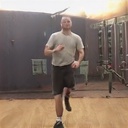}%
\hfill%
\hfill%
\hfill%
\includegraphics[width=0.087\linewidth]{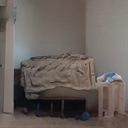}%
\hfill%
\includegraphics[width=0.087\linewidth]{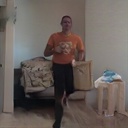}%
\end{subfigure}

% \vspace{0.9pt}

\begin{subfigure}[b]{1.0\linewidth}
\textoverlay{W1}{\includegraphics[width=0.087\linewidth]{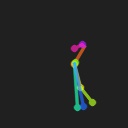}}%
\hfill%
\hfill%
\hfill%
\includegraphics[width=0.087\linewidth]{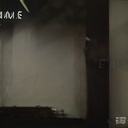}%
\hfill%
\includegraphics[width=0.087\linewidth]{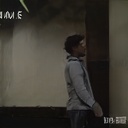}%
\hfill%
\hfill%
\hfill%
\includegraphics[width=0.087\linewidth]{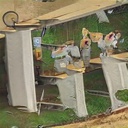}%
\hfill%
\includegraphics[width=0.087\linewidth]{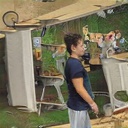}%
\hfill%
\hfill%
\hfill%
\includegraphics[width=0.087\linewidth]{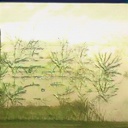}%
\hfill%
\includegraphics[width=0.087\linewidth]{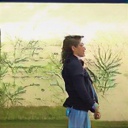}%
\hfill%
\hfill%
\hfill%
\includegraphics[width=0.087\linewidth]{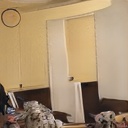}%
\hfill%
\includegraphics[width=0.087\linewidth]{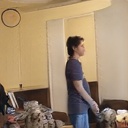}%
\hfill%
\hfill%
\hfill%
\includegraphics[width=0.087\linewidth]{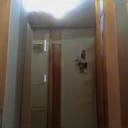}%
\hfill%
\includegraphics[width=0.087\linewidth]{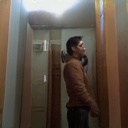}%
\end{subfigure}

% \vspace{0.9pt}

\begin{subfigure}[b]{1.0\linewidth}
\textoverlay{X1}{\includegraphics[width=0.087\linewidth]{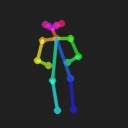}}%
\hfill%
\hfill%
\hfill%
\includegraphics[width=0.087\linewidth]{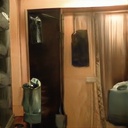}%
\hfill%
\includegraphics[width=0.087\linewidth]{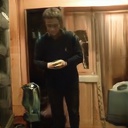}%
\hfill%
\hfill%
\hfill%
\includegraphics[width=0.087\linewidth]{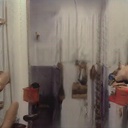}%
\hfill%
\includegraphics[width=0.087\linewidth]{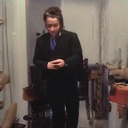}%
\hfill%
\hfill%
\hfill%
\includegraphics[width=0.087\linewidth]{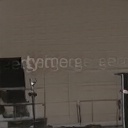}%
\hfill%
\includegraphics[width=0.087\linewidth]{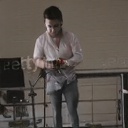}%
\hfill%
\hfill%
\hfill%
\includegraphics[width=0.087\linewidth]{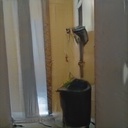}%
\hfill%
\includegraphics[width=0.087\linewidth]{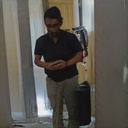}%
\hfill%
\hfill%
\hfill%
\includegraphics[width=0.087\linewidth]{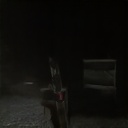}%
\hfill%
\includegraphics[width=0.087\linewidth]{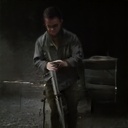}%
\end{subfigure}

% \vspace{0.9pt}

\begin{subfigure}[b]{1.0\linewidth}
\textoverlay{Y1}{\includegraphics[width=0.087\linewidth]{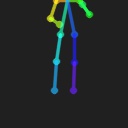}}%
\hfill%
\hfill%
\hfill%
\includegraphics[width=0.087\linewidth]{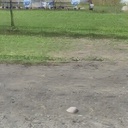}%
\hfill%
\includegraphics[width=0.087\linewidth]{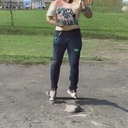}%
\hfill%
\hfill%
\hfill%
\includegraphics[width=0.087\linewidth]{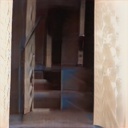}%
\hfill%
\includegraphics[width=0.087\linewidth]{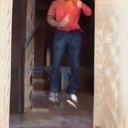}%
\hfill%
\hfill%
\hfill%
\includegraphics[width=0.087\linewidth]{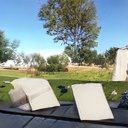}%
\hfill%
\includegraphics[width=0.087\linewidth]{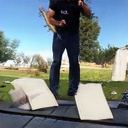}%
\hfill%
\hfill%
\hfill%
\includegraphics[width=0.087\linewidth]{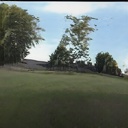}%
\hfill%
\includegraphics[width=0.087\linewidth]{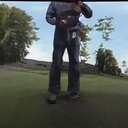}%
\hfill%
\hfill%
\hfill%
\includegraphics[width=0.087\linewidth]{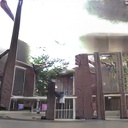}%
\hfill%
\includegraphics[width=0.087\linewidth]{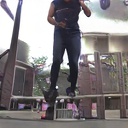}%
\end{subfigure}

% \vspace{0.9pt}

\begin{subfigure}[b]{1.0\linewidth}
\textoverlay{Z1}{\includegraphics[width=0.087\linewidth]{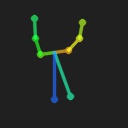}}%
\hfill%
\hfill%
\hfill%
\includegraphics[width=0.087\linewidth]{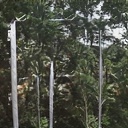}%
\hfill%
\includegraphics[width=0.087\linewidth]{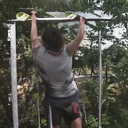}%
\hfill%
\hfill%
\hfill%
\includegraphics[width=0.087\linewidth]{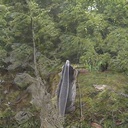}%
\hfill%
\includegraphics[width=0.087\linewidth]{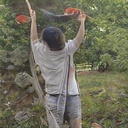}%
\hfill%
\hfill%
\hfill%
\includegraphics[width=0.087\linewidth]{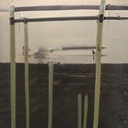}%
\hfill%
\includegraphics[width=0.087\linewidth]{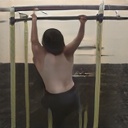}%
\hfill%
\hfill%
\hfill%
\includegraphics[width=0.087\linewidth]{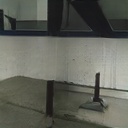}%
\hfill%
\includegraphics[width=0.087\linewidth]{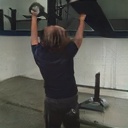}%
\hfill%
\hfill%
\hfill%
\includegraphics[width=0.087\linewidth]{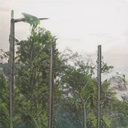}%
\hfill%
\includegraphics[width=0.087\linewidth]{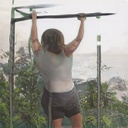}%
\end{subfigure}

% \vspace{0.9pt}

\begin{subfigure}[b]{1.0\linewidth}
\textoverlay{A2}{\includegraphics[width=0.087\linewidth]{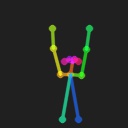}}%
\hfill%
\hfill%
\hfill%
\includegraphics[width=0.087\linewidth]{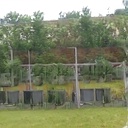}%
\hfill%
\includegraphics[width=0.087\linewidth]{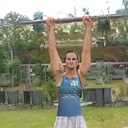}%
\hfill%
\hfill%
\hfill%
\includegraphics[width=0.087\linewidth]{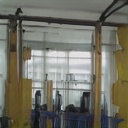}%
\hfill%
\includegraphics[width=0.087\linewidth]{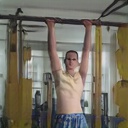}%
\hfill%
\hfill%
\hfill%
\includegraphics[width=0.087\linewidth]{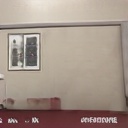}%
\hfill%
\includegraphics[width=0.087\linewidth]{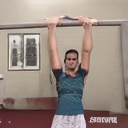}%
\hfill%
\hfill%
\hfill%
\includegraphics[width=0.087\linewidth]{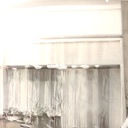}%
\hfill%
\includegraphics[width=0.087\linewidth]{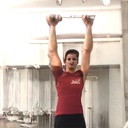}%
\hfill%
\hfill%
\hfill%
\includegraphics[width=0.087\linewidth]{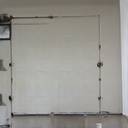}%
\hfill%
\includegraphics[width=0.087\linewidth]{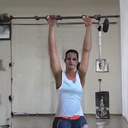}%
\end{subfigure}

% \vspace{0.9pt}

\begin{subfigure}[b]{1.0\linewidth}
\textoverlay{B2}{\includegraphics[width=0.087\linewidth]{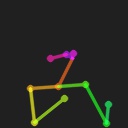}}%
\hfill%
\hfill%
\hfill%
\includegraphics[width=0.087\linewidth]{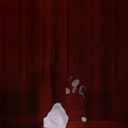}%
\hfill%
\includegraphics[width=0.087\linewidth]{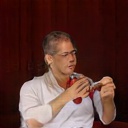}%
\hfill%
\hfill%
\hfill%
\includegraphics[width=0.087\linewidth]{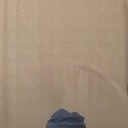}%
\hfill%
\includegraphics[width=0.087\linewidth]{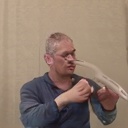}%
\hfill%
\hfill%
\hfill%
\includegraphics[width=0.087\linewidth]{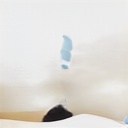}%
\hfill%
\includegraphics[width=0.087\linewidth]{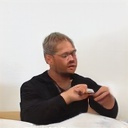}%
\hfill%
\hfill%
\hfill%
\includegraphics[width=0.087\linewidth]{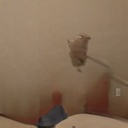}%
\hfill%
\includegraphics[width=0.087\linewidth]{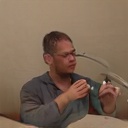}%
\hfill%
\hfill%
\hfill%
\includegraphics[width=0.087\linewidth]{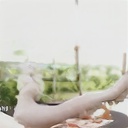}%
\hfill%
\includegraphics[width=0.087\linewidth]{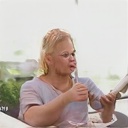}%
\end{subfigure}

% \vspace{0.9pt}

\begin{subfigure}[b]{1.0\linewidth}
\textoverlay{C2}{\includegraphics[width=0.087\linewidth]{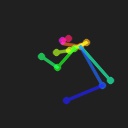}}%
\hfill%
\hfill%
\hfill%
\includegraphics[width=0.087\linewidth]{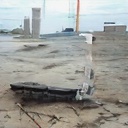}%
\hfill%
\includegraphics[width=0.087\linewidth]{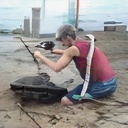}%
\hfill%
\hfill%
\hfill%
\includegraphics[width=0.087\linewidth]{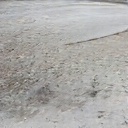}%
\hfill%
\includegraphics[width=0.087\linewidth]{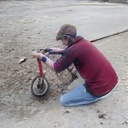}%
\hfill%
\hfill%
\hfill%
\includegraphics[width=0.087\linewidth]{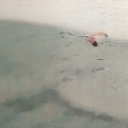}%
\hfill%
\includegraphics[width=0.087\linewidth]{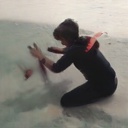}%
\hfill%
\hfill%
\hfill%
\includegraphics[width=0.087\linewidth]{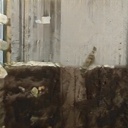}%
\hfill%
\includegraphics[width=0.087\linewidth]{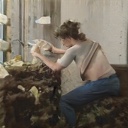}%
\hfill%
\hfill%
\hfill%
\includegraphics[width=0.087\linewidth]{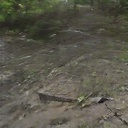}%
\hfill%
\includegraphics[width=0.087\linewidth]{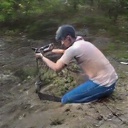}%
\end{subfigure}

% \vspace{0.9pt}

\begin{subfigure}[b]{1.0\linewidth}
\textoverlay{D2}{\includegraphics[width=0.087\linewidth]{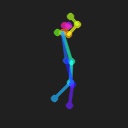}}%
\hfill%
\hfill%
\hfill%
\includegraphics[width=0.087\linewidth]{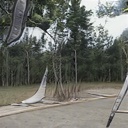}%
\hfill%
\includegraphics[width=0.087\linewidth]{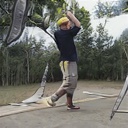}%
\hfill%
\hfill%
\hfill%
\includegraphics[width=0.087\linewidth]{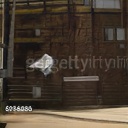}%
\hfill%
\includegraphics[width=0.087\linewidth]{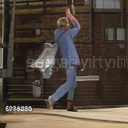}%
\hfill%
\hfill%
\hfill%
\includegraphics[width=0.087\linewidth]{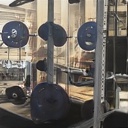}%
\hfill%
\includegraphics[width=0.087\linewidth]{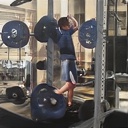}%
\hfill%
\hfill%
\hfill%
\includegraphics[width=0.087\linewidth]{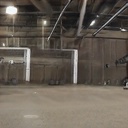}%
\hfill%
\includegraphics[width=0.087\linewidth]{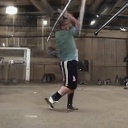}%
\hfill%
\hfill%
\hfill%
\includegraphics[width=0.087\linewidth]{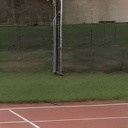}%
\hfill%
\includegraphics[width=0.087\linewidth]{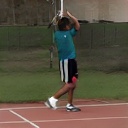}%
\end{subfigure}

% \vspace{-5pt}

\caption{\textbf{Random samples.}}
\label{fig:random_results_1}

\end{figure*}

\begin{figure*}[h!]
\centering

\begin{subfigure}[b]{1.0\linewidth}
\textoverlay{E2}{\includegraphics[width=0.087\linewidth]{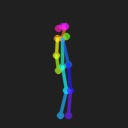}}%
\hfill%
\hfill%
\hfill%
\includegraphics[width=0.087\linewidth]{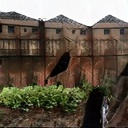}%
\hfill%
\includegraphics[width=0.087\linewidth]{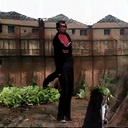}%
\hfill%
\hfill%
\hfill%
\includegraphics[width=0.087\linewidth]{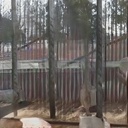}%
\hfill%
\includegraphics[width=0.087\linewidth]{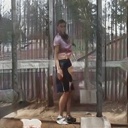}%
\hfill%
\hfill%
\hfill%
\includegraphics[width=0.087\linewidth]{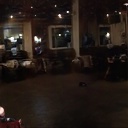}%
\hfill%
\includegraphics[width=0.087\linewidth]{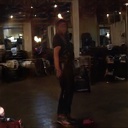}%
\hfill%
\hfill%
\hfill%
\includegraphics[width=0.087\linewidth]{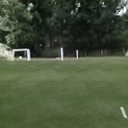}%
\hfill%
\includegraphics[width=0.087\linewidth]{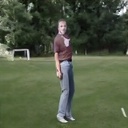}%
\hfill%
\hfill%
\hfill%
\includegraphics[width=0.087\linewidth]{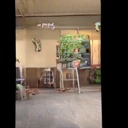}%
\hfill%
\includegraphics[width=0.087\linewidth]{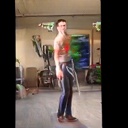}%
\end{subfigure}

% \vspace{0.9pt}

\begin{subfigure}[b]{1.0\linewidth}
\textoverlay{F2}{\includegraphics[width=0.087\linewidth]{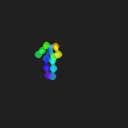}}%
\hfill%
\hfill%
\hfill%
\includegraphics[width=0.087\linewidth]{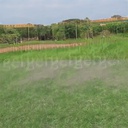}%
\hfill%
\includegraphics[width=0.087\linewidth]{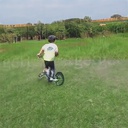}%
\hfill%
\hfill%
\hfill%
\includegraphics[width=0.087\linewidth]{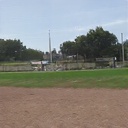}%
\hfill%
\includegraphics[width=0.087\linewidth]{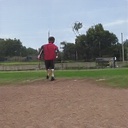}%
\hfill%
\hfill%
\hfill%
\includegraphics[width=0.087\linewidth]{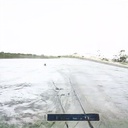}%
\hfill%
\includegraphics[width=0.087\linewidth]{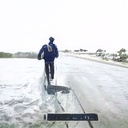}%
\hfill%
\hfill%
\hfill%
\includegraphics[width=0.087\linewidth]{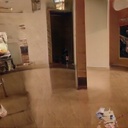}%
\hfill%
\includegraphics[width=0.087\linewidth]{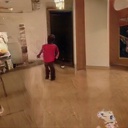}%
\hfill%
\hfill%
\hfill%
\includegraphics[width=0.087\linewidth]{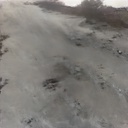}%
\hfill%
\includegraphics[width=0.087\linewidth]{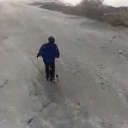}%
\end{subfigure}

% \vspace{0.9pt}

\begin{subfigure}[b]{1.0\linewidth}
\textoverlay{G2}{\includegraphics[width=0.087\linewidth]{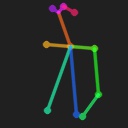}}%
\hfill%
\hfill%
\hfill%
\includegraphics[width=0.087\linewidth]{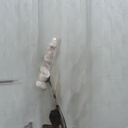}%
\hfill%
\includegraphics[width=0.087\linewidth]{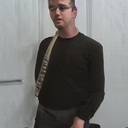}%
\hfill%
\hfill%
\hfill%
\includegraphics[width=0.087\linewidth]{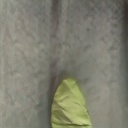}%
\hfill%
\includegraphics[width=0.087\linewidth]{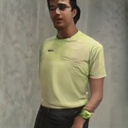}%
\hfill%
\hfill%
\hfill%
\includegraphics[width=0.087\linewidth]{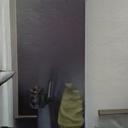}%
\hfill%
\includegraphics[width=0.087\linewidth]{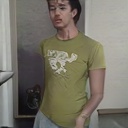}%
\hfill%
\hfill%
\hfill%
\includegraphics[width=0.087\linewidth]{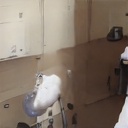}%
\hfill%
\includegraphics[width=0.087\linewidth]{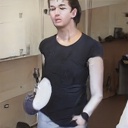}%
\hfill%
\hfill%
\hfill%
\includegraphics[width=0.087\linewidth]{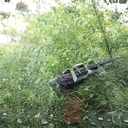}%
\hfill%
\includegraphics[width=0.087\linewidth]{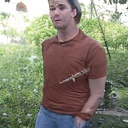}%
\end{subfigure}

% \vspace{0.9pt}

\begin{subfigure}[b]{1.0\linewidth}
\textoverlay{H2}{\includegraphics[width=0.087\linewidth]{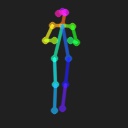}}%
\hfill%
\hfill%
\hfill%
\includegraphics[width=0.087\linewidth]{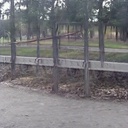}%
\hfill%
\includegraphics[width=0.087\linewidth]{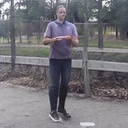}%
\hfill%
\hfill%
\hfill%
\includegraphics[width=0.087\linewidth]{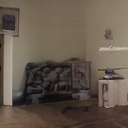}%
\hfill%
\includegraphics[width=0.087\linewidth]{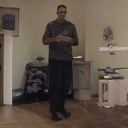}%
\hfill%
\hfill%
\hfill%
\includegraphics[width=0.087\linewidth]{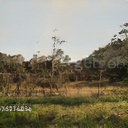}%
\hfill%
\includegraphics[width=0.087\linewidth]{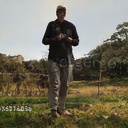}%
\hfill%
\hfill%
\hfill%
\includegraphics[width=0.087\linewidth]{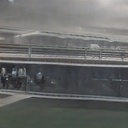}%
\hfill%
\includegraphics[width=0.087\linewidth]{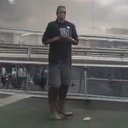}%
\hfill%
\hfill%
\hfill%
\includegraphics[width=0.087\linewidth]{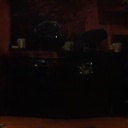}%
\hfill%
\includegraphics[width=0.087\linewidth]{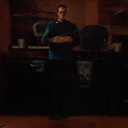}%
\end{subfigure}

% \vspace{0.9pt}

\begin{subfigure}[b]{1.0\linewidth}
\textoverlay{I2}{\includegraphics[width=0.087\linewidth]{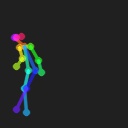}}%
\hfill%
\hfill%
\hfill%
\includegraphics[width=0.087\linewidth]{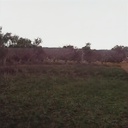}%
\hfill%
\includegraphics[width=0.087\linewidth]{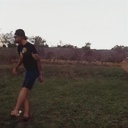}%
\hfill%
\hfill%
\hfill%
\includegraphics[width=0.087\linewidth]{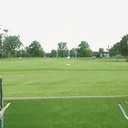}%
\hfill%
\includegraphics[width=0.087\linewidth]{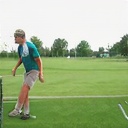}%
\hfill%
\hfill%
\hfill%
\includegraphics[width=0.087\linewidth]{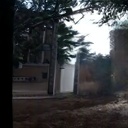}%
\hfill%
\includegraphics[width=0.087\linewidth]{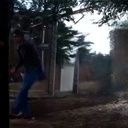}%
\hfill%
\hfill%
\hfill%
\includegraphics[width=0.087\linewidth]{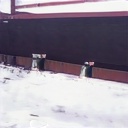}%
\hfill%
\includegraphics[width=0.087\linewidth]{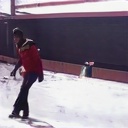}%
\hfill%
\hfill%
\hfill%
\includegraphics[width=0.087\linewidth]{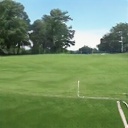}%
\hfill%
\includegraphics[width=0.087\linewidth]{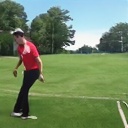}%
\end{subfigure}

% \vspace{0.9pt}

\begin{subfigure}[b]{1.0\linewidth}
\textoverlay{J2}{\includegraphics[width=0.087\linewidth]{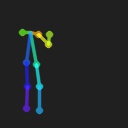}}%
\hfill%
\hfill%
\hfill%
\includegraphics[width=0.087\linewidth]{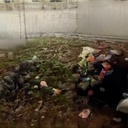}%
\hfill%
\includegraphics[width=0.087\linewidth]{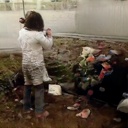}%
\hfill%
\hfill%
\hfill%
\includegraphics[width=0.087\linewidth]{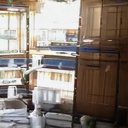}%
\hfill%
\includegraphics[width=0.087\linewidth]{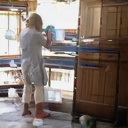}%
\hfill%
\hfill%
\hfill%
\includegraphics[width=0.087\linewidth]{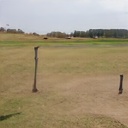}%
\hfill%
\includegraphics[width=0.087\linewidth]{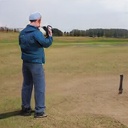}%
\hfill%
\hfill%
\hfill%
\includegraphics[width=0.087\linewidth]{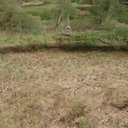}%
\hfill%
\includegraphics[width=0.087\linewidth]{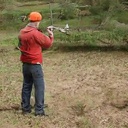}%
\hfill%
\hfill%
\hfill%
\includegraphics[width=0.087\linewidth]{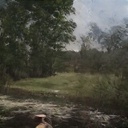}%
\hfill%
\includegraphics[width=0.087\linewidth]{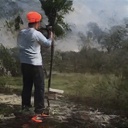}%
\end{subfigure}

% \vspace{0.9pt}

\begin{subfigure}[b]{1.0\linewidth}
\textoverlay{K2}{\includegraphics[width=0.087\linewidth]{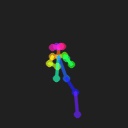}}%
\hfill%
\hfill%
\hfill%
\includegraphics[width=0.087\linewidth]{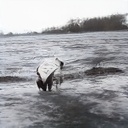}%
\hfill%
\includegraphics[width=0.087\linewidth]{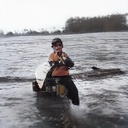}%
\hfill%
\hfill%
\hfill%
\includegraphics[width=0.087\linewidth]{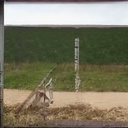}%
\hfill%
\includegraphics[width=0.087\linewidth]{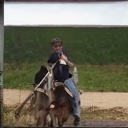}%
\hfill%
\hfill%
\hfill%
\includegraphics[width=0.087\linewidth]{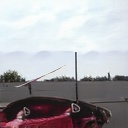}%
\hfill%
\includegraphics[width=0.087\linewidth]{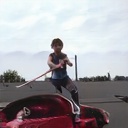}%
\hfill%
\hfill%
\hfill%
\includegraphics[width=0.087\linewidth]{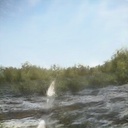}%
\hfill%
\includegraphics[width=0.087\linewidth]{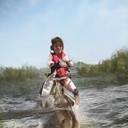}%
\hfill%
\hfill%
\hfill%
\includegraphics[width=0.087\linewidth]{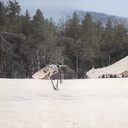}%
\hfill%
\includegraphics[width=0.087\linewidth]{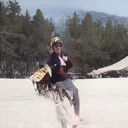}%
\end{subfigure}

% \vspace{0.9pt}

\begin{subfigure}[b]{1.0\linewidth}
\textoverlay{L2}{\includegraphics[width=0.087\linewidth]{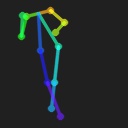}}%
\hfill%
\hfill%
\hfill%
\includegraphics[width=0.087\linewidth]{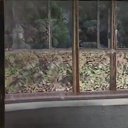}%
\hfill%
\includegraphics[width=0.087\linewidth]{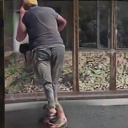}%
\hfill%
\hfill%
\hfill%
\includegraphics[width=0.087\linewidth]{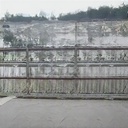}%
\hfill%
\includegraphics[width=0.087\linewidth]{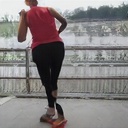}%
\hfill%
\hfill%
\hfill%
\includegraphics[width=0.087\linewidth]{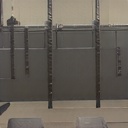}%
\hfill%
\includegraphics[width=0.087\linewidth]{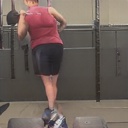}%
\hfill%
\hfill%
\hfill%
\includegraphics[width=0.087\linewidth]{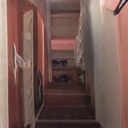}%
\hfill%
\includegraphics[width=0.087\linewidth]{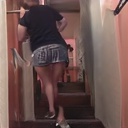}%
\hfill%
\hfill%
\hfill%
\includegraphics[width=0.087\linewidth]{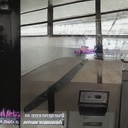}%
\hfill%
\includegraphics[width=0.087\linewidth]{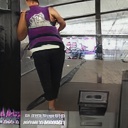}%
\end{subfigure}

% \vspace{0.9pt}

\begin{subfigure}[b]{1.0\linewidth}
\textoverlay{M2}{\includegraphics[width=0.087\linewidth]{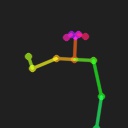}}%
\hfill%
\hfill%
\hfill%
\includegraphics[width=0.087\linewidth]{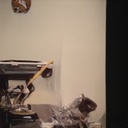}%
\hfill%
\includegraphics[width=0.087\linewidth]{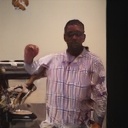}%
\hfill%
\hfill%
\hfill%
\includegraphics[width=0.087\linewidth]{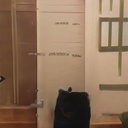}%
\hfill%
\includegraphics[width=0.087\linewidth]{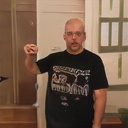}%
\hfill%
\hfill%
\hfill%
\includegraphics[width=0.087\linewidth]{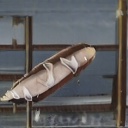}%
\hfill%
\includegraphics[width=0.087\linewidth]{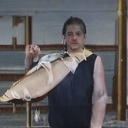}%
\hfill%
\hfill%
\hfill%
\includegraphics[width=0.087\linewidth]{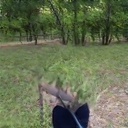}%
\hfill%
\includegraphics[width=0.087\linewidth]{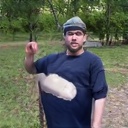}%
\hfill%
\hfill%
\hfill%
\includegraphics[width=0.087\linewidth]{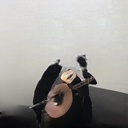}%
\hfill%
\includegraphics[width=0.087\linewidth]{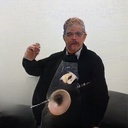}%
\end{subfigure}

% \vspace{0.9pt}

\begin{subfigure}[b]{1.0\linewidth}
\textoverlay{N2}{\includegraphics[width=0.087\linewidth]{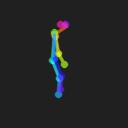}}%
\hfill%
\hfill%
\hfill%
\includegraphics[width=0.087\linewidth]{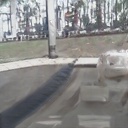}%
\hfill%
\includegraphics[width=0.087\linewidth]{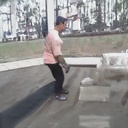}%
\hfill%
\hfill%
\hfill%
\includegraphics[width=0.087\linewidth]{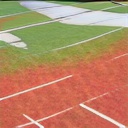}%
\hfill%
\includegraphics[width=0.087\linewidth]{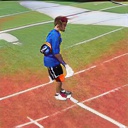}%
\hfill%
\hfill%
\hfill%
\includegraphics[width=0.087\linewidth]{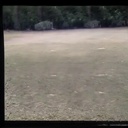}%
\hfill%
\includegraphics[width=0.087\linewidth]{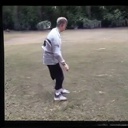}%
\hfill%
\hfill%
\hfill%
\includegraphics[width=0.087\linewidth]{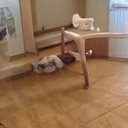}%
\hfill%
\includegraphics[width=0.087\linewidth]{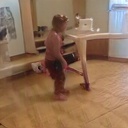}%
\hfill%
\hfill%
\hfill%
\includegraphics[width=0.087\linewidth]{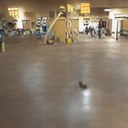}%
\hfill%
\includegraphics[width=0.087\linewidth]{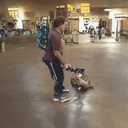}%
\end{subfigure}

% \vspace{0.9pt}

\begin{subfigure}[b]{1.0\linewidth}
\textoverlay{O2}{\includegraphics[width=0.087\linewidth]{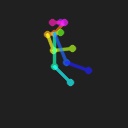}}%
\hfill%
\hfill%
\hfill%
\includegraphics[width=0.087\linewidth]{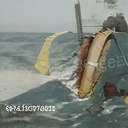}%
\hfill%
\includegraphics[width=0.087\linewidth]{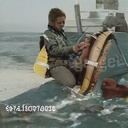}%
\hfill%
\hfill%
\hfill%
\includegraphics[width=0.087\linewidth]{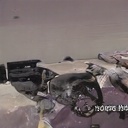}%
\hfill%
\includegraphics[width=0.087\linewidth]{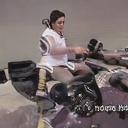}%
\hfill%
\hfill%
\hfill%
\includegraphics[width=0.087\linewidth]{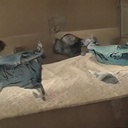}%
\hfill%
\includegraphics[width=0.087\linewidth]{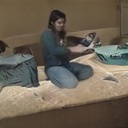}%
\hfill%
\hfill%
\hfill%
\includegraphics[width=0.087\linewidth]{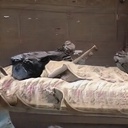}%
\hfill%
\includegraphics[width=0.087\linewidth]{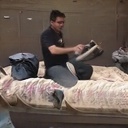}%
\hfill%
\hfill%
\hfill%
\includegraphics[width=0.087\linewidth]{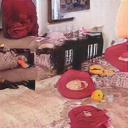}%
\hfill%
\includegraphics[width=0.087\linewidth]{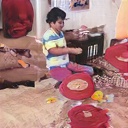}%
\end{subfigure}

% \vspace{0.9pt}

\begin{subfigure}[b]{1.0\linewidth}
\textoverlay{P2}{\includegraphics[width=0.087\linewidth]{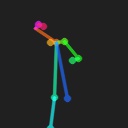}}%
\hfill%
\hfill%
\hfill%
\includegraphics[width=0.087\linewidth]{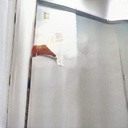}%
\hfill%
\includegraphics[width=0.087\linewidth]{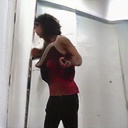}%
\hfill%
\hfill%
\hfill%
\includegraphics[width=0.087\linewidth]{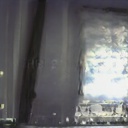}%
\hfill%
\includegraphics[width=0.087\linewidth]{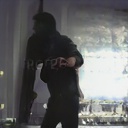}%
\hfill%
\hfill%
\hfill%
\includegraphics[width=0.087\linewidth]{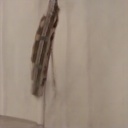}%
\hfill%
\includegraphics[width=0.087\linewidth]{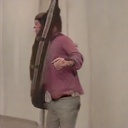}%
\hfill%
\hfill%
\hfill%
\includegraphics[width=0.087\linewidth]{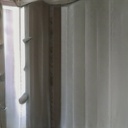}%
\hfill%
\includegraphics[width=0.087\linewidth]{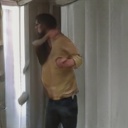}%
\hfill%
\hfill%
\hfill%
\includegraphics[width=0.087\linewidth]{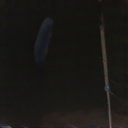}%
\hfill%
\includegraphics[width=0.087\linewidth]{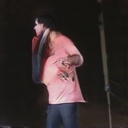}%
\end{subfigure}

% \vspace{0.9pt}

\begin{subfigure}[b]{1.0\linewidth}
\textoverlay{Q2}{\includegraphics[width=0.087\linewidth]{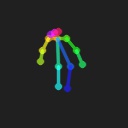}}%
\hfill%
\hfill%
\hfill%
\includegraphics[width=0.087\linewidth]{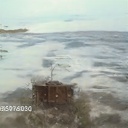}%
\hfill%
\includegraphics[width=0.087\linewidth]{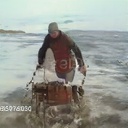}%
\hfill%
\hfill%
\hfill%
\includegraphics[width=0.087\linewidth]{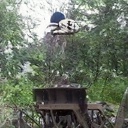}%
\hfill%
\includegraphics[width=0.087\linewidth]{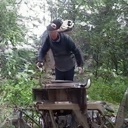}%
\hfill%
\hfill%
\hfill%
\includegraphics[width=0.087\linewidth]{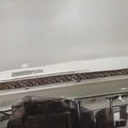}%
\hfill%
\includegraphics[width=0.087\linewidth]{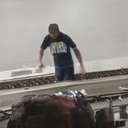}%
\hfill%
\hfill%
\hfill%
\includegraphics[width=0.087\linewidth]{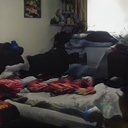}%
\hfill%
\includegraphics[width=0.087\linewidth]{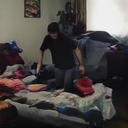}%
\hfill%
\hfill%
\hfill%
\includegraphics[width=0.087\linewidth]{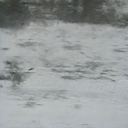}%
\hfill%
\includegraphics[width=0.087\linewidth]{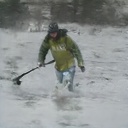}%
\end{subfigure}

% \vspace{0.9pt}

\begin{subfigure}[b]{1.0\linewidth}
\textoverlay{R2}{\includegraphics[width=0.087\linewidth]{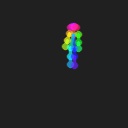}}%
\hfill%
\hfill%
\hfill%
\includegraphics[width=0.087\linewidth]{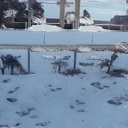}%
\hfill%
\includegraphics[width=0.087\linewidth]{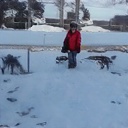}%
\hfill%
\hfill%
\hfill%
\includegraphics[width=0.087\linewidth]{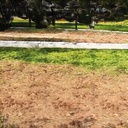}%
\hfill%
\includegraphics[width=0.087\linewidth]{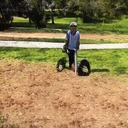}%
\hfill%
\hfill%
\hfill%
\includegraphics[width=0.087\linewidth]{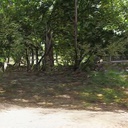}%
\hfill%
\includegraphics[width=0.087\linewidth]{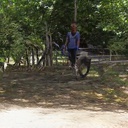}%
\hfill%
\hfill%
\hfill%
\includegraphics[width=0.087\linewidth]{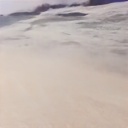}%
\hfill%
\includegraphics[width=0.087\linewidth]{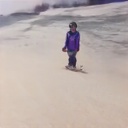}%
\hfill%
\hfill%
\hfill%
\includegraphics[width=0.087\linewidth]{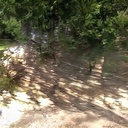}%
\hfill%
\includegraphics[width=0.087\linewidth]{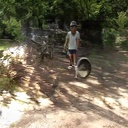}%
\end{subfigure}

% \vspace{0.9pt}

\begin{subfigure}[b]{1.0\linewidth}
\textoverlay{S2}{\includegraphics[width=0.087\linewidth]{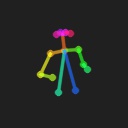}}%
\hfill%
\hfill%
\hfill%
\includegraphics[width=0.087\linewidth]{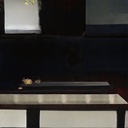}%
\hfill%
\includegraphics[width=0.087\linewidth]{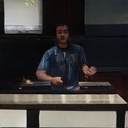}%
\hfill%
\hfill%
\hfill%
\includegraphics[width=0.087\linewidth]{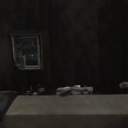}%
\hfill%
\includegraphics[width=0.087\linewidth]{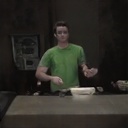}%
\hfill%
\hfill%
\hfill%
\includegraphics[width=0.087\linewidth]{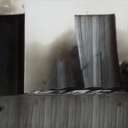}%
\hfill%
\includegraphics[width=0.087\linewidth]{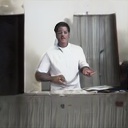}%
\hfill%
\hfill%
\hfill%
\includegraphics[width=0.087\linewidth]{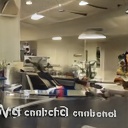}%
\hfill%
\includegraphics[width=0.087\linewidth]{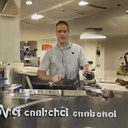}%
\hfill%
\hfill%
\hfill%
\includegraphics[width=0.087\linewidth]{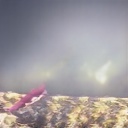}%
\hfill%
\includegraphics[width=0.087\linewidth]{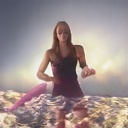}%
\end{subfigure}

% \vspace{-5pt}

\caption{\textbf{Random samples.}}
\label{fig:random_results_2}

\end{figure*}

\begin{figure*}[h!]
\centering

\begin{subfigure}[b]{1.0\linewidth}
\textoverlay{T2}{\includegraphics[width=0.087\linewidth]{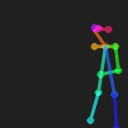}}%
\hfill%
\hfill%
\hfill%
\includegraphics[width=0.087\linewidth]{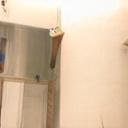}%
\hfill%
\includegraphics[width=0.087\linewidth]{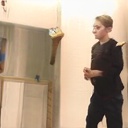}%
\hfill%
\hfill%
\hfill%
\includegraphics[width=0.087\linewidth]{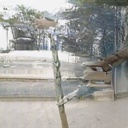}%
\hfill%
\includegraphics[width=0.087\linewidth]{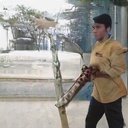}%
\hfill%
\hfill%
\hfill%
\includegraphics[width=0.087\linewidth]{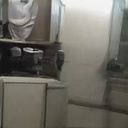}%
\hfill%
\includegraphics[width=0.087\linewidth]{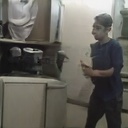}%
\hfill%
\hfill%
\hfill%
\includegraphics[width=0.087\linewidth]{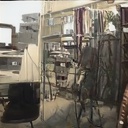}%
\hfill%
\includegraphics[width=0.087\linewidth]{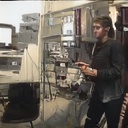}%
\hfill%
\hfill%
\hfill%
\includegraphics[width=0.087\linewidth]{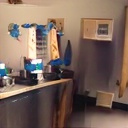}%
\hfill%
\includegraphics[width=0.087\linewidth]{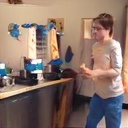}%
\end{subfigure}

% \vspace{0.9pt}

\begin{subfigure}[b]{1.0\linewidth}
\textoverlay{U2}{\includegraphics[width=0.087\linewidth]{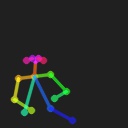}}%
\hfill%
\hfill%
\hfill%
\includegraphics[width=0.087\linewidth]{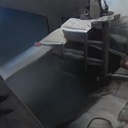}%
\hfill%
\includegraphics[width=0.087\linewidth]{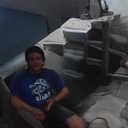}%
\hfill%
\hfill%
\hfill%
\includegraphics[width=0.087\linewidth]{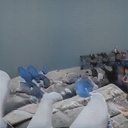}%
\hfill%
\includegraphics[width=0.087\linewidth]{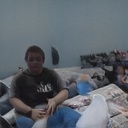}%
\hfill%
\hfill%
\hfill%
\includegraphics[width=0.087\linewidth]{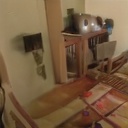}%
\hfill%
\includegraphics[width=0.087\linewidth]{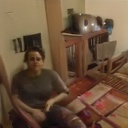}%
\hfill%
\hfill%
\hfill%
\includegraphics[width=0.087\linewidth]{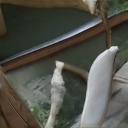}%
\hfill%
\includegraphics[width=0.087\linewidth]{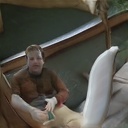}%
\hfill%
\hfill%
\hfill%
\includegraphics[width=0.087\linewidth]{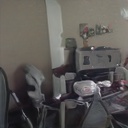}%
\hfill%
\includegraphics[width=0.087\linewidth]{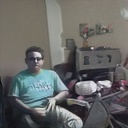}%
\end{subfigure}

% \vspace{0.9pt}

\begin{subfigure}[b]{1.0\linewidth}
\textoverlay{V2}{\includegraphics[width=0.087\linewidth]{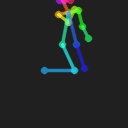}}%
\hfill%
\hfill%
\hfill%
\includegraphics[width=0.087\linewidth]{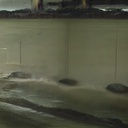}%
\hfill%
\includegraphics[width=0.087\linewidth]{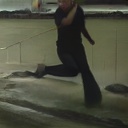}%
\hfill%
\hfill%
\hfill%
\includegraphics[width=0.087\linewidth]{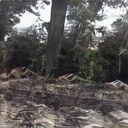}%
\hfill%
\includegraphics[width=0.087\linewidth]{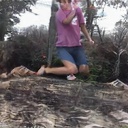}%
\hfill%
\hfill%
\hfill%
\includegraphics[width=0.087\linewidth]{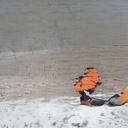}%
\hfill%
\includegraphics[width=0.087\linewidth]{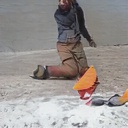}%
\hfill%
\hfill%
\hfill%
\includegraphics[width=0.087\linewidth]{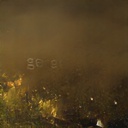}%
\hfill%
\includegraphics[width=0.087\linewidth]{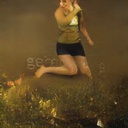}%
\hfill%
\hfill%
\hfill%
\includegraphics[width=0.087\linewidth]{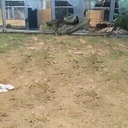}%
\hfill%
\includegraphics[width=0.087\linewidth]{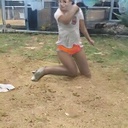}%
\end{subfigure}

% \vspace{0.9pt}

\begin{subfigure}[b]{1.0\linewidth}
\textoverlay{W2}{\includegraphics[width=0.087\linewidth]{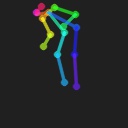}}%
\hfill%
\hfill%
\hfill%
\includegraphics[width=0.087\linewidth]{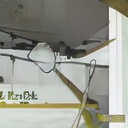}%
\hfill%
\includegraphics[width=0.087\linewidth]{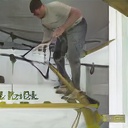}%
\hfill%
\hfill%
\hfill%
\includegraphics[width=0.087\linewidth]{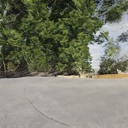}%
\hfill%
\includegraphics[width=0.087\linewidth]{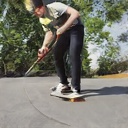}%
\hfill%
\hfill%
\hfill%
\includegraphics[width=0.087\linewidth]{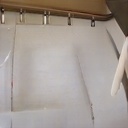}%
\hfill%
\includegraphics[width=0.087\linewidth]{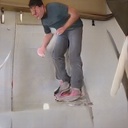}%
\hfill%
\hfill%
\hfill%
\includegraphics[width=0.087\linewidth]{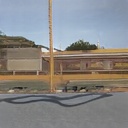}%
\hfill%
\includegraphics[width=0.087\linewidth]{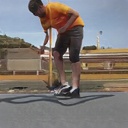}%
\hfill%
\hfill%
\hfill%
\includegraphics[width=0.087\linewidth]{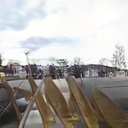}%
\hfill%
\includegraphics[width=0.087\linewidth]{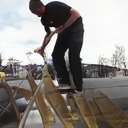}%
\end{subfigure}

% \vspace{0.9pt}

\begin{subfigure}[b]{1.0\linewidth}
\textoverlay{X2}{\includegraphics[width=0.087\linewidth]{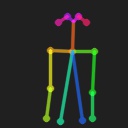}}%
\hfill%
\hfill%
\hfill%
\includegraphics[width=0.087\linewidth]{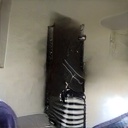}%
\hfill%
\includegraphics[width=0.087\linewidth]{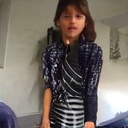}%
\hfill%
\hfill%
\hfill%
\includegraphics[width=0.087\linewidth]{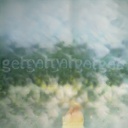}%
\hfill%
\includegraphics[width=0.087\linewidth]{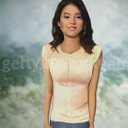}%
\hfill%
\hfill%
\hfill%
\includegraphics[width=0.087\linewidth]{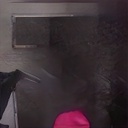}%
\hfill%
\includegraphics[width=0.087\linewidth]{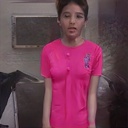}%
\hfill%
\hfill%
\hfill%
\includegraphics[width=0.087\linewidth]{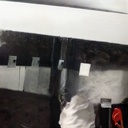}%
\hfill%
\includegraphics[width=0.087\linewidth]{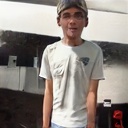}%
\hfill%
\hfill%
\hfill%
\includegraphics[width=0.087\linewidth]{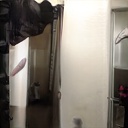}%
\hfill%
\includegraphics[width=0.087\linewidth]{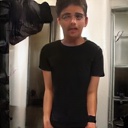}%
\end{subfigure}

% \vspace{0.9pt}

\begin{subfigure}[b]{1.0\linewidth}
\textoverlay{Y2}{\includegraphics[width=0.087\linewidth]{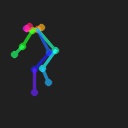}}%
\hfill%
\hfill%
\hfill%
\includegraphics[width=0.087\linewidth]{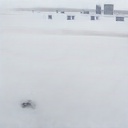}%
\hfill%
\includegraphics[width=0.087\linewidth]{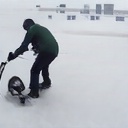}%
\hfill%
\hfill%
\hfill%
\includegraphics[width=0.087\linewidth]{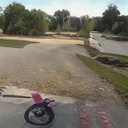}%
\hfill%
\includegraphics[width=0.087\linewidth]{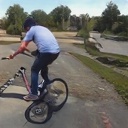}%
\hfill%
\hfill%
\hfill%
\includegraphics[width=0.087\linewidth]{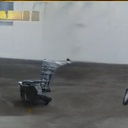}%
\hfill%
\includegraphics[width=0.087\linewidth]{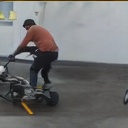}%
\hfill%
\hfill%
\hfill%
\includegraphics[width=0.087\linewidth]{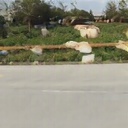}%
\hfill%
\includegraphics[width=0.087\linewidth]{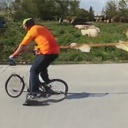}%
\hfill%
\hfill%
\hfill%
\includegraphics[width=0.087\linewidth]{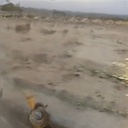}%
\hfill%
\includegraphics[width=0.087\linewidth]{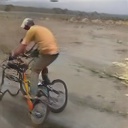}%
\end{subfigure}

% \vspace{0.9pt}

\begin{subfigure}[b]{1.0\linewidth}
\textoverlay{Z2}{\includegraphics[width=0.087\linewidth]{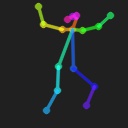}}%
\hfill%
\hfill%
\hfill%
\includegraphics[width=0.087\linewidth]{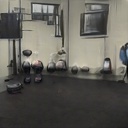}%
\hfill%
\includegraphics[width=0.087\linewidth]{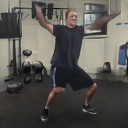}%
\hfill%
\hfill%
\hfill%
\includegraphics[width=0.087\linewidth]{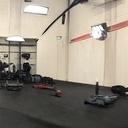}%
\hfill%
\includegraphics[width=0.087\linewidth]{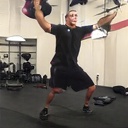}%
\hfill%
\hfill%
\hfill%
\includegraphics[width=0.087\linewidth]{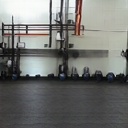}%
\hfill%
\includegraphics[width=0.087\linewidth]{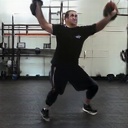}%
\hfill%
\hfill%
\hfill%
\includegraphics[width=0.087\linewidth]{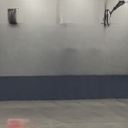}%
\hfill%
\includegraphics[width=0.087\linewidth]{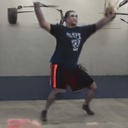}%
\hfill%
\hfill%
\hfill%
\includegraphics[width=0.087\linewidth]{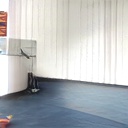}%
\hfill%
\includegraphics[width=0.087\linewidth]{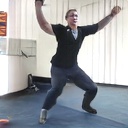}%
\end{subfigure}

% \vspace{0.9pt}

\begin{subfigure}[b]{1.0\linewidth}
\textoverlay{A3}{\includegraphics[width=0.087\linewidth]{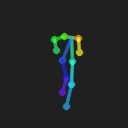}}%
\hfill%
\hfill%
\hfill%
\includegraphics[width=0.087\linewidth]{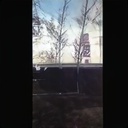}%
\hfill%
\includegraphics[width=0.087\linewidth]{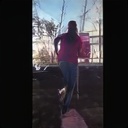}%
\hfill%
\hfill%
\hfill%
\includegraphics[width=0.087\linewidth]{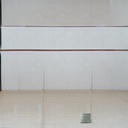}%
\hfill%
\includegraphics[width=0.087\linewidth]{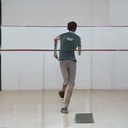}%
\hfill%
\hfill%
\hfill%
\includegraphics[width=0.087\linewidth]{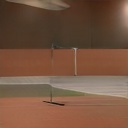}%
\hfill%
\includegraphics[width=0.087\linewidth]{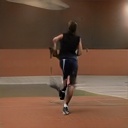}%
\hfill%
\hfill%
\hfill%
\includegraphics[width=0.087\linewidth]{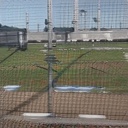}%
\hfill%
\includegraphics[width=0.087\linewidth]{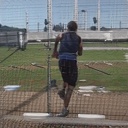}%
\hfill%
\hfill%
\hfill%
\includegraphics[width=0.087\linewidth]{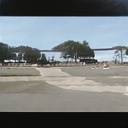}%
\hfill%
\includegraphics[width=0.087\linewidth]{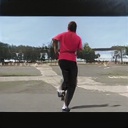}%
\end{subfigure}

% \vspace{0.9pt}

\begin{subfigure}[b]{1.0\linewidth}
\textoverlay{B3}{\includegraphics[width=0.087\linewidth]{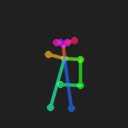}}%
\hfill%
\hfill%
\hfill%
\includegraphics[width=0.087\linewidth]{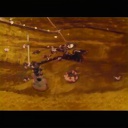}%
\hfill%
\includegraphics[width=0.087\linewidth]{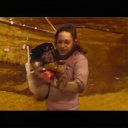}%
\hfill%
\hfill%
\hfill%
\includegraphics[width=0.087\linewidth]{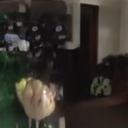}%
\hfill%
\includegraphics[width=0.087\linewidth]{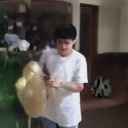}%
\hfill%
\hfill%
\hfill%
\includegraphics[width=0.087\linewidth]{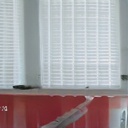}%
\hfill%
\includegraphics[width=0.087\linewidth]{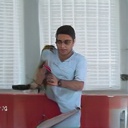}%
\hfill%
\hfill%
\hfill%
\includegraphics[width=0.087\linewidth]{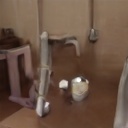}%
\hfill%
\includegraphics[width=0.087\linewidth]{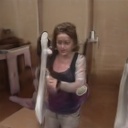}%
\hfill%
\hfill%
\hfill%
\includegraphics[width=0.087\linewidth]{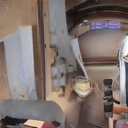}%
\hfill%
\includegraphics[width=0.087\linewidth]{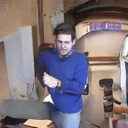}%
\end{subfigure}

% \vspace{0.9pt}

\begin{subfigure}[b]{1.0\linewidth}
\textoverlay{C3}{\includegraphics[width=0.087\linewidth]{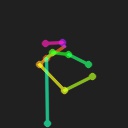}}%
\hfill%
\hfill%
\hfill%
\includegraphics[width=0.087\linewidth]{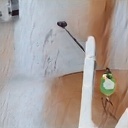}%
\hfill%
\includegraphics[width=0.087\linewidth]{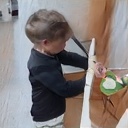}%
\hfill%
\hfill%
\hfill%
\includegraphics[width=0.087\linewidth]{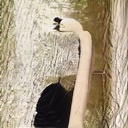}%
\hfill%
\includegraphics[width=0.087\linewidth]{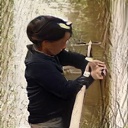}%
\hfill%
\hfill%
\hfill%
\includegraphics[width=0.087\linewidth]{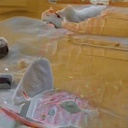}%
\hfill%
\includegraphics[width=0.087\linewidth]{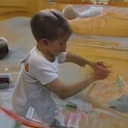}%
\hfill%
\hfill%
\hfill%
\includegraphics[width=0.087\linewidth]{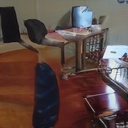}%
\hfill%
\includegraphics[width=0.087\linewidth]{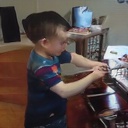}%
\hfill%
\hfill%
\hfill%
\includegraphics[width=0.087\linewidth]{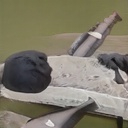}%
\hfill%
\includegraphics[width=0.087\linewidth]{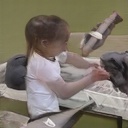}%
\end{subfigure}

% \vspace{0.9pt}

\begin{subfigure}[b]{1.0\linewidth}
\textoverlay{D3}{\includegraphics[width=0.087\linewidth]{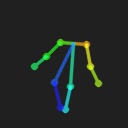}}%
\hfill%
\hfill%
\hfill%
\includegraphics[width=0.087\linewidth]{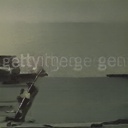}%
\hfill%
\includegraphics[width=0.087\linewidth]{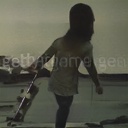}%
\hfill%
\hfill%
\hfill%
\includegraphics[width=0.087\linewidth]{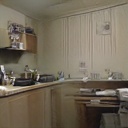}%
\hfill%
\includegraphics[width=0.087\linewidth]{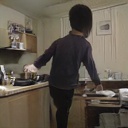}%
\hfill%
\hfill%
\hfill%
\includegraphics[width=0.087\linewidth]{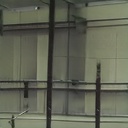}%
\hfill%
\includegraphics[width=0.087\linewidth]{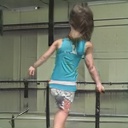}%
\hfill%
\hfill%
\hfill%
\includegraphics[width=0.087\linewidth]{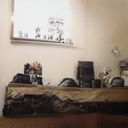}%
\hfill%
\includegraphics[width=0.087\linewidth]{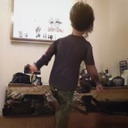}%
\hfill%
\hfill%
\hfill%
\includegraphics[width=0.087\linewidth]{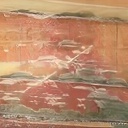}%
\hfill%
\includegraphics[width=0.087\linewidth]{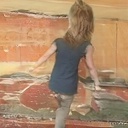}%
\end{subfigure}

% \vspace{0.9pt}

\begin{subfigure}[b]{1.0\linewidth}
\textoverlay{E3}{\includegraphics[width=0.087\linewidth]{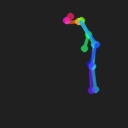}}%
\hfill%
\hfill%
\hfill%
\includegraphics[width=0.087\linewidth]{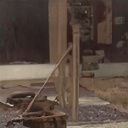}%
\hfill%
\includegraphics[width=0.087\linewidth]{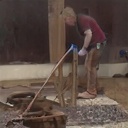}%
\hfill%
\hfill%
\hfill%
\includegraphics[width=0.087\linewidth]{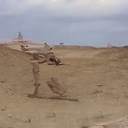}%
\hfill%
\includegraphics[width=0.087\linewidth]{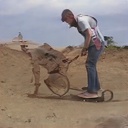}%
\hfill%
\hfill%
\hfill%
\includegraphics[width=0.087\linewidth]{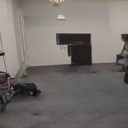}%
\hfill%
\includegraphics[width=0.087\linewidth]{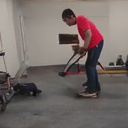}%
\hfill%
\hfill%
\hfill%
\includegraphics[width=0.087\linewidth]{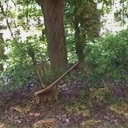}%
\hfill%
\includegraphics[width=0.087\linewidth]{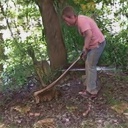}%
\hfill%
\hfill%
\hfill%
\includegraphics[width=0.087\linewidth]{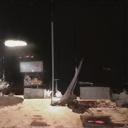}%
\hfill%
\includegraphics[width=0.087\linewidth]{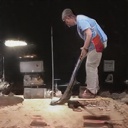}%
\end{subfigure}

% \vspace{0.9pt}

\begin{subfigure}[b]{1.0\linewidth}
\textoverlay{F3}{\includegraphics[width=0.087\linewidth]{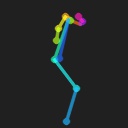}}%
\hfill%
\hfill%
\hfill%
\includegraphics[width=0.087\linewidth]{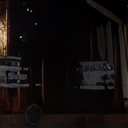}%
\hfill%
\includegraphics[width=0.087\linewidth]{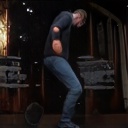}%
\hfill%
\hfill%
\hfill%
\includegraphics[width=0.087\linewidth]{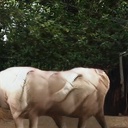}%
\hfill%
\includegraphics[width=0.087\linewidth]{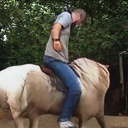}%
\hfill%
\hfill%
\hfill%
\includegraphics[width=0.087\linewidth]{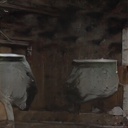}%
\hfill%
\includegraphics[width=0.087\linewidth]{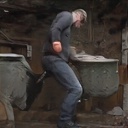}%
\hfill%
\hfill%
\hfill%
\includegraphics[width=0.087\linewidth]{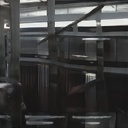}%
\hfill%
\includegraphics[width=0.087\linewidth]{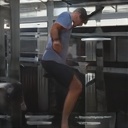}%
\hfill%
\hfill%
\hfill%
\includegraphics[width=0.087\linewidth]{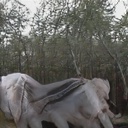}%
\hfill%
\includegraphics[width=0.087\linewidth]{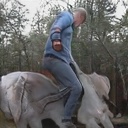}%
\end{subfigure}

% \vspace{0.9pt}

\begin{subfigure}[b]{1.0\linewidth}
\textoverlay{G3}{\includegraphics[width=0.087\linewidth]{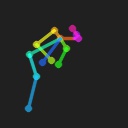}}%
\hfill%
\hfill%
\hfill%
\includegraphics[width=0.087\linewidth]{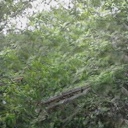}%
\hfill%
\includegraphics[width=0.087\linewidth]{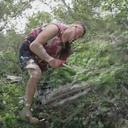}%
\hfill%
\hfill%
\hfill%
\includegraphics[width=0.087\linewidth]{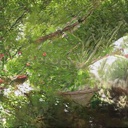}%
\hfill%
\includegraphics[width=0.087\linewidth]{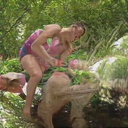}%
\hfill%
\hfill%
\hfill%
\includegraphics[width=0.087\linewidth]{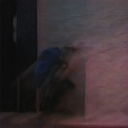}%
\hfill%
\includegraphics[width=0.087\linewidth]{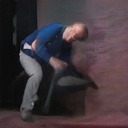}%
\hfill%
\hfill%
\hfill%
\includegraphics[width=0.087\linewidth]{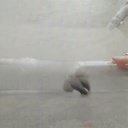}%
\hfill%
\includegraphics[width=0.087\linewidth]{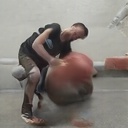}%
\hfill%
\hfill%
\hfill%
\includegraphics[width=0.087\linewidth]{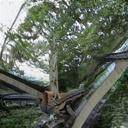}%
\hfill%
\includegraphics[width=0.087\linewidth]{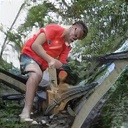}%
\end{subfigure}

% \vspace{0.9pt}

\begin{subfigure}[b]{1.0\linewidth}
\textoverlay{H3}{\includegraphics[width=0.087\linewidth]{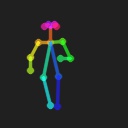}}%
\hfill%
\hfill%
\hfill%
\includegraphics[width=0.087\linewidth]{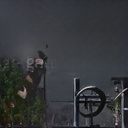}%
\hfill%
\includegraphics[width=0.087\linewidth]{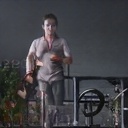}%
\hfill%
\hfill%
\hfill%
\includegraphics[width=0.087\linewidth]{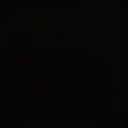}%
\hfill%
\includegraphics[width=0.087\linewidth]{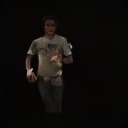}%
\hfill%
\hfill%
\hfill%
\includegraphics[width=0.087\linewidth]{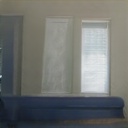}%
\hfill%
\includegraphics[width=0.087\linewidth]{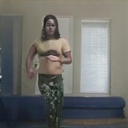}%
\hfill%
\hfill%
\hfill%
\includegraphics[width=0.087\linewidth]{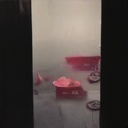}%
\hfill%
\includegraphics[width=0.087\linewidth]{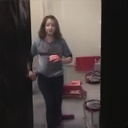}%
\hfill%
\hfill%
\hfill%
\includegraphics[width=0.087\linewidth]{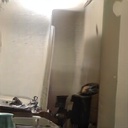}%
\hfill%
\includegraphics[width=0.087\linewidth]{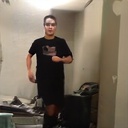}%
\end{subfigure}

% \vspace{-5pt}

\caption{\textbf{Random samples.}}
\label{fig:random_results_3}

\end{figure*}

\begin{figure*}[h!]
\centering

\begin{subfigure}[b]{1.0\linewidth}
\textoverlay{I3}{\includegraphics[width=0.087\linewidth]{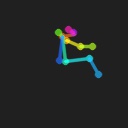}}%
\hfill%
\hfill%
\hfill%
\includegraphics[width=0.087\linewidth]{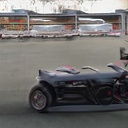}%
\hfill%
\includegraphics[width=0.087\linewidth]{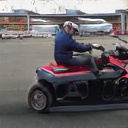}%
\hfill%
\hfill%
\hfill%
\includegraphics[width=0.087\linewidth]{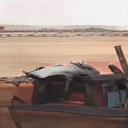}%
\hfill%
\includegraphics[width=0.087\linewidth]{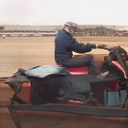}%
\hfill%
\hfill%
\hfill%
\includegraphics[width=0.087\linewidth]{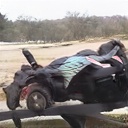}%
\hfill%
\includegraphics[width=0.087\linewidth]{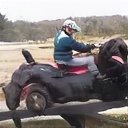}%
\hfill%
\hfill%
\hfill%
\includegraphics[width=0.087\linewidth]{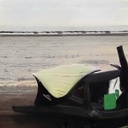}%
\hfill%
\includegraphics[width=0.087\linewidth]{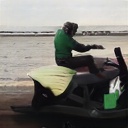}%
\hfill%
\hfill%
\hfill%
\includegraphics[width=0.087\linewidth]{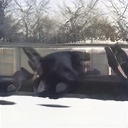}%
\hfill%
\includegraphics[width=0.087\linewidth]{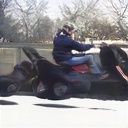}%
\end{subfigure}

% \vspace{0.9pt}

\begin{subfigure}[b]{1.0\linewidth}
\textoverlay{J3}{\includegraphics[width=0.087\linewidth]{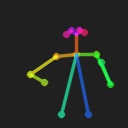}}%
\hfill%
\hfill%
\hfill%
\includegraphics[width=0.087\linewidth]{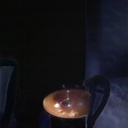}%
\hfill%
\includegraphics[width=0.087\linewidth]{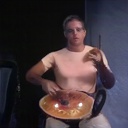}%
\hfill%
\hfill%
\hfill%
\includegraphics[width=0.087\linewidth]{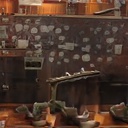}%
\hfill%
\includegraphics[width=0.087\linewidth]{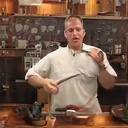}%
\hfill%
\hfill%
\hfill%
\includegraphics[width=0.087\linewidth]{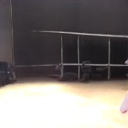}%
\hfill%
\includegraphics[width=0.087\linewidth]{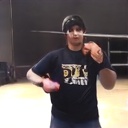}%
\hfill%
\hfill%
\hfill%
\includegraphics[width=0.087\linewidth]{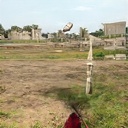}%
\hfill%
\includegraphics[width=0.087\linewidth]{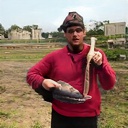}%
\hfill%
\hfill%
\hfill%
\includegraphics[width=0.087\linewidth]{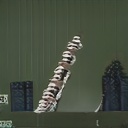}%
\hfill%
\includegraphics[width=0.087\linewidth]{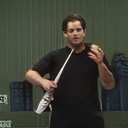}%
\end{subfigure}

% \vspace{0.9pt}

\begin{subfigure}[b]{1.0\linewidth}
\textoverlay{K3}{\includegraphics[width=0.087\linewidth]{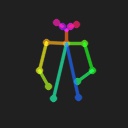}}%
\hfill%
\hfill%
\hfill%
\includegraphics[width=0.087\linewidth]{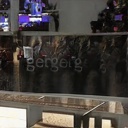}%
\hfill%
\includegraphics[width=0.087\linewidth]{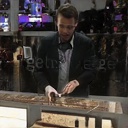}%
\hfill%
\hfill%
\hfill%
\includegraphics[width=0.087\linewidth]{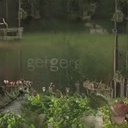}%
\hfill%
\includegraphics[width=0.087\linewidth]{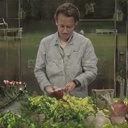}%
\hfill%
\hfill%
\hfill%
\includegraphics[width=0.087\linewidth]{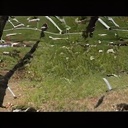}%
\hfill%
\includegraphics[width=0.087\linewidth]{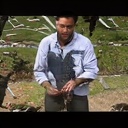}%
\hfill%
\hfill%
\hfill%
\includegraphics[width=0.087\linewidth]{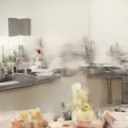}%
\hfill%
\includegraphics[width=0.087\linewidth]{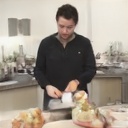}%
\hfill%
\hfill%
\hfill%
\includegraphics[width=0.087\linewidth]{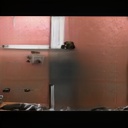}%
\hfill%
\includegraphics[width=0.087\linewidth]{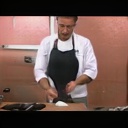}%
\end{subfigure}

% \vspace{0.9pt}

\begin{subfigure}[b]{1.0\linewidth}
\textoverlay{L3}{\includegraphics[width=0.087\linewidth]{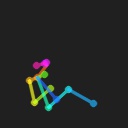}}%
\hfill%
\hfill%
\hfill%
\includegraphics[width=0.087\linewidth]{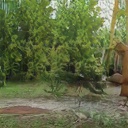}%
\hfill%
\includegraphics[width=0.087\linewidth]{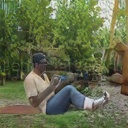}%
\hfill%
\hfill%
\hfill%
\includegraphics[width=0.087\linewidth]{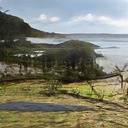}%
\hfill%
\includegraphics[width=0.087\linewidth]{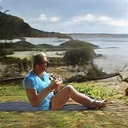}%
\hfill%
\hfill%
\hfill%
\includegraphics[width=0.087\linewidth]{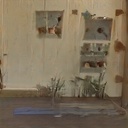}%
\hfill%
\includegraphics[width=0.087\linewidth]{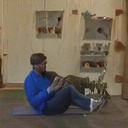}%
\hfill%
\hfill%
\hfill%
\includegraphics[width=0.087\linewidth]{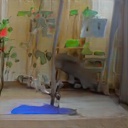}%
\hfill%
\includegraphics[width=0.087\linewidth]{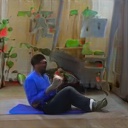}%
\hfill%
\hfill%
\hfill%
\includegraphics[width=0.087\linewidth]{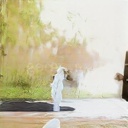}%
\hfill%
\includegraphics[width=0.087\linewidth]{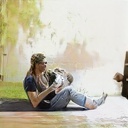}%
\end{subfigure}

% \vspace{0.9pt}

\begin{subfigure}[b]{1.0\linewidth}
\textoverlay{M3}{\includegraphics[width=0.087\linewidth]{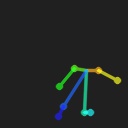}}%
\hfill%
\hfill%
\hfill%
\includegraphics[width=0.087\linewidth]{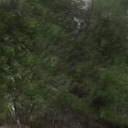}%
\hfill%
\includegraphics[width=0.087\linewidth]{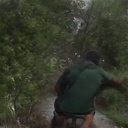}%
\hfill%
\hfill%
\hfill%
\includegraphics[width=0.087\linewidth]{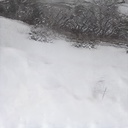}%
\hfill%
\includegraphics[width=0.087\linewidth]{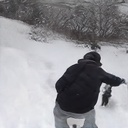}%
\hfill%
\hfill%
\hfill%
\includegraphics[width=0.087\linewidth]{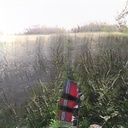}%
\hfill%
\includegraphics[width=0.087\linewidth]{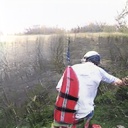}%
\hfill%
\hfill%
\hfill%
\includegraphics[width=0.087\linewidth]{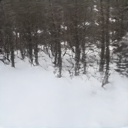}%
\hfill%
\includegraphics[width=0.087\linewidth]{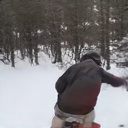}%
\hfill%
\hfill%
\hfill%
\includegraphics[width=0.087\linewidth]{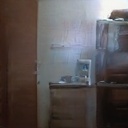}%
\hfill%
\includegraphics[width=0.087\linewidth]{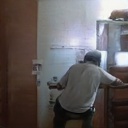}%
\end{subfigure}

% \vspace{0.9pt}

\begin{subfigure}[b]{1.0\linewidth}
\textoverlay{N3}{\includegraphics[width=0.087\linewidth]{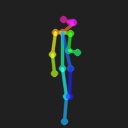}}%
\hfill%
\hfill%
\hfill%
\includegraphics[width=0.087\linewidth]{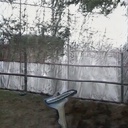}%
\hfill%
\includegraphics[width=0.087\linewidth]{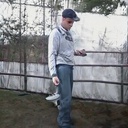}%
\hfill%
\hfill%
\hfill%
\includegraphics[width=0.087\linewidth]{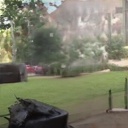}%
\hfill%
\includegraphics[width=0.087\linewidth]{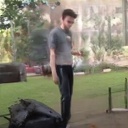}%
\hfill%
\hfill%
\hfill%
\includegraphics[width=0.087\linewidth]{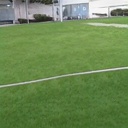}%
\hfill%
\includegraphics[width=0.087\linewidth]{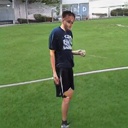}%
\hfill%
\hfill%
\hfill%
\includegraphics[width=0.087\linewidth]{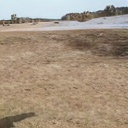}%
\hfill%
\includegraphics[width=0.087\linewidth]{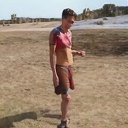}%
\hfill%
\hfill%
\hfill%
\includegraphics[width=0.087\linewidth]{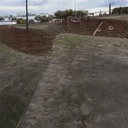}%
\hfill%
\includegraphics[width=0.087\linewidth]{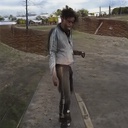}%
\end{subfigure}

% \vspace{0.9pt}

\begin{subfigure}[b]{1.0\linewidth}
\textoverlay{O3}{\includegraphics[width=0.087\linewidth]{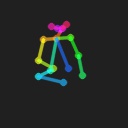}}%
\hfill%
\hfill%
\hfill%
\includegraphics[width=0.087\linewidth]{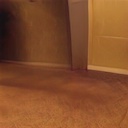}%
\hfill%
\includegraphics[width=0.087\linewidth]{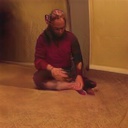}%
\hfill%
\hfill%
\hfill%
\includegraphics[width=0.087\linewidth]{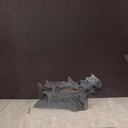}%
\hfill%
\includegraphics[width=0.087\linewidth]{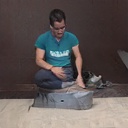}%
\hfill%
\hfill%
\hfill%
\includegraphics[width=0.087\linewidth]{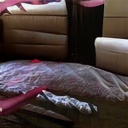}%
\hfill%
\includegraphics[width=0.087\linewidth]{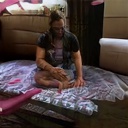}%
\hfill%
\hfill%
\hfill%
\includegraphics[width=0.087\linewidth]{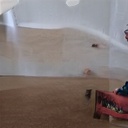}%
\hfill%
\includegraphics[width=0.087\linewidth]{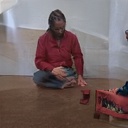}%
\hfill%
\hfill%
\hfill%
\includegraphics[width=0.087\linewidth]{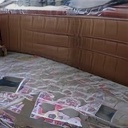}%
\hfill%
\includegraphics[width=0.087\linewidth]{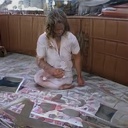}%
\end{subfigure}

% \vspace{0.9pt}

\begin{subfigure}[b]{1.0\linewidth}
\textoverlay{P3}{\includegraphics[width=0.087\linewidth]{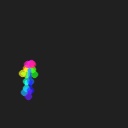}}%
\hfill%
\hfill%
\hfill%
\includegraphics[width=0.087\linewidth]{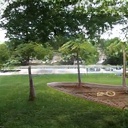}%
\hfill%
\includegraphics[width=0.087\linewidth]{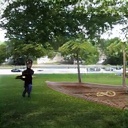}%
\hfill%
\hfill%
\hfill%
\includegraphics[width=0.087\linewidth]{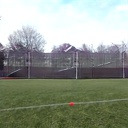}%
\hfill%
\includegraphics[width=0.087\linewidth]{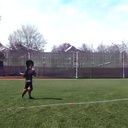}%
\hfill%
\hfill%
\hfill%
\includegraphics[width=0.087\linewidth]{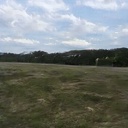}%
\hfill%
\includegraphics[width=0.087\linewidth]{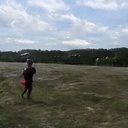}%
\hfill%
\hfill%
\hfill%
\includegraphics[width=0.087\linewidth]{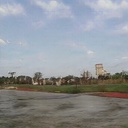}%
\hfill%
\includegraphics[width=0.087\linewidth]{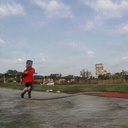}%
\hfill%
\hfill%
\hfill%
\includegraphics[width=0.087\linewidth]{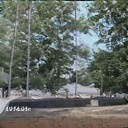}%
\hfill%
\includegraphics[width=0.087\linewidth]{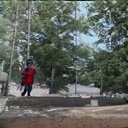}%
\end{subfigure}

% \vspace{0.9pt}

\begin{subfigure}[b]{1.0\linewidth}
\textoverlay{Q3}{\includegraphics[width=0.087\linewidth]{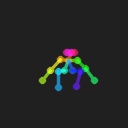}}%
\hfill%
\hfill%
\hfill%
\includegraphics[width=0.087\linewidth]{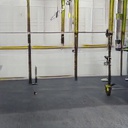}%
\hfill%
\includegraphics[width=0.087\linewidth]{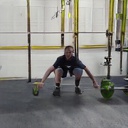}%
\hfill%
\hfill%
\hfill%
\includegraphics[width=0.087\linewidth]{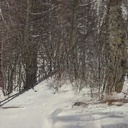}%
\hfill%
\includegraphics[width=0.087\linewidth]{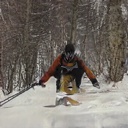}%
\hfill%
\hfill%
\hfill%
\includegraphics[width=0.087\linewidth]{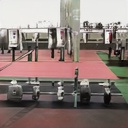}%
\hfill%
\includegraphics[width=0.087\linewidth]{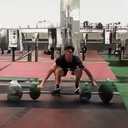}%
\hfill%
\hfill%
\hfill%
\includegraphics[width=0.087\linewidth]{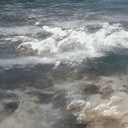}%
\hfill%
\includegraphics[width=0.087\linewidth]{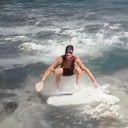}%
\hfill%
\hfill%
\hfill%
\includegraphics[width=0.087\linewidth]{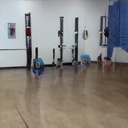}%
\hfill%
\includegraphics[width=0.087\linewidth]{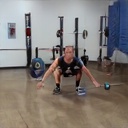}%
\end{subfigure}

% \vspace{0.9pt}

\begin{subfigure}[b]{1.0\linewidth}
\textoverlay{R3}{\includegraphics[width=0.087\linewidth]{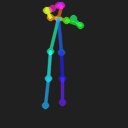}}%
\hfill%
\hfill%
\hfill%
\includegraphics[width=0.087\linewidth]{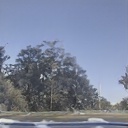}%
\hfill%
\includegraphics[width=0.087\linewidth]{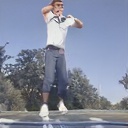}%
\hfill%
\hfill%
\hfill%
\includegraphics[width=0.087\linewidth]{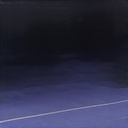}%
\hfill%
\includegraphics[width=0.087\linewidth]{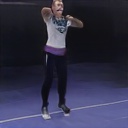}%
\hfill%
\hfill%
\hfill%
\includegraphics[width=0.087\linewidth]{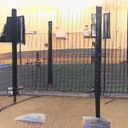}%
\hfill%
\includegraphics[width=0.087\linewidth]{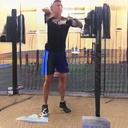}%
\hfill%
\hfill%
\hfill%
\includegraphics[width=0.087\linewidth]{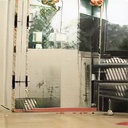}%
\hfill%
\includegraphics[width=0.087\linewidth]{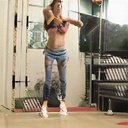}%
\hfill%
\hfill%
\hfill%
\includegraphics[width=0.087\linewidth]{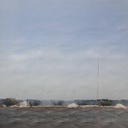}%
\hfill%
\includegraphics[width=0.087\linewidth]{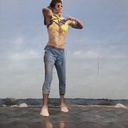}%
\end{subfigure}

% \vspace{0.9pt}

\begin{subfigure}[b]{1.0\linewidth}
\textoverlay{S3}{\includegraphics[width=0.087\linewidth]{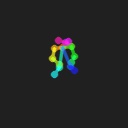}}%
\hfill%
\hfill%
\hfill%
\includegraphics[width=0.087\linewidth]{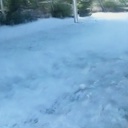}%
\hfill%
\includegraphics[width=0.087\linewidth]{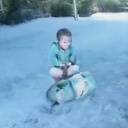}%
\hfill%
\hfill%
\hfill%
\includegraphics[width=0.087\linewidth]{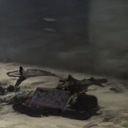}%
\hfill%
\includegraphics[width=0.087\linewidth]{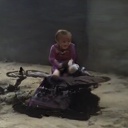}%
\hfill%
\hfill%
\hfill%
\includegraphics[width=0.087\linewidth]{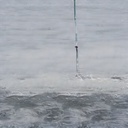}%
\hfill%
\includegraphics[width=0.087\linewidth]{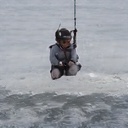}%
\hfill%
\hfill%
\hfill%
\includegraphics[width=0.087\linewidth]{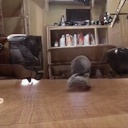}%
\hfill%
\includegraphics[width=0.087\linewidth]{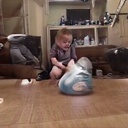}%
\hfill%
\hfill%
\hfill%
\includegraphics[width=0.087\linewidth]{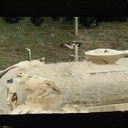}%
\hfill%
\includegraphics[width=0.087\linewidth]{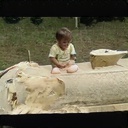}%
\end{subfigure}

% \vspace{0.9pt}

\begin{subfigure}[b]{1.0\linewidth}
\textoverlay{T3}{\includegraphics[width=0.087\linewidth]{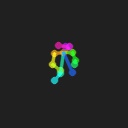}}%
\hfill%
\hfill%
\hfill%
\includegraphics[width=0.087\linewidth]{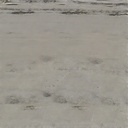}%
\hfill%
\includegraphics[width=0.087\linewidth]{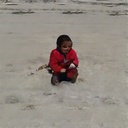}%
\hfill%
\hfill%
\hfill%
\includegraphics[width=0.087\linewidth]{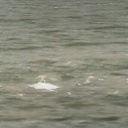}%
\hfill%
\includegraphics[width=0.087\linewidth]{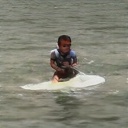}%
\hfill%
\hfill%
\hfill%
\includegraphics[width=0.087\linewidth]{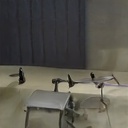}%
\hfill%
\includegraphics[width=0.087\linewidth]{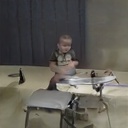}%
\hfill%
\hfill%
\hfill%
\includegraphics[width=0.087\linewidth]{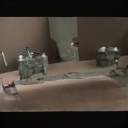}%
\hfill%
\includegraphics[width=0.087\linewidth]{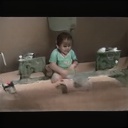}%
\hfill%
\hfill%
\hfill%
\includegraphics[width=0.087\linewidth]{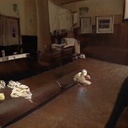}%
\hfill%
\includegraphics[width=0.087\linewidth]{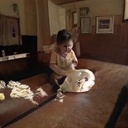}%
\end{subfigure}

% \vspace{0.9pt}

\begin{subfigure}[b]{1.0\linewidth}
\textoverlay{U3}{\includegraphics[width=0.087\linewidth]{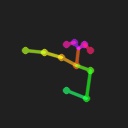}}%
\hfill%
\hfill%
\hfill%
\includegraphics[width=0.087\linewidth]{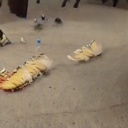}%
\hfill%
\includegraphics[width=0.087\linewidth]{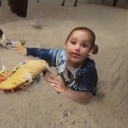}%
\hfill%
\hfill%
\hfill%
\includegraphics[width=0.087\linewidth]{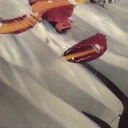}%
\hfill%
\includegraphics[width=0.087\linewidth]{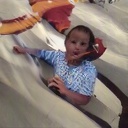}%
\hfill%
\hfill%
\hfill%
\includegraphics[width=0.087\linewidth]{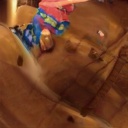}%
\hfill%
\includegraphics[width=0.087\linewidth]{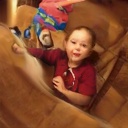}%
\hfill%
\hfill%
\hfill%
\includegraphics[width=0.087\linewidth]{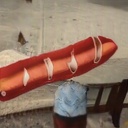}%
\hfill%
\includegraphics[width=0.087\linewidth]{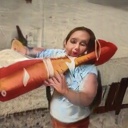}%
\hfill%
\hfill%
\hfill%
\includegraphics[width=0.087\linewidth]{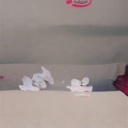}%
\hfill%
\includegraphics[width=0.087\linewidth]{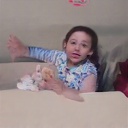}%
\end{subfigure}

% \vspace{0.9pt}

\begin{subfigure}[b]{1.0\linewidth}
\textoverlay{V3}{\includegraphics[width=0.087\linewidth]{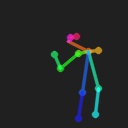}}%
\hfill%
\hfill%
\hfill%
\includegraphics[width=0.087\linewidth]{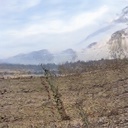}%
\hfill%
\includegraphics[width=0.087\linewidth]{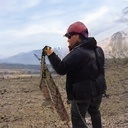}%
\hfill%
\hfill%
\hfill%
\includegraphics[width=0.087\linewidth]{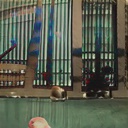}%
\hfill%
\includegraphics[width=0.087\linewidth]{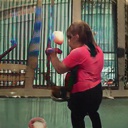}%
\hfill%
\hfill%
\hfill%
\includegraphics[width=0.087\linewidth]{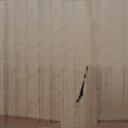}%
\hfill%
\includegraphics[width=0.087\linewidth]{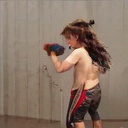}%
\hfill%
\hfill%
\hfill%
\includegraphics[width=0.087\linewidth]{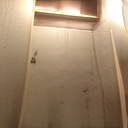}%
\hfill%
\includegraphics[width=0.087\linewidth]{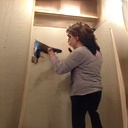}%
\hfill%
\hfill%
\hfill%
\includegraphics[width=0.087\linewidth]{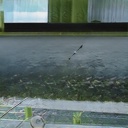}%
\hfill%
\includegraphics[width=0.087\linewidth]{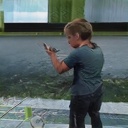}%
\end{subfigure}

% \vspace{0.9pt}

\begin{subfigure}[b]{1.0\linewidth}
\textoverlay{W3}{\includegraphics[width=0.087\linewidth]{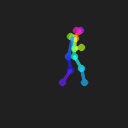}}%
\hfill%
\hfill%
\hfill%
\includegraphics[width=0.087\linewidth]{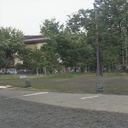}%
\hfill%
\includegraphics[width=0.087\linewidth]{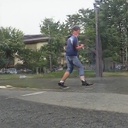}%
\hfill%
\hfill%
\hfill%
\includegraphics[width=0.087\linewidth]{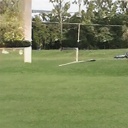}%
\hfill%
\includegraphics[width=0.087\linewidth]{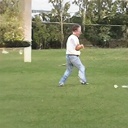}%
\hfill%
\hfill%
\hfill%
\includegraphics[width=0.087\linewidth]{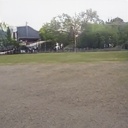}%
\hfill%
\includegraphics[width=0.087\linewidth]{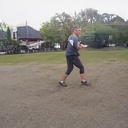}%
\hfill%
\hfill%
\hfill%
\includegraphics[width=0.087\linewidth]{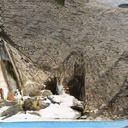}%
\hfill%
\includegraphics[width=0.087\linewidth]{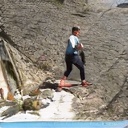}%
\hfill%
\hfill%
\hfill%
\includegraphics[width=0.087\linewidth]{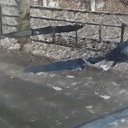}%
\hfill%
\includegraphics[width=0.087\linewidth]{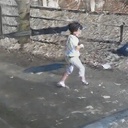}%
\end{subfigure}

% \vspace{-5pt}

\caption{\textbf{Random samples.}}
\label{fig:random_results_4}

\end{figure*}

Please see Figures~\ref{fig:random_results_0}~\ref{fig:random_results_1}~\ref{fig:random_results_2}~\ref{fig:random_results_3}~\ref{fig:random_results_4} for random uncurated samples from our model.

\end{document}